\documentclass[10pt,twocolumn,letterpaper]{article}

\usepackage{cvpr}
\usepackage{times}
\usepackage{graphicx}
\usepackage{amsmath}
\usepackage{amssymb}
\usepackage{bm}

\usepackage{comment}
\usepackage{color}
\usepackage{rotating}
\usepackage{overpic}

\usepackage[
   pagebackref=false,
   breaklinks=true,
   letterpaper=true,
   colorlinks,
   bookmarks=false]{hyperref}

\cvprfinalcopy
\def\cvprPaperID{1542}

\newcommand{\Loss}{\mathcal{L}}           
\newcommand{\LossLog}{{\Loss}^{\log{}}}   

\begin{document}

\title{\vspace{-18pt}\resizebox{\textwidth}{!}{%
   Affinity CNN: %
      Learning Pixel-Centric Pairwise Relations for Figure/Ground Embedding%
}}

\author{
   ~Michael Maire\\
   TTI Chicago\\
   {\tt\small mmaire@ttic.edu}\\
   \and
   ~Takuya Narihira\\
   Sony Corp.\\
   {\tt\small takuya.narihira@jp.sony.com}\\
   \and
   Stella X. Yu\\
   UC Berkeley / ICSI\\
   {\tt \small stellayu@berkeley.edu}
}

\maketitle

\begin{abstract}
Spectral embedding provides a framework for solving perceptual organization
problems, including image segmentation and figure/ground organization.  From an
affinity matrix describing pairwise relationships between pixels, it clusters
pixels into regions, and, using a complex-valued extension, orders pixels
according to layer.  We train a convolutional neural network (CNN) to directly
predict the pairwise relationships that define this affinity matrix.  Spectral
embedding then resolves these predictions into a globally-consistent
segmentation and figure/ground organization of the scene.  Experiments
demonstrate significant benefit to this direct coupling compared to prior works
which use explicit intermediate stages, such as edge detection, on the pathway
from image to affinities.  Our results suggest spectral embedding as a
powerful alternative to the conditional random field (CRF)-based globalization
schemes typically coupled to deep neural networks.
\end{abstract}

\section{Introduction}
\label{sec:introduction}

Systems for perceptual organization of scenes are commonly architected around a
pipeline of intermediate stages.  For example, image segmentation follows from
edge detection~\cite{FMM:CVPR:2003,UCM,gPb-UCM,SE,H4L}; figure/ground,
occlusion, or depth layering follows from reasoning over discrete contours or
regions~\cite{RFM:ECCV:2006,HSEH:ICCV:2007,LZLX:ICCV:2011,YHRF:PAMI:2011,
JGCC:CVPR:2012} with some systems also reliant on motion cues~\cite{
SH:IJCV:2009,HY:ECCV:2010,SBMAM:CVPR:2011,SLP:CVPR:2014}.  This trend holds
even in light of rapid advancements from designs centered on convolutional
neural networks (CNNs).  Rather than directly focus on image
segmentation, recent CNN architectures~\cite{N4,DeepEdge,DeepContour,H4L}
target edge detection.  Turaga~\etal~\cite{TMJRHBDS:Neural:2010} make the
connection between affinity learning and segmentation, yet restrict affinities
to be precisely local edge strengths.  Pure CNN approaches for depth from a
single image do focus on directly constructing the desired
output~\cite{EPF:NIPS:2014,EF:CVPR:2015}.  However, these works do not address
the problem of perceptual grouping without fixed semantic classes.

\begin{figure}
   \begin{center}
      \includegraphics[width=0.95\linewidth, clip=true, trim=0.25in 0.4in 0.40in 0.20in]{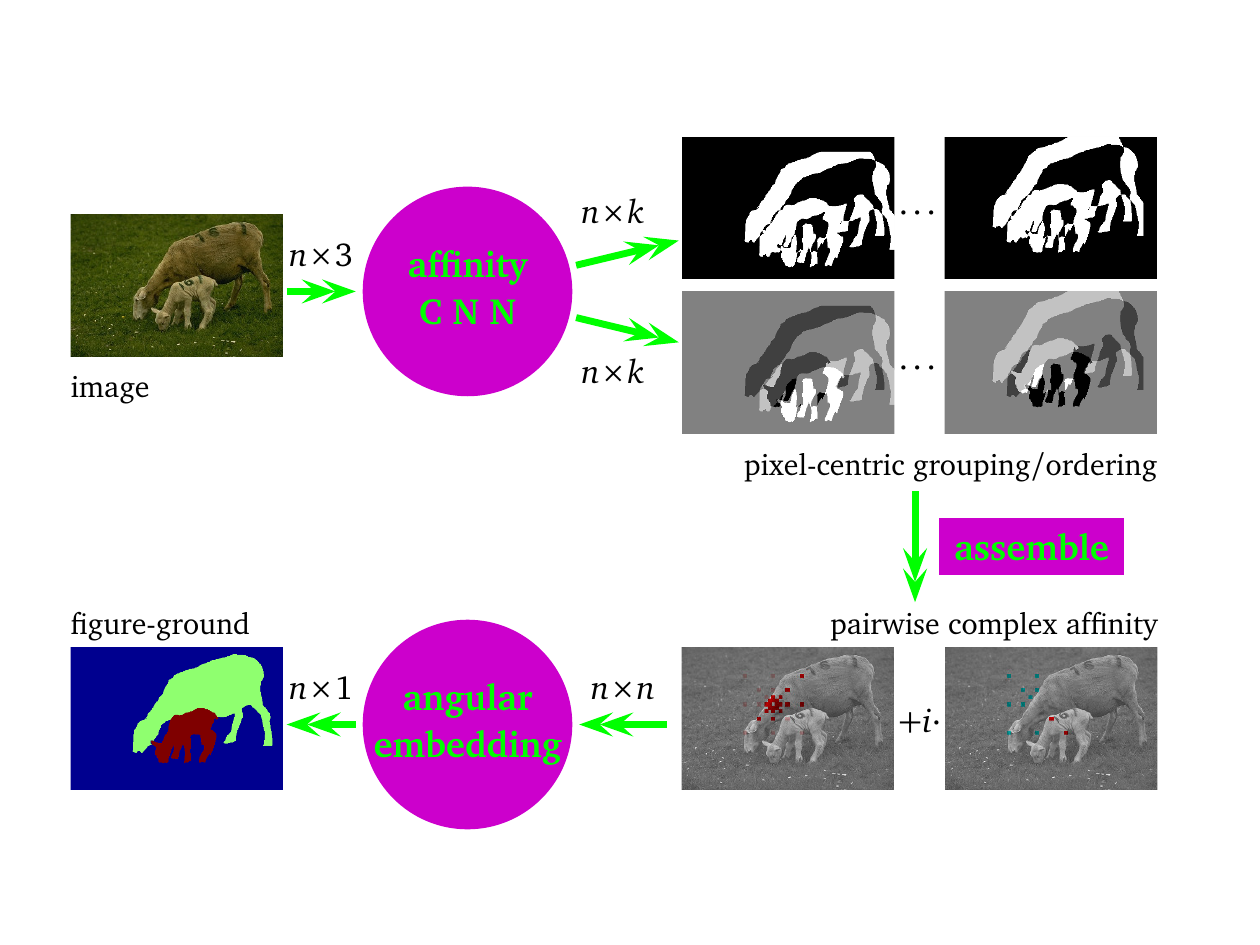}
   \end{center}
   \vspace{-0.05\linewidth}
   \caption{
      \textbf{System architecture.}
         We send an image through a CNN which is trained to predict the
         grouping and ordering relations between each of the $n$ pixels and
         its neighbors at $k$ displacements laid out in a fixed stencil
         pattern.  We assemble these $n\!\times\! 2k$ pixel-centric
         relations into a sparse $n\!\times\! n$ complex affinity matrix
         between pixels, each row indicating a pixel's affinity with others.
         Shown above is the row for the pixel at the center of a log-polar
         sampling pattern; its positive/negative relations with neighbors
         are marked by red/cyan squares overlaid on the image.  We feed the
         pairwise affinity matrix into Angular Embedding for global
         integration, producing an eigenvector representation that reveals
         figure-ground organization: we know not only which pixels go
         together, but also which pixels go in front.
   }
   \label{fig:system}
\end{figure}

We engineer a system for simultaneous segmentation and figure/ground
organization by directly connecting a CNN to an inference algorithm which
produces a globally consistent scene interpretation.  Training the CNN with
a target appropriate for the inference procedure eliminates the need for
hand-designed intermediate stages such as edge detection.  Our strategy
parallels recent work connecting CNNs and conditional random fields (CRFs) for
semantic segmentation~\cite{DeepLabCRF,LSRH:arXiv:2015,WSLCPY:ICCV:2015}.  A
crucial difference, however, is that we handle the generic, or class
independent, image partitioning problem.  In this context, spectral embedding,
and specifically Angular Embedding (AE)~\cite{Yu:CVPR:2009,AE}, is a more
natural inference algorithm.  Figure~\ref{fig:system} illustrates our
architecture.

Angular Embedding, an extension of the spectral relaxation of Normalized
Cuts~\cite{NCuts} to complex-valued affinities, provides a mathematical
framework for solving joint grouping and ranking problems.  Previous works
established this framework as a basis for segmentation and figure/ground
organization~\cite{Maire:ECCV:2010} as well as object-part grouping and
segmentation~\cite{MYP:ICCV:2011}.  We follow the spirit of~\cite{
Maire:ECCV:2010}, but employ major changes to achieve high-quality
figure/ground results:
\begin{itemize}
   \item{
      We reformulate segmentation and figure/ground layering in terms of an
      energy model with pairwise forces between pixels.  Pixels either bind
      together (group) or differentially repel (layer separation), with
      strength of interaction modulated by confidence in the prediction.
   }
   \item{
      We train a CNN to directly predict all data-dependent terms in the model.
   }
   \item{
      We predict interactions across multiple distance scales and use
      an efficient solver~\cite{MY:ICCV:2013} for spectral embedding.
   }
\end{itemize}

Our new energy model replaces the ad-hoc boundary-centric interactions employed
by~\cite{Maire:ECCV:2010}.  Our CNN replaces hand-designed features.  Together
they facilitate learning of pairwise interactions across a regular stencil
pattern.  Choosing a sparse stencil pattern, yet including both short- and
long-range connections, allows us to incorporate multiscale cues while
remaining computationally efficient.

Section~\ref{sec:spectral} develops our model for segmentation and
figure/ground while providing the necessary background on Angular Embedding.
Section~\ref{sec:learning} details the structure of our CNN for predicting
pairwise interaction terms in the model.

As our model is fully learned, it could be trained according to different
notions of segmentation and figure/ground.  For example, consistent definitions
for figure/ground include true depth ordering as in~\cite{EPF:NIPS:2014},
object class-specific foreground/background separation as in~\cite{
MYP:ICCV:2011}, and boundary ownership or occlusion as in~\cite{RFM:ECCV:2006,
FMM:JOV:2007,Maire:ECCV:2010}.  We focus on the latter and define segmentation
as a region partition and figure/ground as an ordering of regions by occlusion
layer.  The Berkeley segmentation dataset (BSDS) provides ground-truth
annotation of this form~\cite{BSDS,FMM:JOV:2007}.  We demonstrate segmentation
results competitive with the state-of-the-art on the BSDS
benchmark~\cite{BSDS-www}, while simultaneously generating high-quality
figure/ground output.

The occlusion layering interpretation of figure/ground is the one most likely
to be portable across datasets; it corresponds to a mid-level perceptual task.
We find this to be precisely the case for our learned model.  Trained on BSDS,
it generates quite reasonable output when tested on other image sources,
including the PASCAL VOC dataset~\cite{PASCAL}.  We believe this to be a
significant advance in fully automatic perceptual organization.
Section~\ref{sec:experiments} presents experimental results across all
datasets, while Section~\ref{sec:conclusion} concludes.

\section{Spectral Embedding \& Generalized Affinity}
\label{sec:spectral}

\begin{figure}
   \begin{center}
      \begin{minipage}[t]{0.29\linewidth}
         \vspace{0pt}
         \begin{scriptsize}
            Given Pairwise:\\
            \hspace*{5pt}Ordering~$\Theta(\cdot,\cdot)$\\
            \hspace*{5pt}Confidence~$C(\cdot,\cdot)$\\
            Recover:\\
            \hspace*{5pt}Global ordering $\theta(p)$\\
            \hspace*{5pt}$p \rightarrow z(p) = e^{i\theta(p)}$
         \end{scriptsize}
      \end{minipage}
      \hfill
      \begin{minipage}[t]{0.60\linewidth}
         \vspace{0pt}
         \begin{center}
         \hspace{-30pt}
         \begin{overpic}[width=1.0\linewidth]{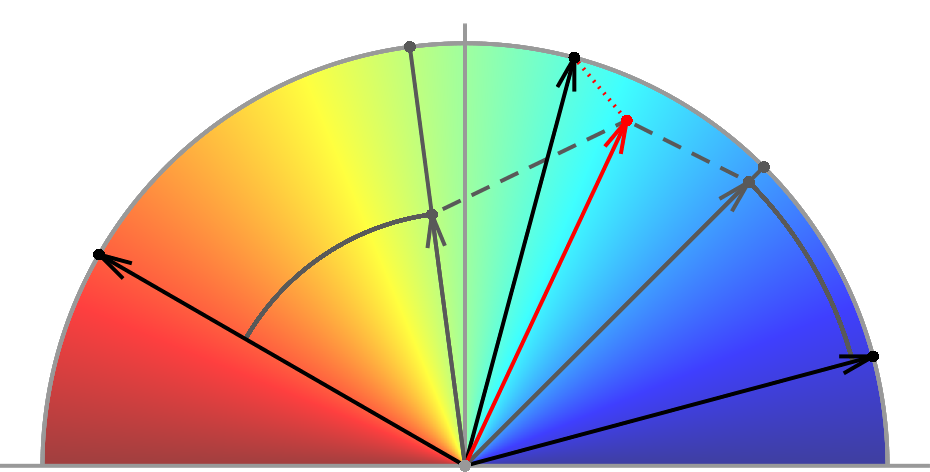}
            \definecolor{AxisColor}{rgb}{0.5,0.5,0.5}
            \definecolor{RotColor}{rgb}{0,0,0}
            \definecolor{OrderingColor}{rgb}{0,0,0}
            \definecolor{ConfidenceColor}{rgb}{0,0,0}
            \definecolor{PredColor}{rgb}{1,0,0}
            \put(50.5,42){\scriptsize{\textcolor{AxisColor}{$i$}}}
            \put(49,-4){\scriptsize{\textcolor{AxisColor}{$0$}}}
            \put(94.5,-4){\scriptsize{\textcolor{AxisColor}{$1$}}}
            \put(0,-4){\scriptsize{\textcolor{AxisColor}{$-1$}}}
            \put(61,46){\scriptsize{$z(p)$}}
            \put(95,12){\scriptsize{$z(q)$}}
            \put(0,25){\scriptsize{$z(r)$}}
            \put(33,48){\scriptsize{\textcolor{RotColor}{$z(r)e^{i\Theta(p,r)}$}}}
            \put(81,35){\scriptsize{\textcolor{RotColor}{$z(q)e^{i\Theta(p,q)}$}}}
            \put(47,5){\begin{rotate}{97.5}\scriptsize{\textcolor{ConfidenceColor}{$C(p,r)$}}\end{rotate}}
            \put(65.5,9.5){\begin{rotate}{45}\scriptsize{\textcolor{ConfidenceColor}{$C(p,q)$}}\end{rotate}}
            \put(26,19.5){\begin{rotate}{35}\scriptsize{\textcolor{OrderingColor}{$\Theta(p,r)$}}\end{rotate}}
            \put(79,26){\begin{rotate}{-58}\scriptsize{\textcolor{OrderingColor}{$\Theta(p,q)$}}\end{rotate}}
            \put(67.5,39){\small{\textcolor{PredColor}{$\bm{\tilde{z}(p)}$}}}
         \end{overpic}
         \end{center}
      \end{minipage}
   \end{center}
   \vspace{-0.01\linewidth}
   \caption{
      \textbf{Angular Embedding}~\cite{AE}\textbf{.}
         Given $(C,\Theta)$ capturing pairwise relationships between nodes,
         the Angular Embedding task is to map those nodes onto the unit
         semicircle, such that their resulting absolute positions respect
         confidence-weighted relative pairwise ordering
         (Equation~\ref{eq:ae_error}).  Relative ordering is identified with
         rotation in the complex plane.  For node $p$, $\theta(p) = \arg(z(p))$
         recovers its global rank order from its embedding $z(p)$.
   }
   \label{fig:ae}
\end{figure}
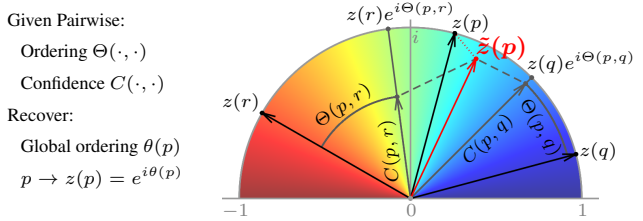

We abstract the figure/ground problem to that of assigning each pixel $p$ a
rank $\theta(p)$, such that $\theta(\cdot)$ orders pixels by occlusion layer.
Assume we are given estimates of the relative order $\Theta(p,q)$ between many
pairs of pixels $p$~and~$q$.  The task is then to find $\theta(\cdot)$ that
agrees as best as possible with these pairwise estimates.  Angular
Embedding~\cite{AE} addresses this optimization problem by minimizing error
$\varepsilon$:
\begin{equation}
   \varepsilon =
      \sum_p
         \frac{\sum_q C(p,q)}{\sum_{p,q} C(p,q)} \cdot |z(p)-\tilde{z}(p)|^2
   \label{eq:ae_error}
\end{equation}
where $C(p,q)$ accounts for possibly differing confidences in the pairwise
estimates and $\theta(p)$ is replaced by $z(p) = e^{i\theta(p)}$.  As
Figure~\ref{fig:ae} shows, this mathematical convenience permits interpretation
of $z(\cdot)$ as an embedding into the complex plane, with desired ordering
$\theta(\cdot)$ corresponding to absolute angle.  $\tilde{z}(p)$ is defined as
the consensus embedding location for $p$ according to its neighbors and
$\Theta$:
\begin{align}
   \tilde{z}(p) &= \sum_q \tilde{C}(p,q) \cdot e^{i\Theta(p,q)} \cdot z(q)\\
   \tilde{C}(p,q) &= \frac{C(p,q)}{\sum_q C(p,q)}
\end{align}

Relaxing the unit norm constraint on $z(\cdot)$ yields a generalized
eigenproblem:
\begin{equation}
   W z = \lambda D z
   \label{eq:ae_eig}
\end{equation}
with $D$ and $W$ defined in terms of $C$ and $\Theta$ by:
\begin{align}
   D &= \mathrm{Diag}(C1_n) \\
   W &= C \bullet e^{i\Theta}
\end{align}
where $n$ is the number of pixels, $1_n$ is a column vector of ones,
$\mathrm{Diag}(\cdot)$ is a matrix with its vector argument on the main
diagonal, and $\bullet$ denotes the matrix Hadamard product.

\begin{figure}
   \begin{center}
      \includegraphics{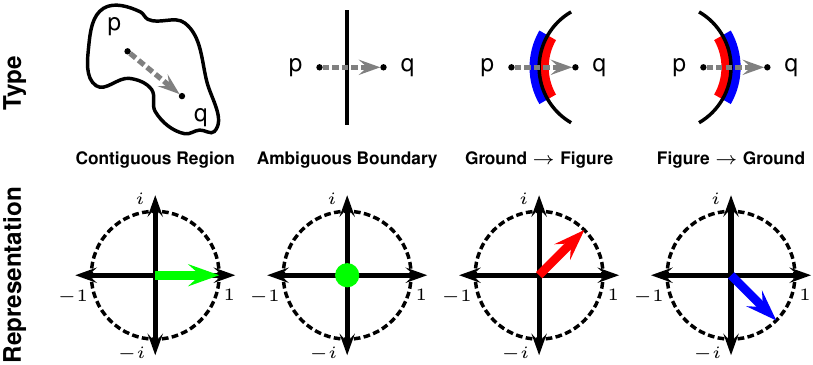}
   \end{center}
   \caption{
      \textbf{Complex affinities for grouping and figure/ground.}
         An angular displacement, corresponding to relative figure/ground or
         depth ordering, along with a confidence on that displacement, specify
         pairwise local grouping relationships between pixels.  A single
         complex number encodes confidence as magnitude and displacement as
         angle from the positive real axis.  Four basic interaction types span
         the space of possible pairwise pixel relationships.
      \emph{\textbf{Contiguous region:}}
         Pixels $p$ and $q$ lie in the same region.  A vector along the
         positive real axis represents high confidence on zero relative
         displacement.
      \emph{\textbf{Ambiguous boundary:}}
         Pixels $p$ and $q$ lie in different regions whose interface admits
         no cues for discriminating displacement.  The shared boundary could
         be a surface marking or depth discontinuity with either of $p$ or $q$
         in front.  The origin of the complex plane represents zero confidence
         on the correct relationship.
      \emph{\textbf{Figure transition:}}
         As boundary convexity tends to indicate foreground, moving from $p$
         to $q$ likely transitions from ground to figure.  We have high
         confidence on positive angular displacement.
      \emph{\textbf{Ground transition:}}
         In the reverse case, $q$ is ground with respect to $p$, and the
         complex representation has negative angle.
   }
   \label{fig:boundary_type}
\end{figure}

For $\Theta$ everywhere zero ($W = C$), this eigenproblem is identical to
the spectral relaxation of Normalized Cuts~\cite{NCuts}, in which the second
and higher eigenvectors encode grouping~\cite{NCuts,gPb-UCM}.  With nonzero
entries in $\Theta$, the first of the now complex-valued eigenvectors is
nontrivial and its angle encodes rank ordering while the subsequent
eigenvectors still encode grouping~\cite{Maire:ECCV:2010}.  We use the same
decoding procedure as~\cite{Maire:ECCV:2010} to read off this information.

Specifically, given eigenvectors, $\{z_0,z_1,...,z_{m-1}\}$, and corresponding
eigenvalues, $\lambda_0 \leq \lambda_1 \leq ... \leq \lambda_{m-1}$, solving
Equation~\ref{eq:ae_eig}, $\theta(p) = \arg(z_0(p))$ recovers figure/ground
ordering.  Treating the eigenvectors as an embedding of pixels into
$\mathbb{C}^m$, distance in this embedding space reveals perceptual grouping.
We follow~\cite{gPb-UCM,Maire:ECCV:2010} to recover both boundaries
and segmentation from the embedding by taking the (spatial) gradient of
eigenvectors and applying the watershed transform.  This is equivalent to a
form of agglomerative clustering in the embedding space, with merging
constrained to be between neighbors in the image domain.

A remaining issue, solved by~\cite{MYP:ICCV:2011}, is to avoid circular
wrap-around in angular span by guaranteeing that the solution fits within a
wedge of the complex plane.  It suffices to rescale $\Theta$ by
$\frac{\pi}{2}(1_{n}^T|\Theta|1_{n})^{-1}$ prior to embedding.

Having chosen Angular Embedding as our inference procedure, it remains to
define the pairwise pixel relationships $C(p,q)$ and $\Theta(p,q)$.  In the
special case of Normalized Cuts, $C(p,q)$ represents a clustering affinity, or
confidence on zero separation (in both clustering and figure/ground).  For the
more general case, we must also predict non-zero figure/ground separation
values and assign them confidences.

\begin{figure}
   \begin{center}
      \begin{minipage}[t]{0.37\linewidth}
         \vspace{0pt}
         \begin{center}
            \includegraphics[width=1.0\linewidth, clip=true, trim=1.0in 2.50in 1.0in 2.50in]{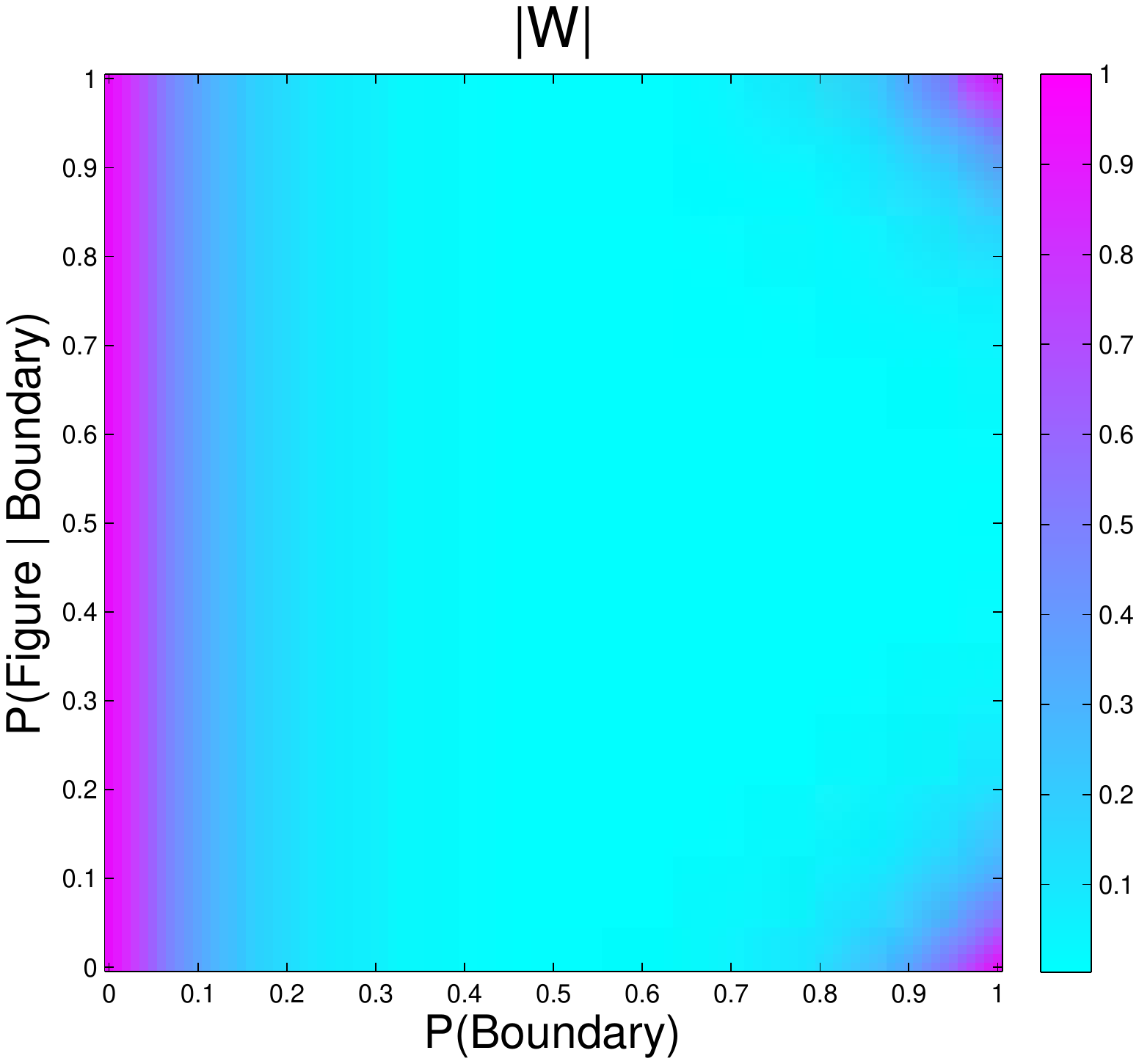}\\
            \includegraphics[width=1.0\linewidth, clip=true, trim=1.0in 2.50in 1.0in 2.50in]{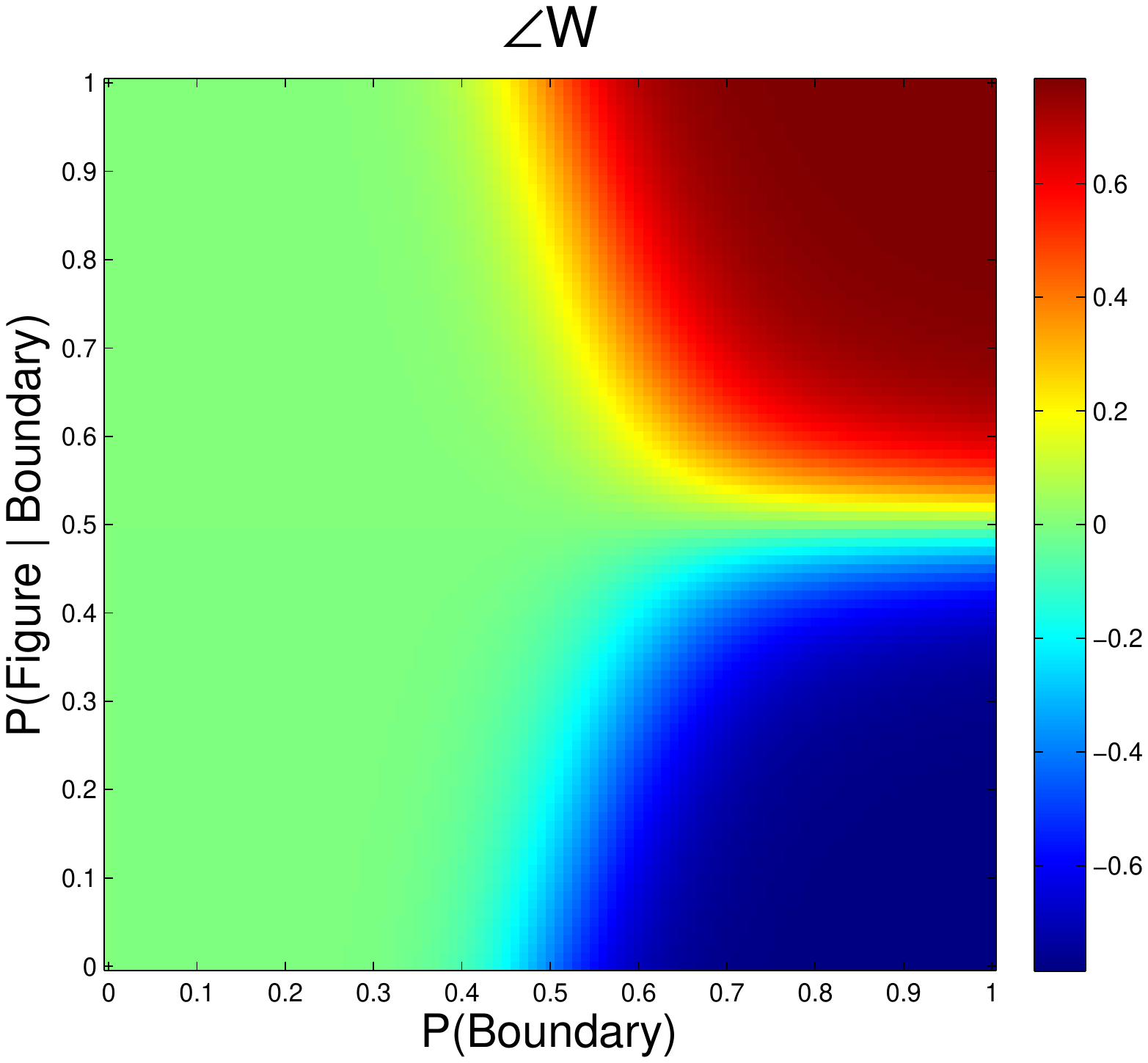}
         \end{center}
      \end{minipage}
      \hfill
      \begin{minipage}[t]{0.62\linewidth}
         \vspace{0.375cm}
         \begin{center}
            \includegraphics[width=1.00\linewidth, clip=true, trim=1.25in 2.7in 1.25in 2.55in]{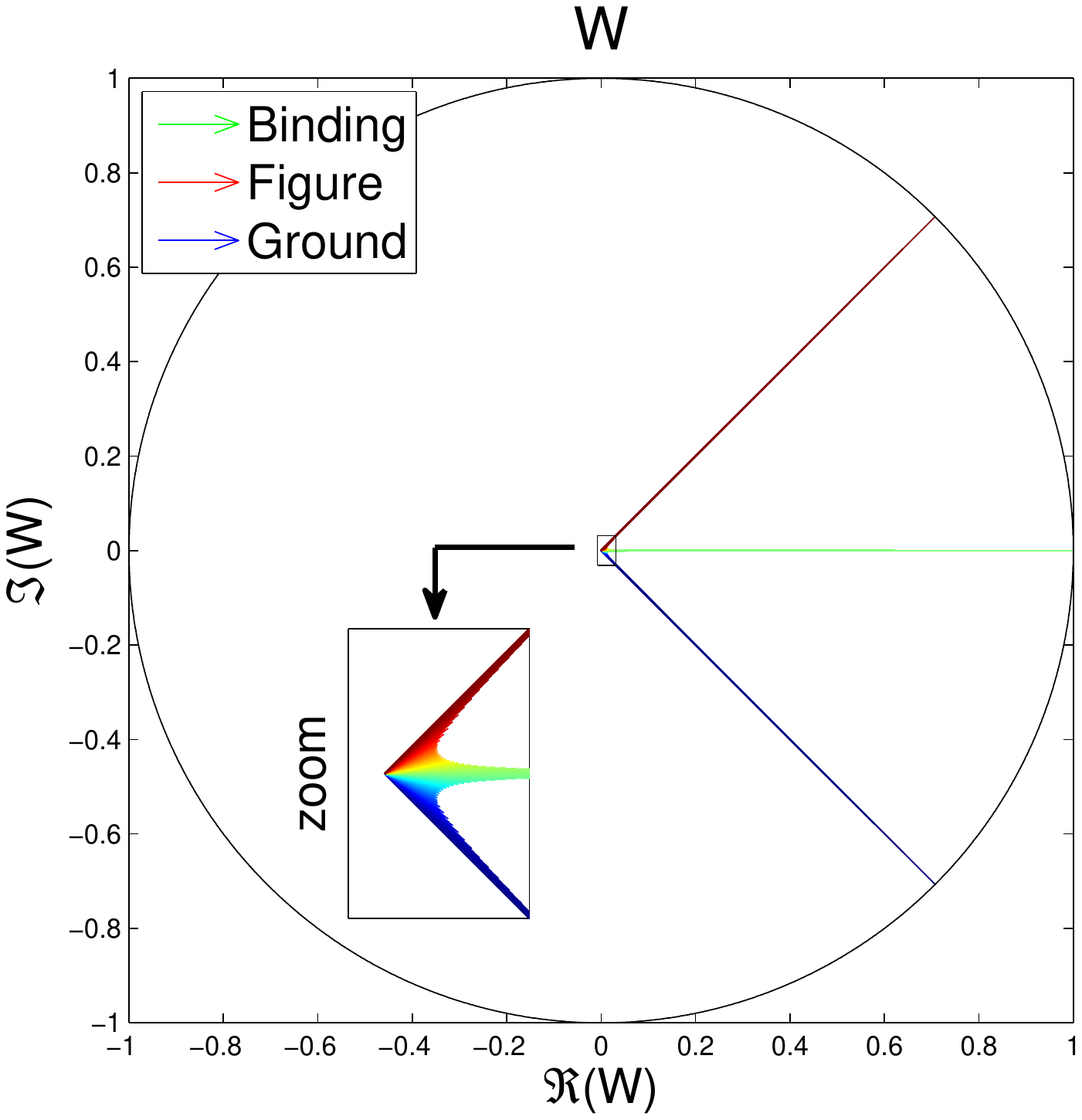}
         \end{center}
      \end{minipage}
   \end{center}
   \vspace{-0.01\linewidth}
   \caption{
      \textbf{Generalized affinity.}
         Combining the base cases in Figure~\ref{fig:boundary_type}, we express
         generalized affinity $W$ as the sum of a binding force acting along
         the positive real axis, and figure and ground displacement forces
         acting at angles.  In absence of any strong boundary, the binding
         force dominates, linking pixels together.  In presence of a strong
         and discriminative boundary, either the figure or ground force
         dominates, triggering displacement.  Under conditions of uncertainty,
         all forces are weak.
      \emph{\textbf{Left:}}
         The plot for $|W|$ illustrates total force strength, while the plot
         for $\angle{W}$ shows the dominant force.
      \emph{\textbf{Right:}}
         Complex-valued $W$ varies smoothly across its configuration space,
         yet  exhibits four distinct modes (binding, figure, ground,
         uncertain).  Smooth transitions occur in the region of uncertainty at
         the origin.
   }
   \label{fig:generalized_affinity}
\end{figure}

Let us develop the model in terms of probabilities:
\begin{align}
e(p)   &= Pr(p~\mathrm{lies~on~a~boundary})\\
b(p,q) &= Pr(\mathrm{seg}(p) \ne \mathrm{seg}(q))\\
f(p,q) &= Pr(\mathrm{figural}(p,q)~|~\mathrm{seg}(p) \ne \mathrm{seg}(q)))
\label{eq:f}\\
g(p,q) &= Pr(\mathrm{figural}(q,p)~|~\mathrm{seg}(p) \ne \mathrm{seg}(q)))
\end{align}
where $\mathrm{seg}(p)$ is the region (segment) containing pixel $p$ and
$\mathrm{figural}(p,q)$ means that $q$ is figure with respect to $p$,
according to the true segmentation and figure/ground ordering.  $b(p,q)$ is
the probability that some boundary separates $p$ and $q$.  $f(p,q)$ and
$g(p,q)$ are conditional probabilities of figure and ground, respectively.
Note $g(p,q) = 1 - f(p,q)$.

There are three possible transitions between $p$ and $q$: none (same region),
ground $\rightarrow$ figure, and figure $\rightarrow$ ground.  Selecting the
most likely, the probabilities of erroneously binding $p$ and $q$ into the
same region, transitioning to figure, or transitioning to ground are
respectively:
\begin{align}
   E_B(p,q) &= b(p,q) \\
   E_F(p,q) &= 1 - (1-e(p))b(p,q)(1-e(q))f(p,q) \\
   E_G(p,q) &= 1 - (1-e(p))b(p,q)(1-e(q))g(p,q)
\end{align}
where $(1-e(p))b(p,q)(1-e(q))$ is the probability that there is a boundary
between $p$ and $q$, but that neither $p$ nor $q$ themselves lie on any
boundary.  Figure/ground repulsion forces act long-range and across boundaries.
We convert to confidence via exponential reweighting:
\begin{align}
   C_B(p,q) &= \exp(-E_B(p,q)/\sigma_b) \\
   C_F(p,q) &= \exp(-E_F(p,q)/\sigma_f) \\
   C_G(p,q) &= \exp(-E_G(p,q)/\sigma_g)
\end{align}
where $\sigma_b$ and $\sigma_f = \sigma_g$ control scaling.  Using a fixed
angle $\phi$ for the rotational action of figure/ground transitions
($\Theta(p,q) = \pm \phi$), we obtain complex-valued affinities:
\begin{align}
   W_B(p,q) &= C_B(p,q) \\
   W_F(p,q) &= C_F(p,q)\exp(i\phi) \\
   W_G(p,q) &= C_G(p,q)\exp(-i\phi)
\end{align}

Figure~\ref{fig:boundary_type} illustrates how $W_B$ (shown in green), $W_F$
(red), and $W_G$ (blue) cover the base cases in the space of pairwise
grouping relationships.  Combining them into a single energy model
(generalized affinity) spans the entire space:
\begin{equation}
   W(p,q) = W_B(p,q) + W_F(p,q) + W_G(p,q)
\end{equation}
One can regard $W(p,q)$ as a sum of binding, figure transition, and ground
transition forces acting between $p$ and $q$.
Figure~\ref{fig:generalized_affinity} plots the configuration space of
$W(p,q)$ in terms of $b(p,q)$ and $f(p,q)$.  As the areas of this space in
which each component force is strong do not overlap, $W$ behaves in distinct
modes, with a smooth transition between them through the area of weak affinity
near the origin.

Learning to predict $e(p)$, $b(p,q)$, and $f(p,q)$ suffices to determine all
components of $W$.  For computational efficiency, we predict pairwise
relationships between each pixel and its immediate neighbors across multiple
spatial scales.  This defines a multiscale sparse $W$.  As an adjustment
prior to feeding $W$ to the Angular Embedding solver of~\cite{MY:ICCV:2013},
we enforce Hermitian symmetry by assigning:
\begin{equation}
   W \leftarrow (W + W^*)/2
\end{equation}

\section{Affinity Learning}
\label{sec:learning}

Supervised training of our system proceeds from a collection of images and
associated ground-truth, $\{(I_0,S_0,R_0),(I_1,S_1,R_1),\ldots\}$.  Here,
$I_k$ is an image defined on domain $\Omega_k \subset \mathbb{N}^2$.
$S_k: \Omega_k \rightarrow \mathbb{N}$ is a segmentation mapping
each pixel to a region id, and $R_k: \Omega_k \rightarrow \mathbb{R}$ is an
rank ordering of pixels according to figure/ground layering.  This data
defines ground-truth pairwise relationships:
\begin{align}
   \tilde{b}_k(p,q) &= 1 - \delta(S(p)-S(q))
   \label{eq:b_train}\\
   \tilde{f}_k(p,q) &= (sign(R(q)-R(p))+1)/2
   \label{eq:f_train}
\end{align}
As $f(p,q)$ is a conditional probability (Equation~\ref{eq:f}), we only
generate training examples $\tilde{f}_k(p,q)$ for pairs $(p,q)$ satisfying
$\tilde{b}_k(p,q) = 1$.

In all experiments, we sample pixel pairs $(p,q)$ from a multiscale stencil
pattern.  For each pixel $p$, we consider as $q$ each of its $8$ immediate
neighbors in the pixel grid, across $3$ scales (distances of $1$, $4$, and
$16$ pixels).  The stencil pattern thus consists of $24$ neighbors total.
We train $48$ predictors, $b(\cdot,\cdot)$ and $f(\cdot,\cdot)$ at each of the
$24$ offsets, for describing the pairwise affinity between a pixel and its
neighbors.  We derive the predictor $e(\cdot)$ as a local average of
$b(\cdot,\cdot)$:
\begin{equation}
   e(p) = \frac{1}{8} \sum_{q \in \mathcal{N}_1(p)} b(p,q)
\end{equation}
where $\mathcal{N}_1(p)$ consists of the $8$ neighbors to $p$ at fine-scale.

\begin{figure}
   \begin{center}
      \includegraphics[width=1.00\linewidth]{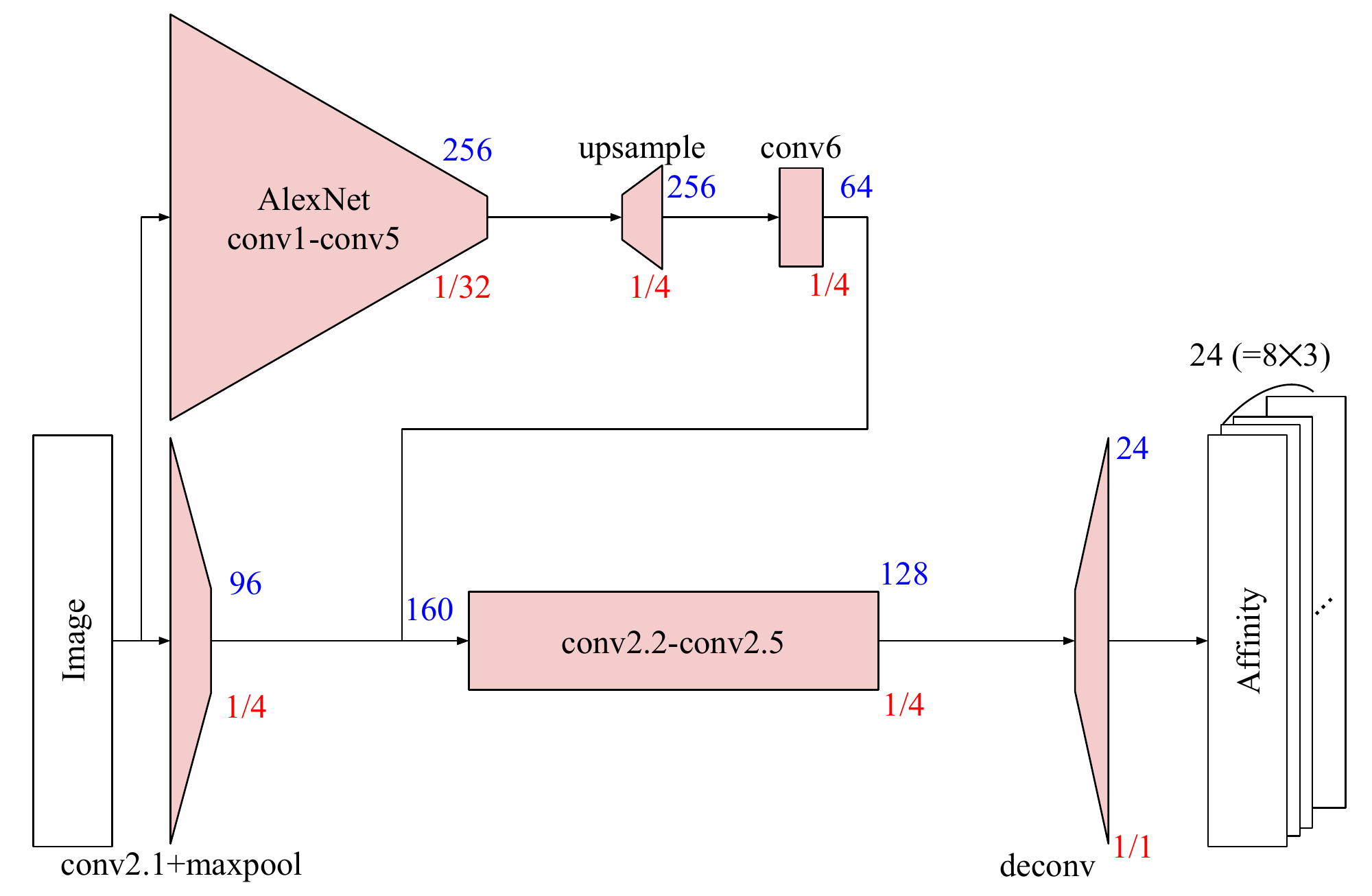}
   \end{center}
   \vspace{-0.01\linewidth}
   \caption{
      \textbf{Deep Affinity Network.}
         Our CNN produces an output feature map with spatial resolution
         matching the input image and whose channels encode pairwise affinity
         estimates between each pixel and $8$ neighbors across $3$ scales.
         Its internal architecture derives from that of previous networks
         \cite{EPF:NIPS:2014,EF:CVPR:2015,NMY:ICCV:2015} for generating
         per-pixel predictions using multiple receptive fields.  At each
         stage, red labels denote the spatial size ratio of feature map to
         input image.   Blue labels denote the number of channels in each
         feature map.
   }
   \label{fig:cnn}
\end{figure}

\begin{figure*}
   \setlength\fboxsep{0pt}
   \begin{center}
      \begin{minipage}[t]{0.12\linewidth}
         \vspace{0pt}
         \begin{center}
            \fbox{\includegraphics[width=1.00\linewidth]{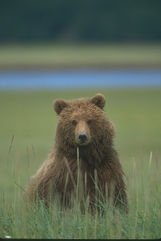}}\\
            \vspace{0.01\linewidth}
            \fbox{\includegraphics[width=1.00\linewidth]{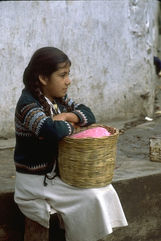}}\\
            \vspace{0.01\linewidth}
            \scriptsize{\textbf{\textsf{Image}}}
         \end{center}
      \end{minipage}
      \hfill
      \begin{minipage}[t]{0.372\linewidth}
         \vspace{0pt}
         \begin{center}
         \begin{minipage}[t]{0.32258\linewidth}
            \vspace{0pt}
            \begin{center}
               \fbox{\includegraphics[width=1.00\linewidth]{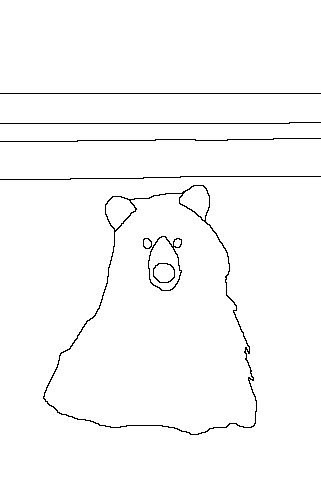}}\\
               \vspace{0.01\linewidth}
               \fbox{\includegraphics[width=1.00\linewidth]{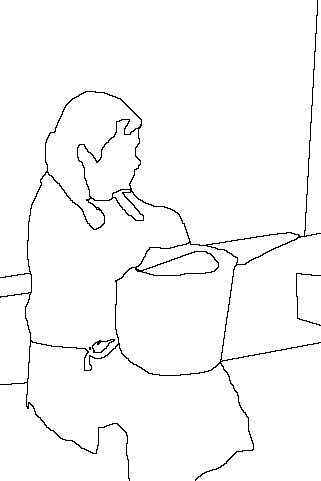}}\\
               \vspace{0.01\linewidth}
               \scriptsize{\textbf{\textsf{Segmentation}}}
            \end{center}
         \end{minipage}
         \begin{minipage}[t]{0.32258\linewidth}
            \vspace{0pt}
            \begin{center}
               \fbox{\includegraphics[width=1.00\linewidth]{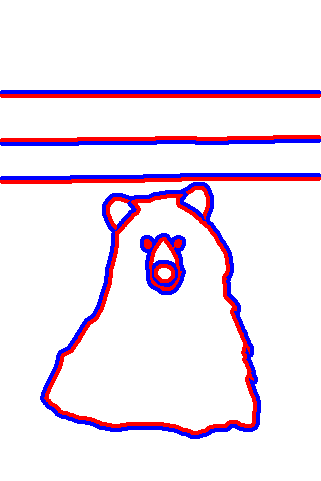}}\\
               \vspace{0.01\linewidth}
               \fbox{\includegraphics[width=1.00\linewidth]{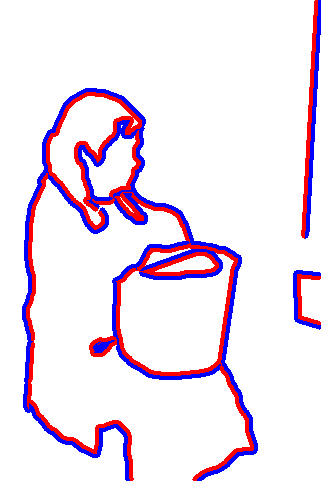}}\\
               \vspace{0.01\linewidth}
               \scriptsize{\textbf{\textsf{Local F/G Labels}}}
            \end{center}
         \end{minipage}
         \begin{minipage}[t]{0.32258\linewidth}
            \vspace{0pt}
            \begin{center}
               \fbox{\includegraphics[width=1.00\linewidth]{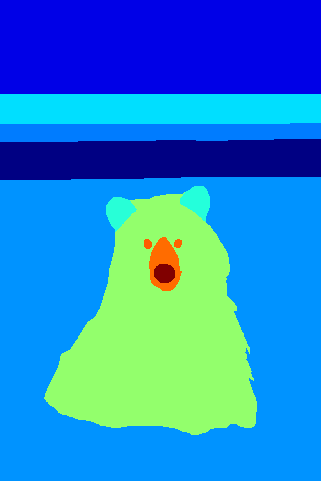}}\\
               \vspace{0.01\linewidth}
               \fbox{\includegraphics[width=1.00\linewidth]{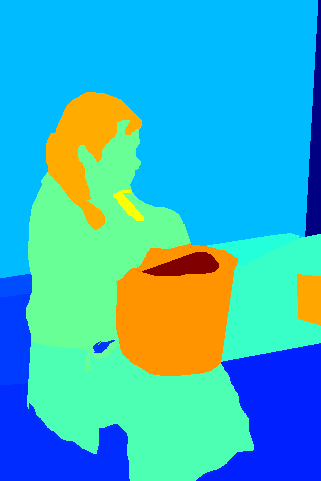}}\\
               \vspace{0.01\linewidth}
               \scriptsize{\textbf{\textsf{Globalized F/G}}}
            \end{center}
         \end{minipage}\\
         \vspace{1pt}
         \rule{0.925\linewidth}{0.5pt}\\
         \scriptsize{\textbf{\textsf{Ground-truth}}}
         \end{center}
      \end{minipage}
      \hfill
      \begin{minipage}[t]{0.496\linewidth}
         \vspace{0pt}
         \begin{center}
         \begin{minipage}[t]{0.24194\linewidth}
            \vspace{0pt}
            \begin{center}
               \fbox{\includegraphics[width=1.00\linewidth]{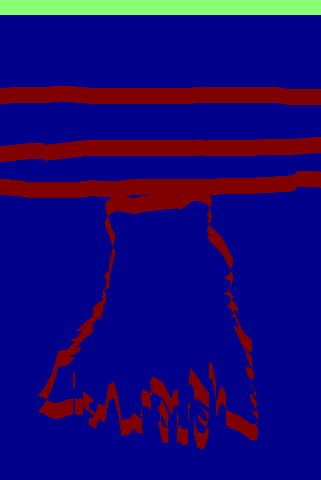}}\\
               \vspace{0.01\linewidth}
               \fbox{\includegraphics[width=1.00\linewidth]{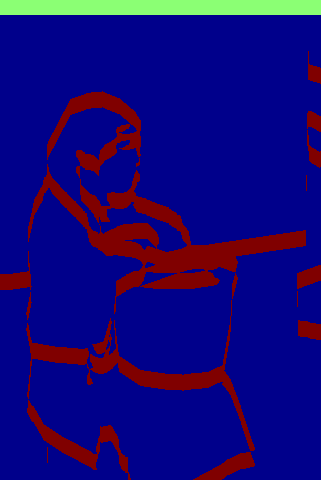}}\\
               \vspace{0.01\linewidth}
               \scriptsize{$\tilde{b}(p,p+d)$}
            \end{center}
         \end{minipage}
         \begin{minipage}[t]{0.24194\linewidth}
            \vspace{0pt}
            \begin{center}
               \fbox{\includegraphics[width=1.00\linewidth]{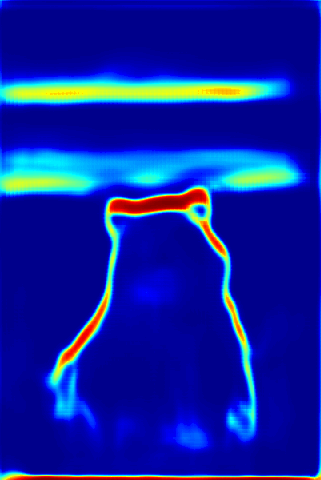}}\\
               \vspace{0.01\linewidth}
               \fbox{\includegraphics[width=1.00\linewidth]{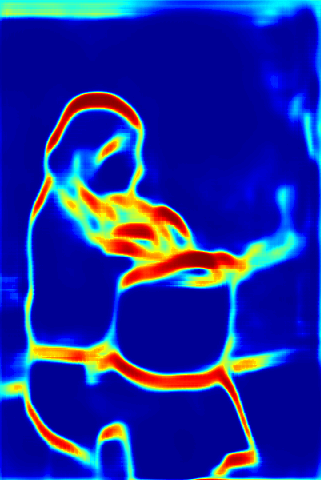}}\\
               \vspace{0.01\linewidth}
               \scriptsize{$b(p,p+d)$}
            \end{center}
         \end{minipage}
         \begin{minipage}[t]{0.24194\linewidth}
            \vspace{0pt}
            \begin{center}
               \fbox{\includegraphics[width=1.00\linewidth]{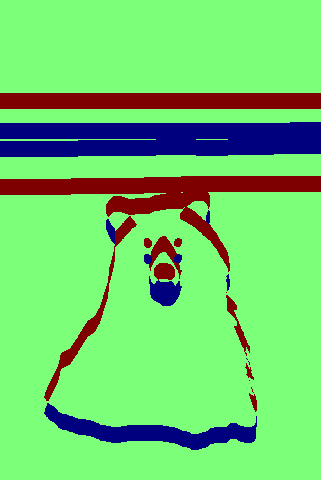}}\\
               \vspace{0.01\linewidth}
               \fbox{\includegraphics[width=1.00\linewidth]{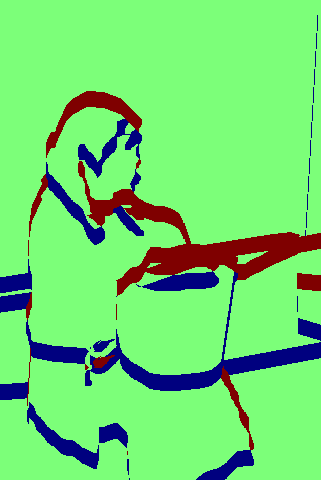}}\\
               \vspace{0.01\linewidth}
               \scriptsize{$\tilde{f}(p,p+d)$}
            \end{center}
         \end{minipage}
         \begin{minipage}[t]{0.24194\linewidth}
            \vspace{0pt}
            \begin{center}
               \fbox{\includegraphics[width=1.00\linewidth]{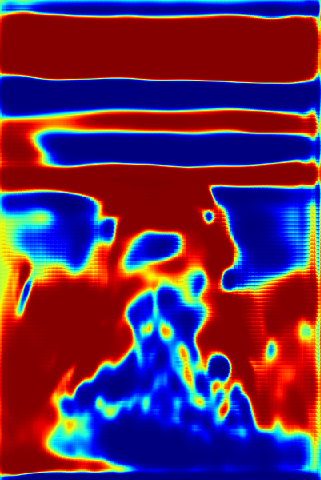}}\\
               \vspace{0.01\linewidth}
               \fbox{\includegraphics[width=1.00\linewidth]{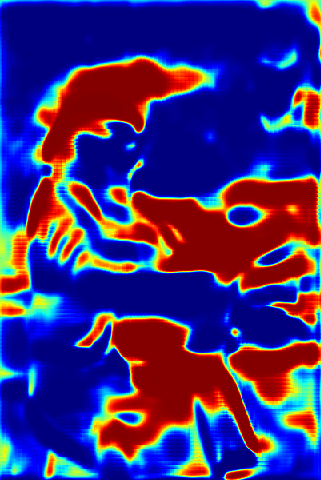}}\\
               \vspace{0.01\linewidth}
               \scriptsize{$f(p,p+d)$}
            \end{center}
         \end{minipage}\\
         \vspace{1pt}
         \rule{0.925\linewidth}{0.5pt}\\
         \scriptsize{\textbf{\textsf{Training}}}
         \end{center}
      \end{minipage}
   \end{center}
   \vspace{-0.01\linewidth}
   \caption{
      \textbf{Affinity learning for segmentation and figure/ground.}
      \emph{\textbf{Ground-truth assembly (left):}}
         Given only ground-truth segmentation~\cite{BSDS} and \emph{local}
         figure/ground labels in the form of boundary
         ownership~\cite{FMM:JOV:2007}, we infer a global ground-truth
         figure/ground order by running Angular Embedding with pairwise
         interactions defined by the local ground-truth.
      \emph{\textbf{Affinity training (right):}}
         The ground-truth segmentation serves to train pairwise grouping
         probability $b(\cdot,\cdot)$, while the globalized ground-truth
         figure/ground trains $f(\cdot,\cdot)$.  Shown are ground-truth
         training targets $\tilde{b}$, $\tilde{f}$, and model predictions
         $b$, $f$, for one component of our stencil: the relationship between
         pixel $p$ and its neighbor at relative offset $d = (-16,0)$.
         Ground-truth $\tilde{b}$ is binary (blue=$0$, red=$1$).  $\tilde{f}$
         is also binary, except pixel pairs in the same region (shown green)
         are ignored.  As $f$ is masked by $b$ at test time, we require only
         that $f(p,q)$ be correct when $b(p,q)$ is close to $1$.
   }
   \label{fig:bsds-training}
\end{figure*}

Choosing a CNN to implement these predictors, we regard the problem as mapping
an input image to a $48$ channel output over the same domain.  We adapt prior
CNN designs for predicting output quantities at every pixel~\cite{
EPF:NIPS:2014,EF:CVPR:2015,NMY:ICCV:2015} to our somewhat higher-dimensional
prediction task.  Specifically, we reuse the basic network design of 
\cite{NMY:ICCV:2015}, which first passes a large-scale coarse receptive field
through an AlexNet~\cite{alexNet12}-like subnetwork.  It appends this
subnetwork's output into a second scale subnetwork acting on a finer
receptive field.  Figure~\ref{fig:cnn} provides a complete layer diagram.
In modifying~\cite{NMY:ICCV:2015}, we increase the size of the penultimate
feature map as well as the output dimensionality.

For modularity at training time, we separately train two networks, one for
$b(\cdot,\cdot)$ and one for $f(\cdot,\cdot)$, each with the layer architecture
of Figure~\ref{fig:cnn}.  We use modified Caffe~\cite{caffe14} for training
with log loss between truth $\tilde{y}$ and prediction $y$ applied to each
output pixel-wise:
\begin{equation}
   \LossLog(\tilde{y}, y) = - \tilde{y}\log(y) - (1-\tilde{y})\log(1-y)
   \label{eq:loss_log}
\end{equation}
making the total loss for $b(\cdot,\cdot)$:
\begin{equation}
   \Loss =
      \frac{1}{|\Omega||\mathcal{N}(p)|}
         \sum_{p \in \Omega} \sum_{q \in \mathcal{N}(p)}
            \LossLog(\tilde{b}(p,q), b(p,q))
   \label{eq:loss}
\end{equation}
with an analogous loss applied for $f(\cdot,\cdot)$.  Here  $\mathcal{N}(p)$
denotes all $24$ neighbors of $p$ according to the stencil pattern.

Using stochastic gradient descent with random initialization and momentum of
$0.9$, we train with batch size $32$ for $5000$ mini-batch iterations.
Learning rates for each layer are tuned by hand.  We utilize data augmentation
in the form of translation and left-right mirroring of examples.

\begin{figure*}
   \setlength\fboxsep{0pt}
   \begin{center}
      \begin{minipage}[t]{0.33\linewidth}
         \vspace{0pt}
         \begin{center}
         \begin{minipage}[t]{0.48485\linewidth}
            \vspace{0pt}
            \begin{center}
               \fbox{\includegraphics[width=1.00\linewidth]{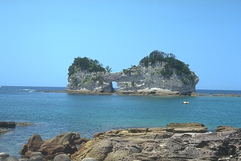}}\\
               \vspace{0.01\linewidth}
               \fbox{\includegraphics[width=1.00\linewidth]{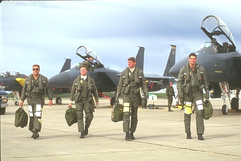}}\\
               \vspace{0.01\linewidth}
               \fbox{\includegraphics[width=1.00\linewidth]{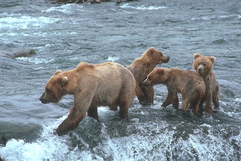}}\\
               \vspace{0.01\linewidth}
               \fbox{\includegraphics[width=1.00\linewidth]{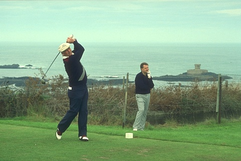}}\\
               \vspace{0.01\linewidth}
               \fbox{\includegraphics[width=1.00\linewidth]{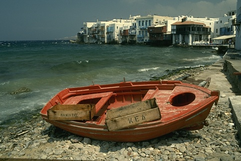}}\\
               \vspace{0.01\linewidth}
               \fbox{\includegraphics[width=1.00\linewidth]{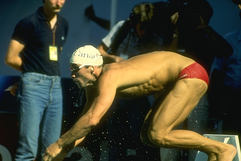}}\\
               \vspace{0.01\linewidth}
               \fbox{\includegraphics[width=1.00\linewidth]{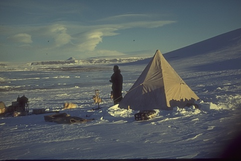}}\\
               \vspace{0.01\linewidth}
               \fbox{\includegraphics[width=1.00\linewidth]{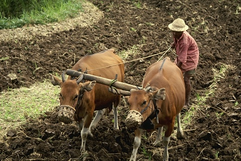}}\\
               \vspace{0.01\linewidth}
               \fbox{\includegraphics[width=1.00\linewidth]{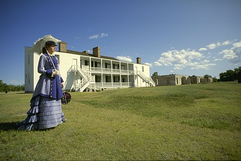}}\\
               \vspace{0.01\linewidth}
               \fbox{\includegraphics[width=1.00\linewidth]{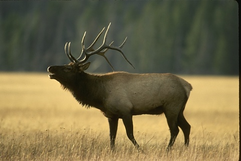}}\\
               \vspace{0.01\linewidth}
               \scriptsize{\textbf{\textsf{Image}}}
            \end{center}
         \end{minipage}
         \begin{minipage}[t]{0.48485\linewidth}
            \vspace{0pt}
            \begin{center}
               \fbox{\includegraphics[width=1.00\linewidth]{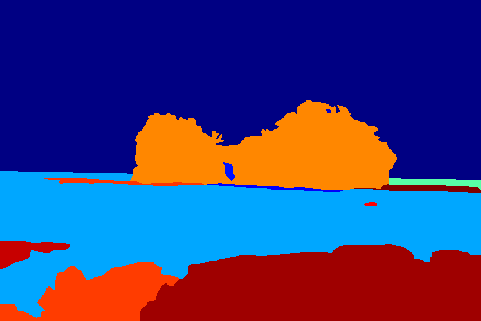}}\\
               \vspace{0.01\linewidth}
               \fbox{\includegraphics[width=1.00\linewidth]{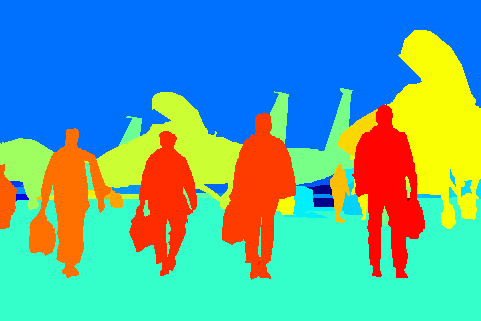}}\\
               \vspace{0.01\linewidth}
               \fbox{\includegraphics[width=1.00\linewidth]{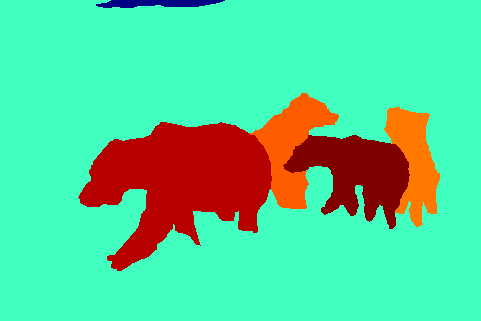}}\\
               \vspace{0.01\linewidth}
               \fbox{\includegraphics[width=1.00\linewidth]{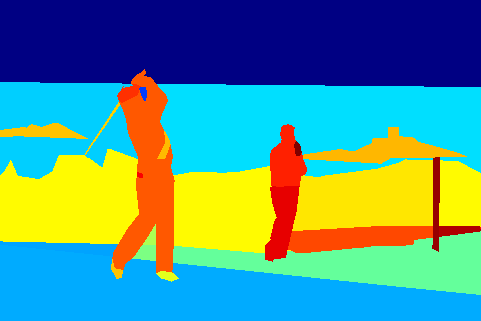}}\\
               \vspace{0.01\linewidth}
               \fbox{\includegraphics[width=1.00\linewidth]{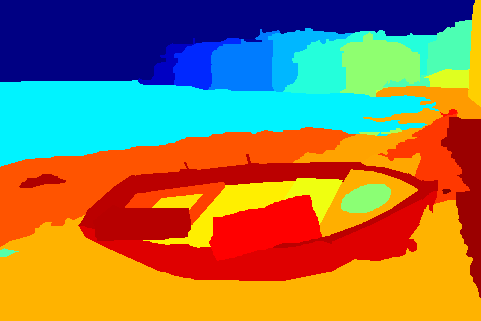}}\\
               \vspace{0.01\linewidth}
               \fbox{\includegraphics[width=1.00\linewidth]{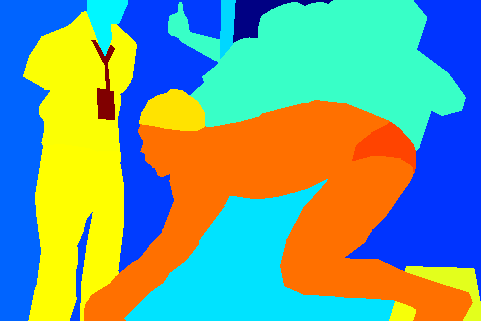}}\\
               \vspace{0.01\linewidth}
               \fbox{\includegraphics[width=1.00\linewidth]{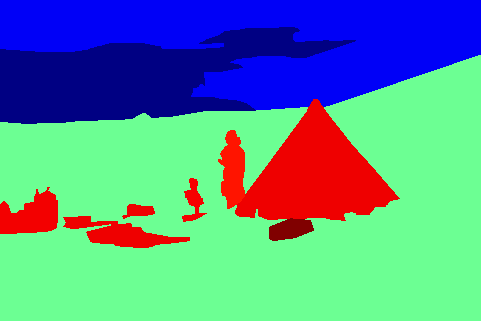}}\\
               \vspace{0.01\linewidth}
               \fbox{\includegraphics[width=1.00\linewidth]{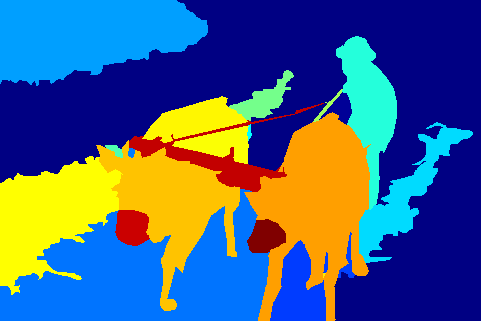}}\\
               \vspace{0.01\linewidth}
               \fbox{\includegraphics[width=1.00\linewidth]{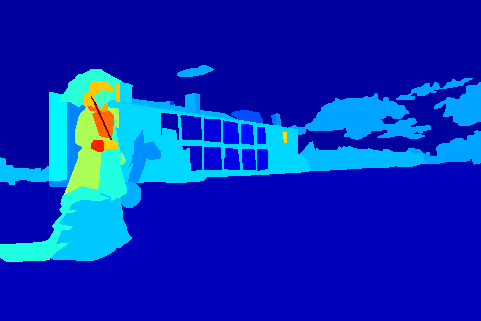}}\\
               \vspace{0.01\linewidth}
               \fbox{\includegraphics[width=1.00\linewidth]{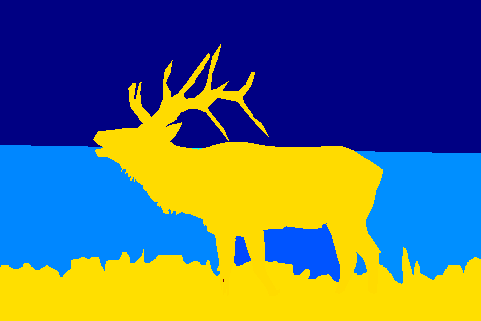}}\\
               \vspace{0.01\linewidth}
               \scriptsize{\textbf{\textsf{Ground-truth F/G}}}
            \end{center}
         \end{minipage}
         \end{center}
      \end{minipage}
      \hfill
      \begin{minipage}[t]{0.165\linewidth}
         \vspace{0pt}
         \begin{center}
         \begin{minipage}[t]{0.96970\linewidth}
            \vspace{0pt}
            \begin{center}
               \fbox{\includegraphics[width=1.00\linewidth]{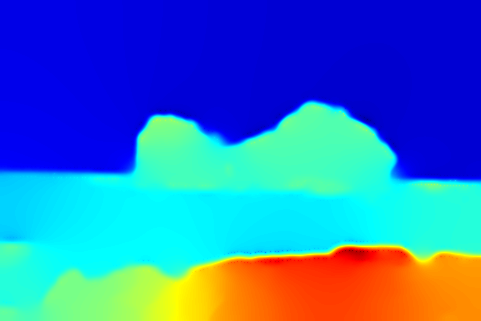}}\\
               \vspace{0.01\linewidth}
               \fbox{\includegraphics[width=1.00\linewidth]{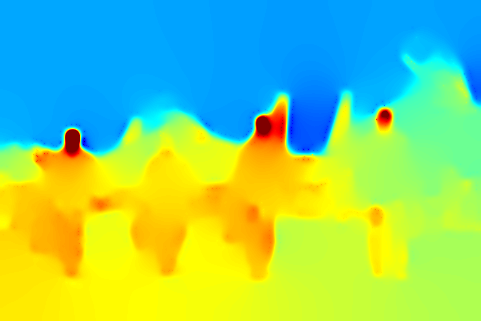}}\\
               \vspace{0.01\linewidth}
               \fbox{\includegraphics[width=1.00\linewidth]{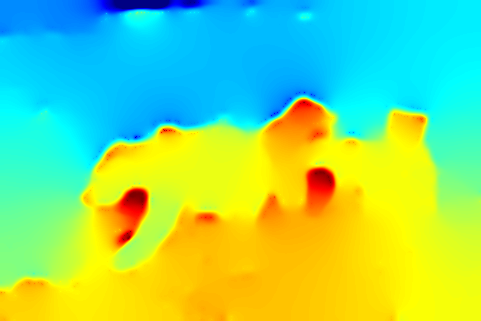}}\\
               \vspace{0.01\linewidth}
               \fbox{\includegraphics[width=1.00\linewidth]{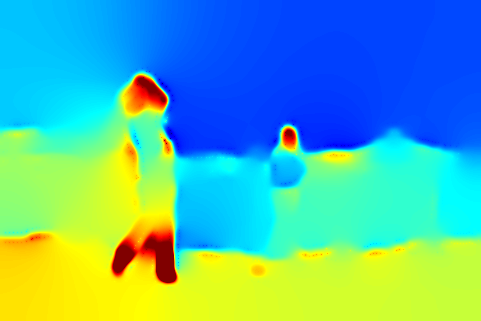}}\\
               \vspace{0.01\linewidth}
               \fbox{\includegraphics[width=1.00\linewidth]{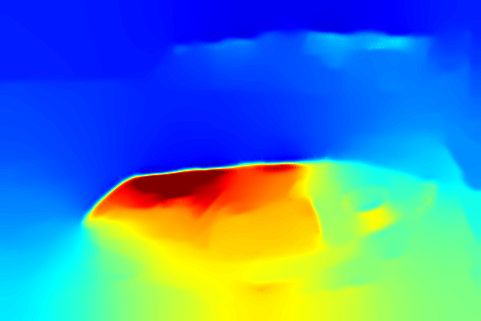}}\\
               \vspace{0.01\linewidth}
               \fbox{\includegraphics[width=1.00\linewidth]{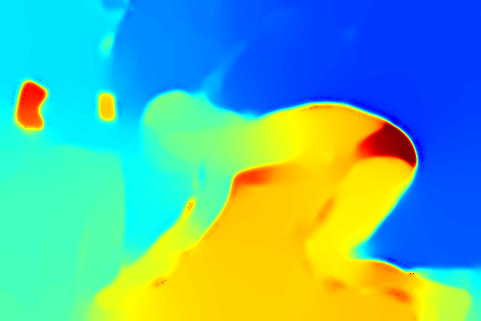}}\\
               \vspace{0.01\linewidth}
               \fbox{\includegraphics[width=1.00\linewidth]{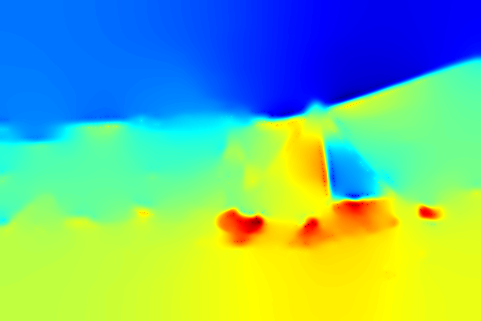}}\\
               \vspace{0.01\linewidth}
               \fbox{\includegraphics[width=1.00\linewidth]{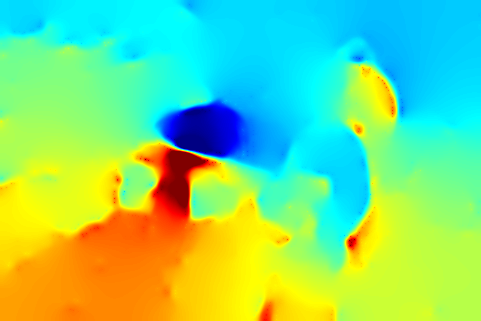}}\\
               \vspace{0.01\linewidth}
               \fbox{\includegraphics[width=1.00\linewidth]{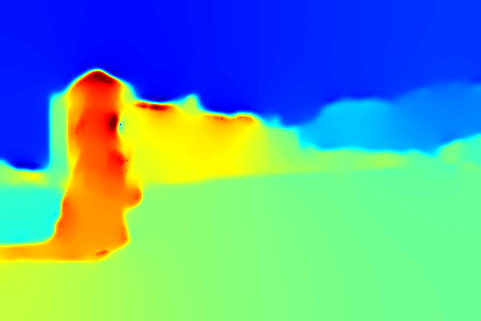}}\\
               \vspace{0.01\linewidth}
               \fbox{\includegraphics[width=1.00\linewidth]{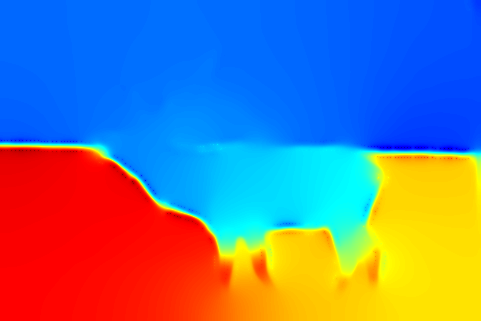}}\\
               \vspace{0.01\linewidth}
               \scriptsize{\textbf{\textsf{Spectral F/G}}}
            \end{center}
         \end{minipage}\\
         \vspace{1.0pt}
         {\rule{0.90\linewidth}{0.5pt}}\\
         \scriptsize{\textsf{\textbf{Maire}~\cite{Maire:ECCV:2010}}}
         \end{center}
      \end{minipage}
      \hfill
      \begin{minipage}[t]{0.495\linewidth}
         \vspace{0pt}
         \begin{center}
         \begin{minipage}[t]{0.32323\linewidth}
            \vspace{0pt}
            \begin{center}
               \fbox{\includegraphics[width=1.00\linewidth]{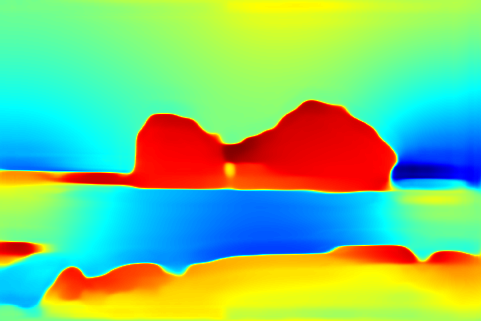}}\\
               \vspace{0.01\linewidth}
               \fbox{\includegraphics[width=1.00\linewidth]{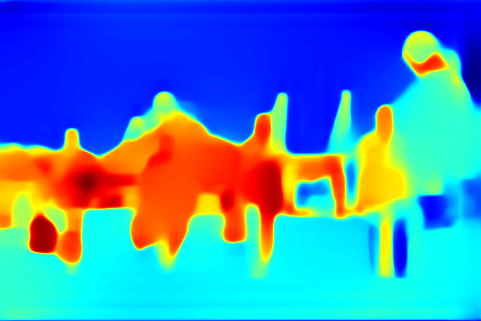}}\\
               \vspace{0.01\linewidth}
               \fbox{\includegraphics[width=1.00\linewidth]{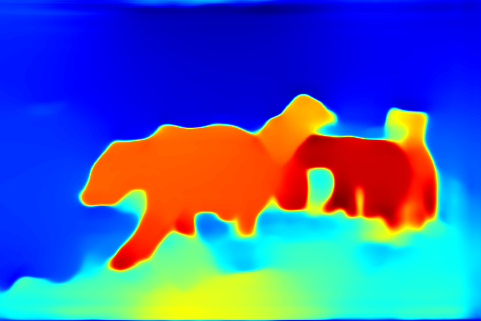}}\\
               \vspace{0.01\linewidth}
               \fbox{\includegraphics[width=1.00\linewidth]{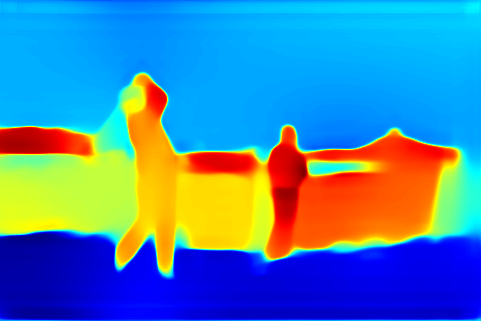}}\\
               \vspace{0.01\linewidth}
               \fbox{\includegraphics[width=1.00\linewidth]{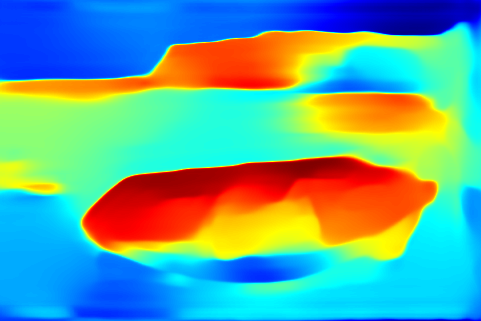}}\\
               \vspace{0.01\linewidth}
               \fbox{\includegraphics[width=1.00\linewidth]{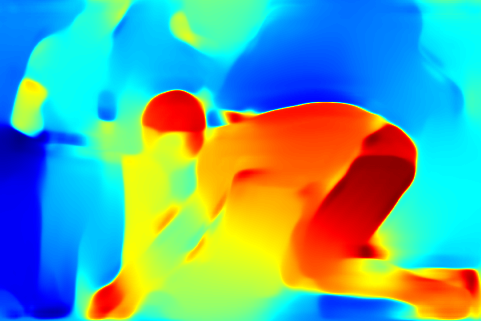}}\\
               \vspace{0.01\linewidth}
               \fbox{\includegraphics[width=1.00\linewidth]{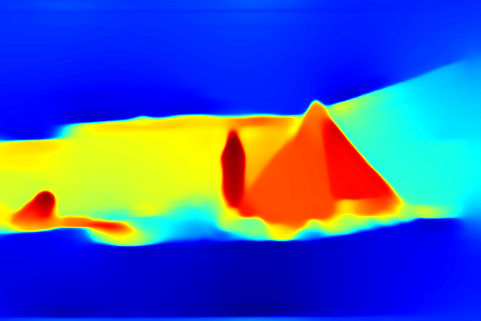}}\\
               \vspace{0.01\linewidth}
               \fbox{\includegraphics[width=1.00\linewidth]{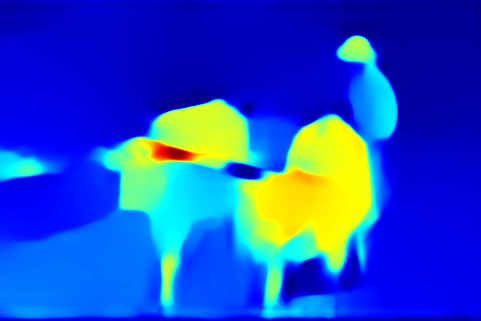}}\\
               \vspace{0.01\linewidth}
               \fbox{\includegraphics[width=1.00\linewidth]{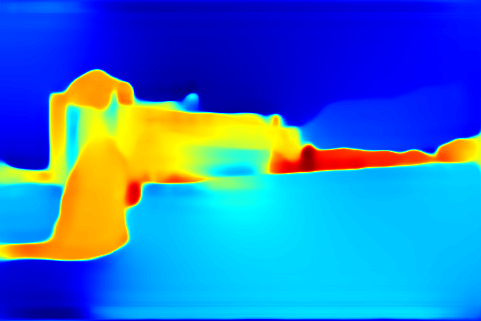}}\\
               \vspace{0.01\linewidth}
               \fbox{\includegraphics[width=1.00\linewidth]{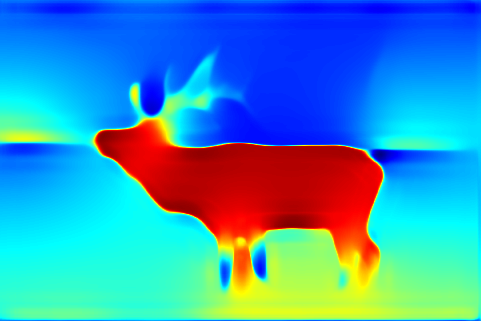}}\\
               \vspace{0.01\linewidth}
               \scriptsize{\textbf{\textsf{Spectral F/G}}}
            \end{center}
         \end{minipage}
         \begin{minipage}[t]{0.32323\linewidth}
            \vspace{0pt}
            \begin{center}
               \fbox{\includegraphics[width=1.00\linewidth]{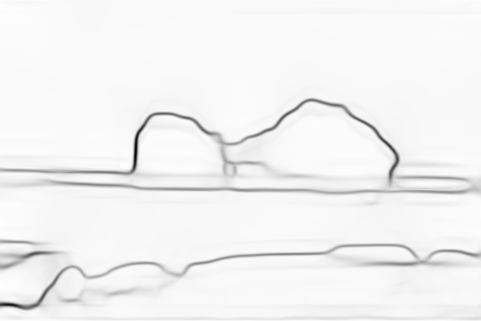}}\\
               \vspace{0.01\linewidth}
               \fbox{\includegraphics[width=1.00\linewidth]{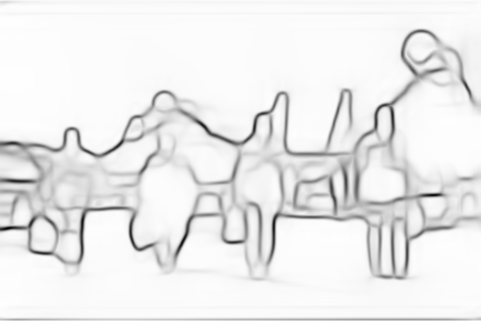}}\\
               \vspace{0.01\linewidth}
               \fbox{\includegraphics[width=1.00\linewidth]{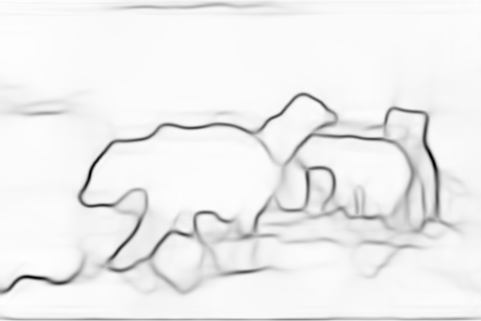}}\\
               \vspace{0.01\linewidth}
               \fbox{\includegraphics[width=1.00\linewidth]{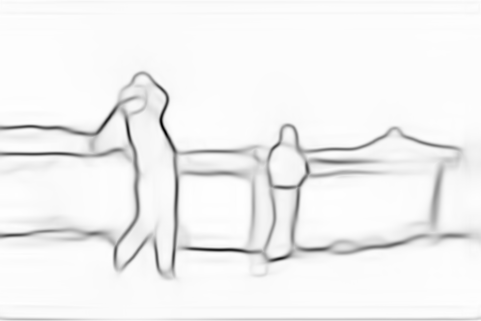}}\\
               \vspace{0.01\linewidth}
               \fbox{\includegraphics[width=1.00\linewidth]{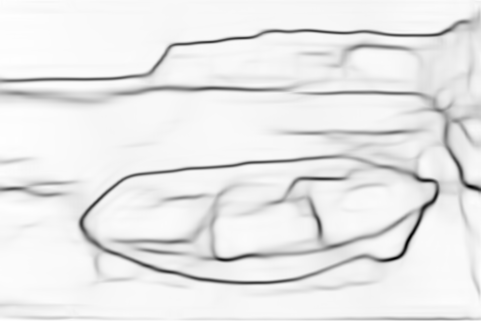}}\\
               \vspace{0.01\linewidth}
               \fbox{\includegraphics[width=1.00\linewidth]{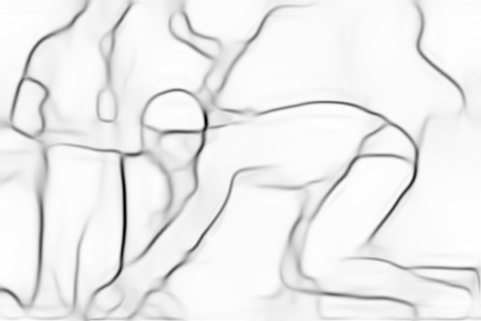}}\\
               \vspace{0.01\linewidth}
               \fbox{\includegraphics[width=1.00\linewidth]{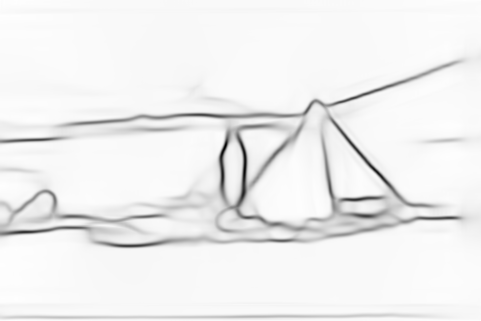}}\\
               \vspace{0.01\linewidth}
               \fbox{\includegraphics[width=1.00\linewidth]{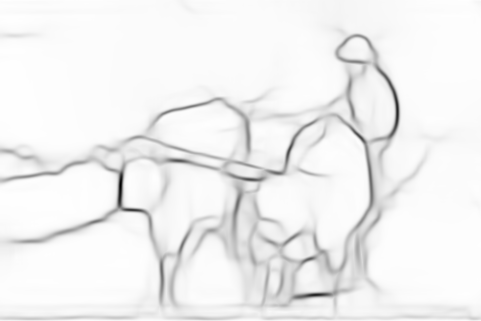}}\\
               \vspace{0.01\linewidth}
               \fbox{\includegraphics[width=1.00\linewidth]{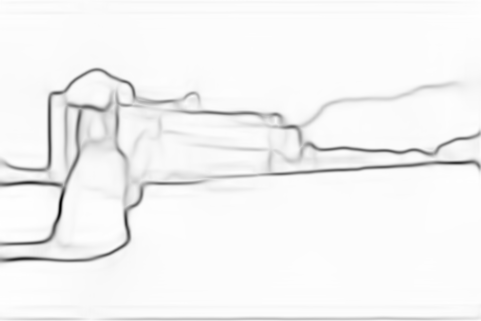}}\\
               \vspace{0.01\linewidth}
               \fbox{\includegraphics[width=1.00\linewidth]{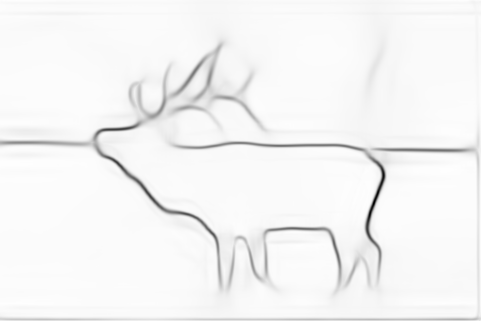}}\\
               \vspace{0.01\linewidth}
               \scriptsize{\textbf{\textsf{Spectral Boundaries}}}
            \end{center}
         \end{minipage}
         \begin{minipage}[t]{0.32323\linewidth}
            \vspace{0pt}
            \begin{center}
               \fbox{\includegraphics[width=1.00\linewidth]{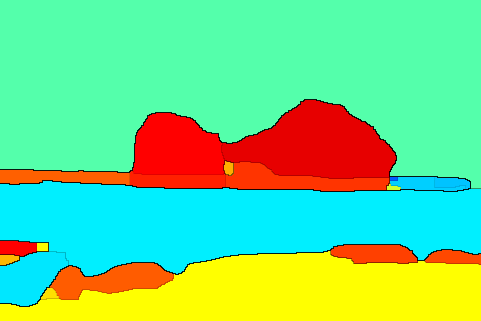}}\\
               \vspace{0.01\linewidth}
               \fbox{\includegraphics[width=1.00\linewidth]{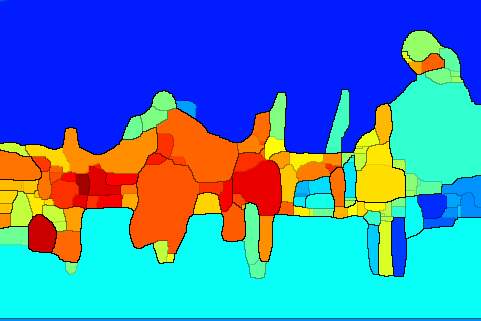}}\\
               \vspace{0.01\linewidth}
               \fbox{\includegraphics[width=1.00\linewidth]{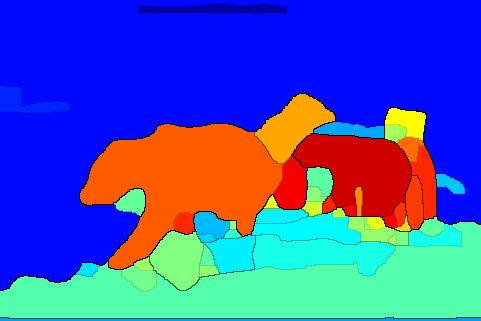}}\\
               \vspace{0.01\linewidth}
               \fbox{\includegraphics[width=1.00\linewidth]{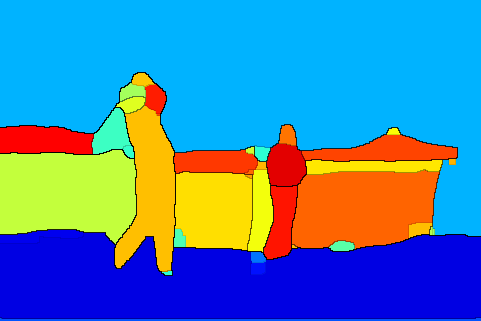}}\\
               \vspace{0.01\linewidth}
               \fbox{\includegraphics[width=1.00\linewidth]{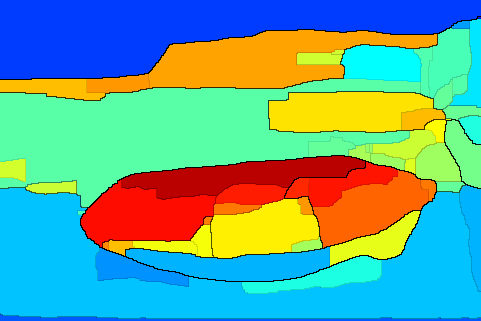}}\\
               \vspace{0.01\linewidth}
               \fbox{\includegraphics[width=1.00\linewidth]{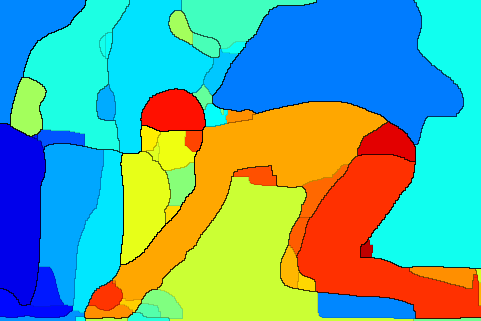}}\\
               \vspace{0.01\linewidth}
               \fbox{\includegraphics[width=1.00\linewidth]{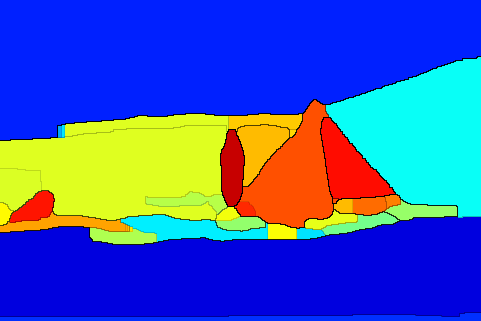}}\\
               \vspace{0.01\linewidth}
               \fbox{\includegraphics[width=1.00\linewidth]{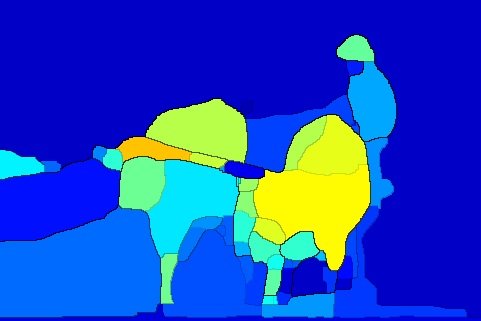}}\\
               \vspace{0.01\linewidth}
               \fbox{\includegraphics[width=1.00\linewidth]{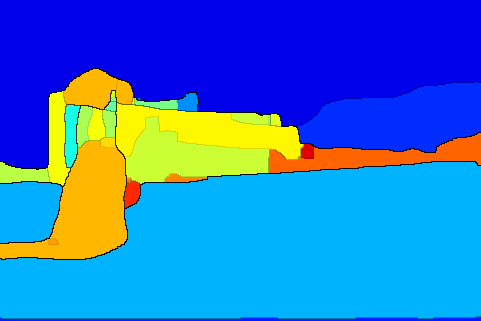}}\\
               \vspace{0.01\linewidth}
               \fbox{\includegraphics[width=1.00\linewidth]{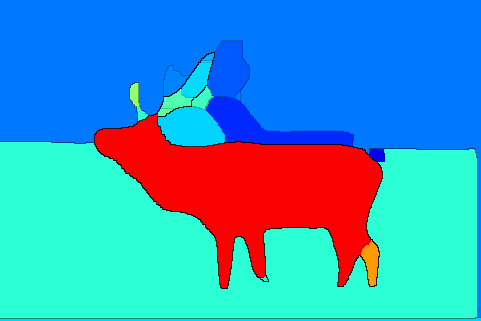}}\\
               \vspace{0.01\linewidth}
               \scriptsize{\textbf{\textsf{Segmentation + F/G}}}
            \end{center}
         \end{minipage}\\
         \vspace{1pt}
         \rule{0.975\linewidth}{0.5pt}\\
         \scriptsize{\textbf{\textsf{Our System}}}
         \end{center}
      \end{minipage}
   \end{center}
   \vspace{-0.01\linewidth}
   \caption{
      \textbf{Image segmentation and figure/ground results.}
         We compare our system to ground-truth and the results of
         Maire~\cite{Maire:ECCV:2010}.  Spectral F/G shows per-pixel
         figure/ground ordering according to the result of Angular Embedding.
         The colormap matches Figure~\ref{fig:ae}, with red denoting figure and
         blue denoting background.  Spectral boundaries show soft boundary
         strength encoded by the eigenvectors.  These boundaries generate a
         hierarchical segmentation~\cite{gPb-UCM}, one level of which we
         display in the final column with per-pixel figure/ground averaged
         over regions.  Note the drastic improvement in results
         over~\cite{Maire:ECCV:2010}.  While~\cite{Maire:ECCV:2010} reflects a
         strong lower-region bias for figure, our system learns to use image
         content and extracts foreground objects.  All examples are from our
         resplit figure/ground test subset of BSDS.
   }
   \label{fig:bsds-results}
\end{figure*}

\section{Experiments}
\label{sec:experiments}

Training our system for the generic perceptual task of segmentation and
figure/ground layering requires a dataset fully annotated in this form.
While there appears to be renewed interest in creating large-scale datasets
with such annotation~\cite{Amodal}, none has yet been released.  We therefore
use the Berkeley segmentation dataset~\cite{BSDS} for training.  Though it
consists of $500$ images total, only $200$ have been annotated with
ground-truth figure/ground~\cite{FMM:JOV:2007}.  We resplit this subset of
$200$ images into $150$ for training and $50$ for testing.

The following subsections detail, how, even with such scarcity of training
data, our system achieves substantial improvements in figure/ground quality
over prior work.

\subsection{BSDS: Ground-truth Figure/Ground}
\label{sec:bsds-gt-fg}

Our model formulation relies on dense labeling of pixel relationships.  The
BSDS ground-truth provides a dense segmentation in the from of a region map,
but only defines local figure/ground relationships between pixels immediately
adjacent along a region boundary~\cite{FMM:JOV:2007}.  We would like to train
predictors for long-range figure/ground relationships (our multiscale stencil
pattern) in addition to short-range.

Figure~\ref{fig:bsds-training} illustrates our method for overcoming this
limitation.  Given perfect (\eg ground-truth) short-range predictions as
input, Angular Embedding generates an extremely high-quality global
figure/ground estimate.  In a real setting, we want robustness by having many
estimates of pairwise relations over many scales.  Ground-truth short-range
connections suffice as they are perfect estimates.  We use the globalized
ground-truth figure/ground map (column $4$ in Figure~\ref{fig:bsds-training})
as our training signal $R$ in Equation~\ref{eq:f_train}.  The usual
ground-truth segmentation serves as $S$ in Equation~\ref{eq:b_train}.

\subsection{BSDS: Segmentation \& Figure/Ground Results}
\label{sec:bsds-seg-fg}

Figure~\ref{fig:bsds-results} shows results on some examples from our $50$
image test set.  Compared to the previous attempt~\cite{Maire:ECCV:2010} to
use Angular Embedding as an inference engine for figure/ground, our results
are strikingly better.  It is visually apparent that our system improves on
every single example in terms of figure/ground.

Our segmentation, as measured by boundary quality, is comparable to that of
similar systems using spectral clustering for segmentation
alone~\cite{gPb-UCM}.  On the standard boundary precision recall benchmark
on BSDS, our spectral boundaries achieve an F-measure of $0.68$, identical to
that of the spectral component (``spectral Pb'') of the gPb boundary
detector~\cite{gPb-UCM}.  Thus, as a segmentation engine our system is on par
with the previous best spectral-clustering based systems.

As a system for joint segmentation and figure/ground organization, our system
has few competitors.  Use of Angular Embedding to solve both problems at once
is unique to our system, and~\cite{Maire:ECCV:2010,MYP:ICCV:2011}.
Figure~\ref{fig:bsds-results} shows an obvious huge jump in figure/ground
performance over~\cite{Maire:ECCV:2010}.

\subsection{BSDS: Figure/Ground Benchmark}
\label{sec:bsds-fg-bench}

To our knowledge, there is not a well-established benchmarking methodology
for dense figure/ground predictions.  While \cite{VS:ECCV:2014} propose
metrics coupling figure/ground classification accuracy along boundaries to
boundary detection performance, we develop a simpler alternative.

Given a per-pixel figure/ground ordering assignment, and a segmentation
partitioning an image into regions, we can easily order the regions according
to figure/ground layering.  Simply assign each region a rank order equal to the
mean figure/ground order of its member pixels.  For robustness to minor
misalignment between the figure/ground assignment and the boundaries of regions
in the segmentation, we use median in place of mean.

This transfer procedure serves as a basis for comparing different
figure/ground orderings.  We transfer them both onto the same segmentation.
In particular, given predicted figure/ground ordering $\theta(\cdot)$,
ground-truth figure/ground ordering $\tilde{\theta}(\cdot)$, and ground-truth
segmentation $S$, we transfer each of $\theta(\cdot)$ and
$\tilde{\theta}(\cdot)$ onto $S$.  This gives two orderings of the regions in
$S$, which we compare according to the following metrics:
\begin{itemize}
   \item{
      Pairwise region accuracy (R-ACC):  For each pair of neighboring regions
      in $S$, if the ground-truth figure/ground assignment shows them to be
      in different layers, we test whether the predicted relative ordering of
      these regions matches the ground-truth relative ordering.  That is, we
      measure accuracy on the classification problem of predicting which
      region is in front.
   }
   \item{
      Boundary ownership accuracy (B-ACC):  We define the front region as
      owning the pixels on the common boundary of the region pair and measure
      the per-pixel accuracy of predicting boundary ownership.  This is a
      reweighting of R-ACC.  In R-ACC, all region pairs straddling a
      ground-truth figure/ground boundary count equally.  In B-ACC, their
      importance is weighted according to length of the boundary.
   }
   \item{
      Boundary ownership of foreground regions (B-ACC-50, B-ACC-25):
      Identical to B-ACC, except we only consider boundaries which belong to
      the foreground-most $50\%$ or $25\%$ of regions in the ground-truth
      figure/ground ordering of each image.  These metrics emphasize
      the importance of correct predictions for foreground objects while
      ignoring more distant objects.
   }
\end{itemize}

Note that $S$ need not be the ground-truth segmentation.  We can project
ground-truth figure/ground onto any segmentation (say, a machine-generated
one) and compare to predicted figure/ground projected onto that segmentation.

\begin{table}
   \begin{center}
   \begin{footnotesize}
   \begin{tabular}{l|c|c|c|c}
      \multicolumn{1}{c|}{Segmentation:} & \multicolumn{4}{c}{{Figure/Ground Prediction Accuracy}}\\
      \multicolumn{1}{c|}{Ground-truth}
                                         & R-ACC           & B-ACC           & B-ACC-50        & B-ACC-25\\
      \hline
      F/G: Ours                          & $\mathbf{0.62}$ & $\mathbf{0.69}$ & $\mathbf{0.72}$ & $\mathbf{0.73}$\\
      F/G: Maire~\cite{Maire:ECCV:2010}  & $0.56$          & $0.58$          & $0.56$          & $0.56$\\
      \multicolumn{5}{c}{~}\\
      \multicolumn{1}{c|}{Segmentation:} & \multicolumn{4}{c}{{Figure/Ground Prediction Accuracy}}\\
      \multicolumn{1}{c|}{Ours}
                                         & R-ACC           & B-ACC           & B-ACC-50        & B-ACC-25\\
      \hline
      F/G: Ours                          & $\mathbf{0.66}$ & $\mathbf{0.70}$ & $\mathbf{0.69}$ & $\mathbf{0.67}$\\
      F/G: Maire~\cite{Maire:ECCV:2010}  & $0.59$          & $0.62$          & $0.61$          & $0.58$\\
   \end{tabular}
   \end{footnotesize}
   \end{center}
   \caption{
      \textbf{Figure/ground benchmark results.}
         After transferring figure/ground predictions onto either ground-truth
         (upper table) or our own (lower table) segmentations, we quantify
         accuracy of local relative relationships.  R-ACC is pairwise
         region accuracy: considering all pairs of neighboring regions, what
         fraction are correctly ordered by relative figure/ground?  B-ACC is
         boundary ownership accuracy: what fraction of boundary pixels have
         correct figure ownership assigned?  B-ACC-50 and B-ACC-25 restrict
         measurement to the boundaries of the $50\%$ and $25\%$ most foreground
         regions (a proxy for foreground objects).  Our system dramatically
         outperforms~\cite{Maire:ECCV:2010} across all metrics.
   }
   \label{tab:benchmarks}
\end{table}

\begin{figure*}
   \setlength\fboxsep{0pt}
   \begin{center}
      \begin{minipage}[t]{0.165\linewidth}
         \vspace{0pt}
         \begin{center}
         \begin{minipage}[t]{0.96970\linewidth}
            \vspace{0pt}
            \begin{center}
               \fbox{\includegraphics[width=1.00\linewidth]{figs/results_bsds_images_249061_sm.png}}\\
               \vspace{0.01\linewidth}
               \fbox{\includegraphics[width=1.00\linewidth]{figs/results_bsds_images_245051_sm.png}}\\
               \vspace{0.01\linewidth}
               \fbox{\includegraphics[width=1.00\linewidth]{figs/results_bsds_images_94079_sm.png}}\\
               \vspace{0.01\linewidth}
               \fbox{\includegraphics[width=1.00\linewidth]{figs/results_bsds_images_286092_sm.png}}\\
               \vspace{0.01\linewidth}
               \fbox{\includegraphics[width=1.00\linewidth]{figs/results_bsds_images_118020_sm.png}}\\
               \vspace{0.01\linewidth}
               \fbox{\includegraphics[width=1.00\linewidth]{figs/results_bsds_images_153077_sm.png}}\\
               \vspace{0.01\linewidth}
               \fbox{\includegraphics[width=1.00\linewidth]{figs/results_bsds_images_188063_sm.png}}\\
               \vspace{0.01\linewidth}
               \fbox{\includegraphics[width=1.00\linewidth]{figs/results_bsds_images_202012_sm.png}}\\
               \vspace{0.01\linewidth}
               \fbox{\includegraphics[width=1.00\linewidth]{figs/results_bsds_images_216053_sm.png}}\\
               \vspace{0.01\linewidth}
               \fbox{\includegraphics[width=1.00\linewidth]{figs/results_bsds_images_41004_sm.png}}\\
               \vspace{0.01\linewidth}
               \scriptsize{\textbf{\textsf{Image}}}
            \end{center}
         \end{minipage}
         \end{center}
      \end{minipage}
      \hfill
      \begin{minipage}[t]{0.495\linewidth}
         \vspace{0pt}
         \begin{center}
         \begin{minipage}[t]{0.32323\linewidth}
            \vspace{0pt}
            \begin{center}
               \fbox{\includegraphics[width=1.00\linewidth]{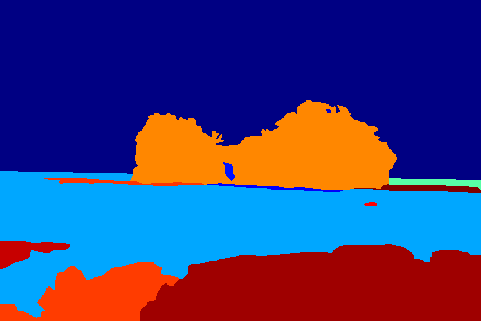}}\\
               \vspace{0.01\linewidth}
               \fbox{\includegraphics[width=1.00\linewidth]{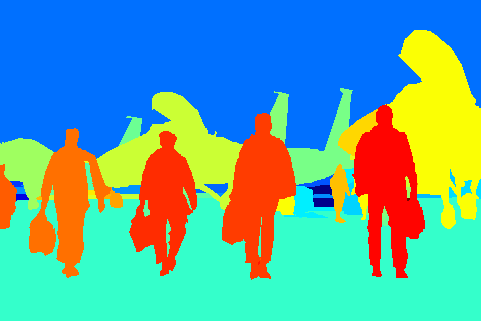}}\\
               \vspace{0.01\linewidth}
               \fbox{\includegraphics[width=1.00\linewidth]{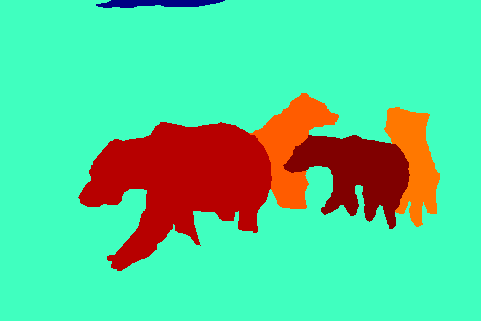}}\\
               \vspace{0.01\linewidth}
               \fbox{\includegraphics[width=1.00\linewidth]{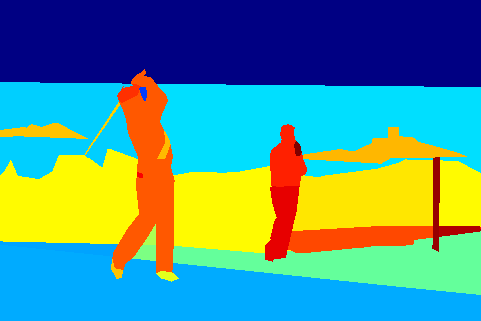}}\\
               \vspace{0.01\linewidth}
               \fbox{\includegraphics[width=1.00\linewidth]{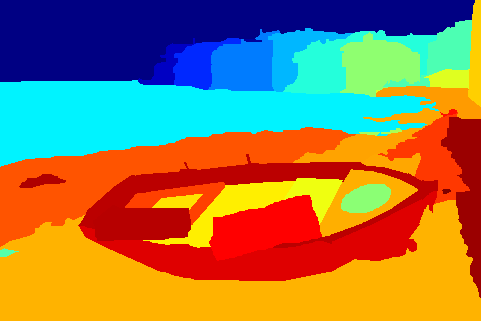}}\\
               \vspace{0.01\linewidth}
               \fbox{\includegraphics[width=1.00\linewidth]{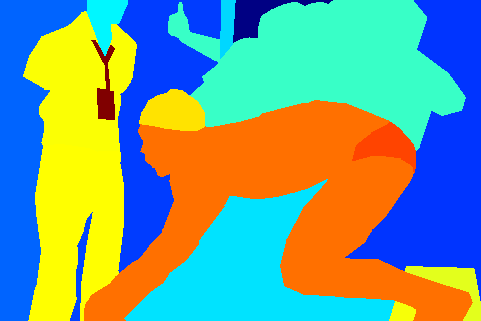}}\\
               \vspace{0.01\linewidth}
               \fbox{\includegraphics[width=1.00\linewidth]{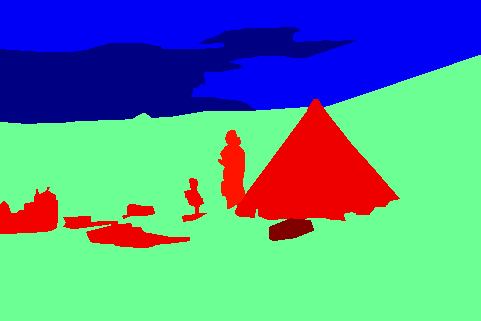}}\\
               \vspace{0.01\linewidth}
               \fbox{\includegraphics[width=1.00\linewidth]{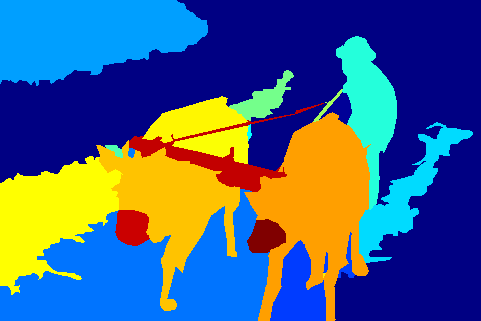}}\\
               \vspace{0.01\linewidth}
               \fbox{\includegraphics[width=1.00\linewidth]{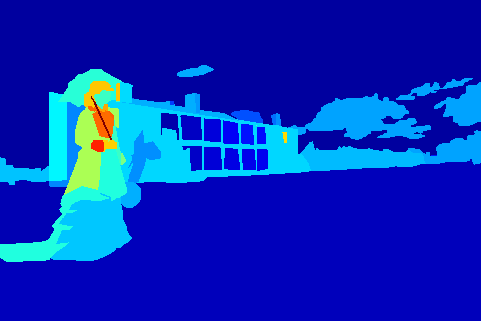}}\\
               \vspace{0.01\linewidth}
               \fbox{\includegraphics[width=1.00\linewidth]{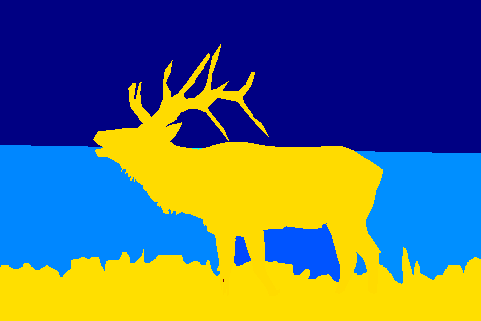}}\\
               \vspace{0.01\linewidth}
               \scriptsize{\textbf{\textsf{F/G: Ground-truth}}}
            \end{center}
         \end{minipage}
         \begin{minipage}[t]{0.32323\linewidth}
            \vspace{0pt}
            \begin{center}
               \fbox{\includegraphics[width=1.00\linewidth]{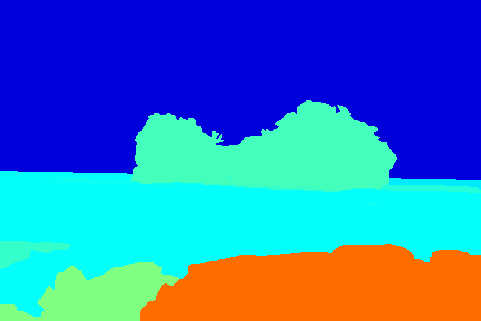}}\\
               \vspace{0.01\linewidth}
               \fbox{\includegraphics[width=1.00\linewidth]{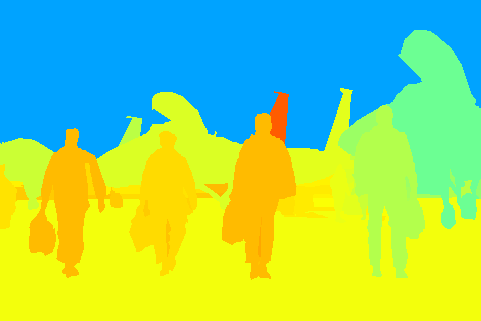}}\\
               \vspace{0.01\linewidth}
               \fbox{\includegraphics[width=1.00\linewidth]{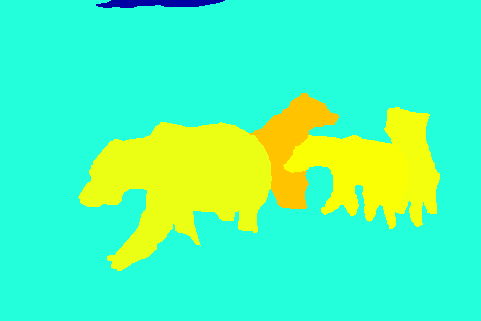}}\\
               \vspace{0.01\linewidth}
               \fbox{\includegraphics[width=1.00\linewidth]{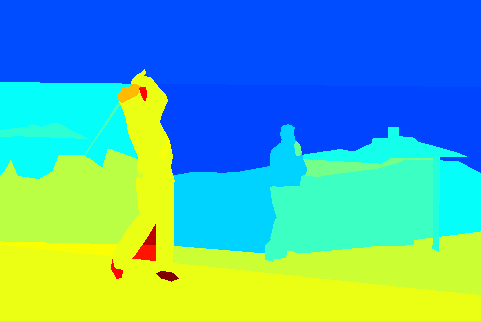}}\\
               \vspace{0.01\linewidth}
               \fbox{\includegraphics[width=1.00\linewidth]{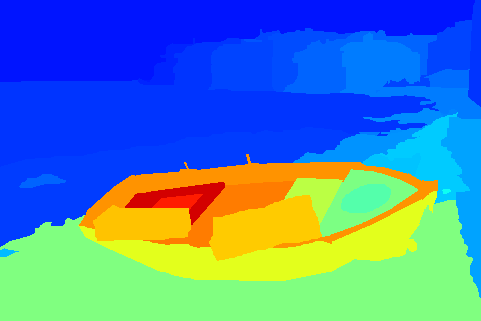}}\\
               \vspace{0.01\linewidth}
               \fbox{\includegraphics[width=1.00\linewidth]{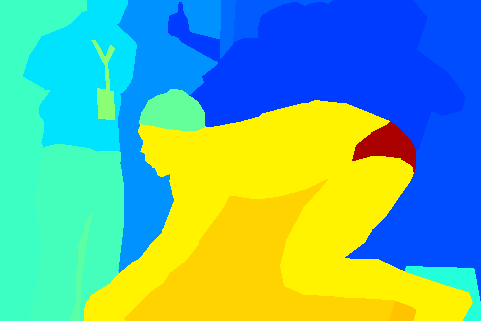}}\\
               \vspace{0.01\linewidth}
               \fbox{\includegraphics[width=1.00\linewidth]{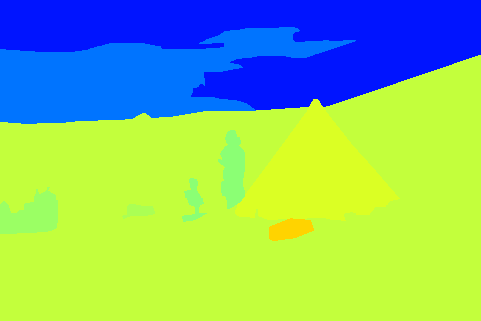}}\\
               \vspace{0.01\linewidth}
               \fbox{\includegraphics[width=1.00\linewidth]{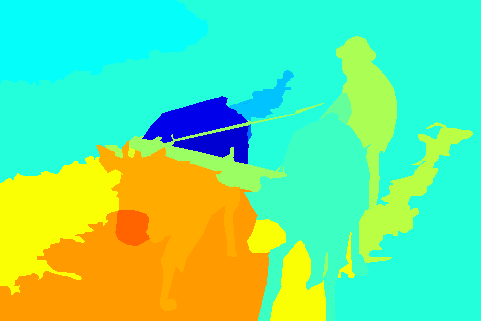}}\\
               \vspace{0.01\linewidth}
               \fbox{\includegraphics[width=1.00\linewidth]{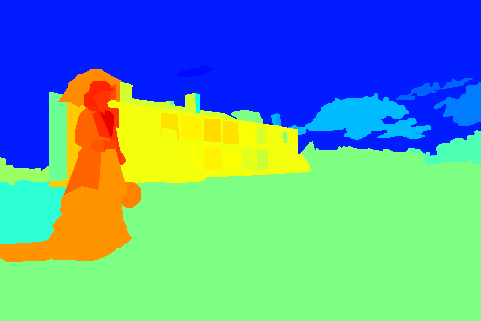}}\\
               \vspace{0.01\linewidth}
               \fbox{\includegraphics[width=1.00\linewidth]{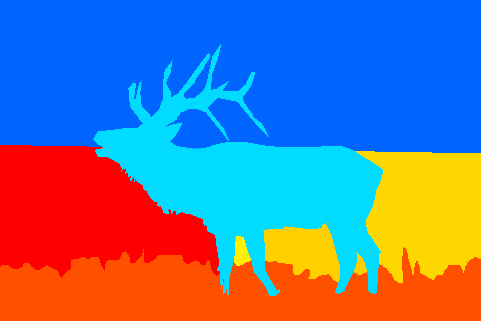}}\\
               \vspace{0.01\linewidth}
               \scriptsize{\textsf{\textbf{F/G: Maire}~\cite{Maire:ECCV:2010}}}
            \end{center}
         \end{minipage}
         \begin{minipage}[t]{0.32323\linewidth}
            \vspace{0pt}
            \begin{center}
               \fbox{\includegraphics[width=1.00\linewidth]{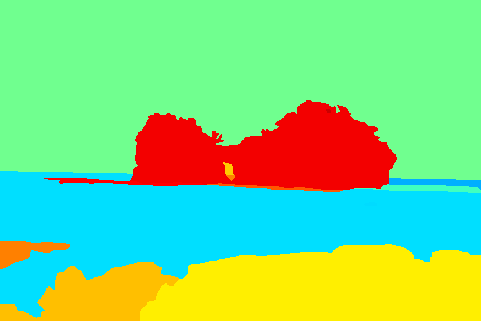}}\\
               \vspace{0.01\linewidth}
               \fbox{\includegraphics[width=1.00\linewidth]{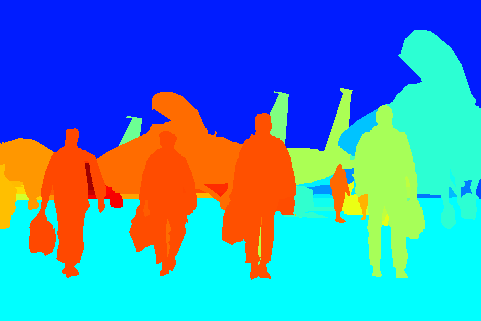}}\\
               \vspace{0.01\linewidth}
               \fbox{\includegraphics[width=1.00\linewidth]{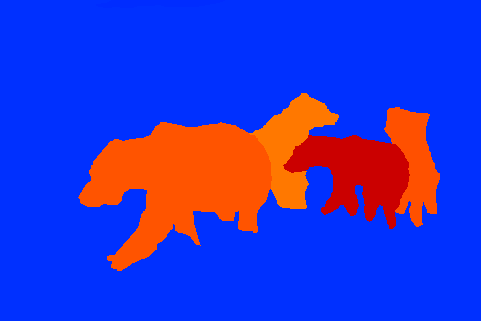}}\\
               \vspace{0.01\linewidth}
               \fbox{\includegraphics[width=1.00\linewidth]{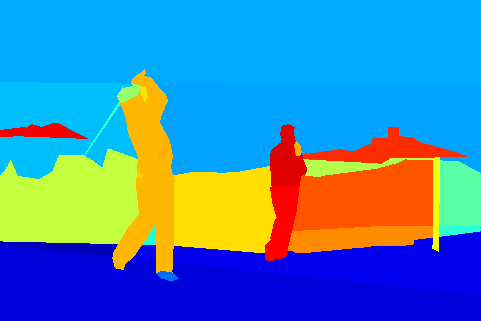}}\\
               \vspace{0.01\linewidth}
               \fbox{\includegraphics[width=1.00\linewidth]{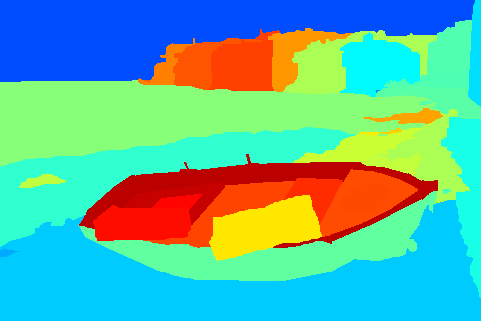}}\\
               \vspace{0.01\linewidth}
               \fbox{\includegraphics[width=1.00\linewidth]{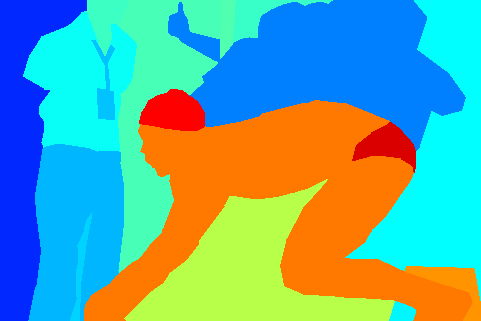}}\\
               \vspace{0.01\linewidth}
               \fbox{\includegraphics[width=1.00\linewidth]{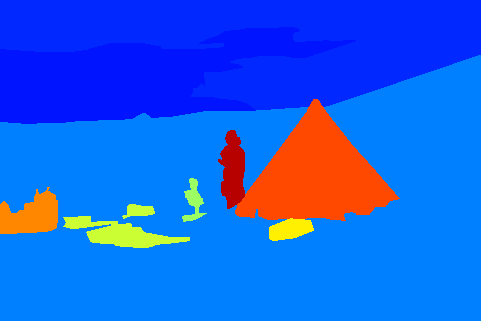}}\\
               \vspace{0.01\linewidth}
               \fbox{\includegraphics[width=1.00\linewidth]{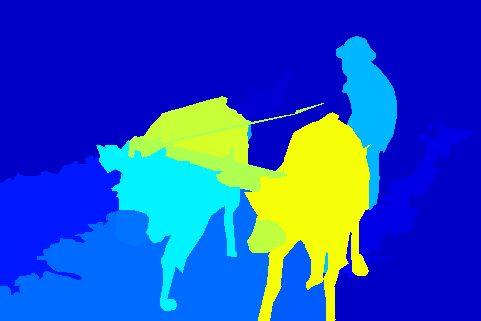}}\\
               \vspace{0.01\linewidth}
               \fbox{\includegraphics[width=1.00\linewidth]{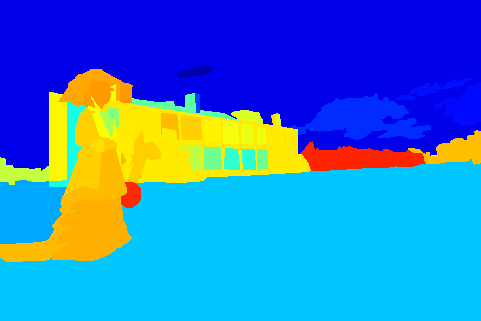}}\\
               \vspace{0.01\linewidth}
               \fbox{\includegraphics[width=1.00\linewidth]{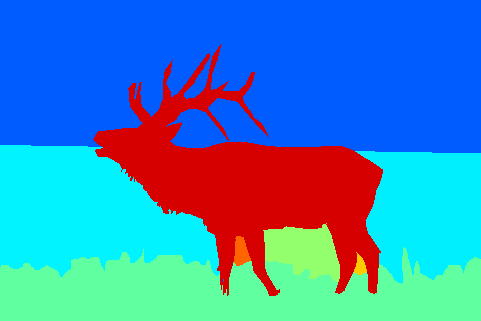}}\\
               \vspace{0.01\linewidth}
               \scriptsize{\textbf{\textsf{F/G: Ours}}}
            \end{center}
         \end{minipage}\\
         \vspace{1pt}
         \rule{0.975\linewidth}{0.5pt}\\
         \scriptsize{\textbf{\textsf{Figure/Ground Transferred onto Ground-truth Segmentation}}}
         \end{center}
      \end{minipage}
      \hfill
      \begin{minipage}[t]{0.33\linewidth}
         \vspace{0pt}
         \begin{center}
         \begin{minipage}[t]{0.48485\linewidth}
            \vspace{0pt}
            \begin{center}
               \fbox{\includegraphics[width=1.00\linewidth]{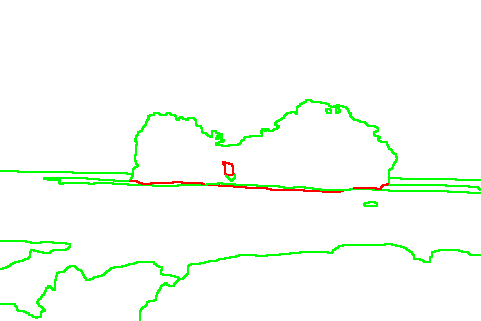}}\\
               \vspace{0.01\linewidth}
               \fbox{\includegraphics[width=1.00\linewidth]{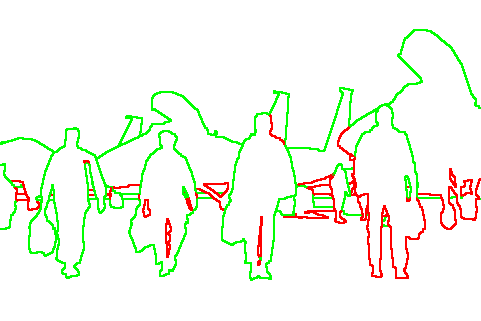}}\\
               \vspace{0.01\linewidth}
               \fbox{\includegraphics[width=1.00\linewidth]{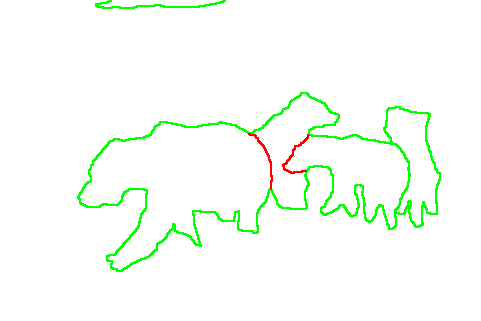}}\\
               \vspace{0.01\linewidth}
               \fbox{\includegraphics[width=1.00\linewidth]{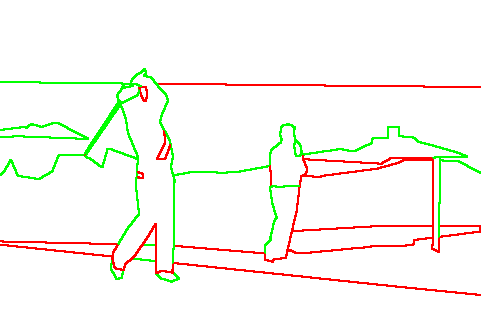}}\\
               \vspace{0.01\linewidth}
               \fbox{\includegraphics[width=1.00\linewidth]{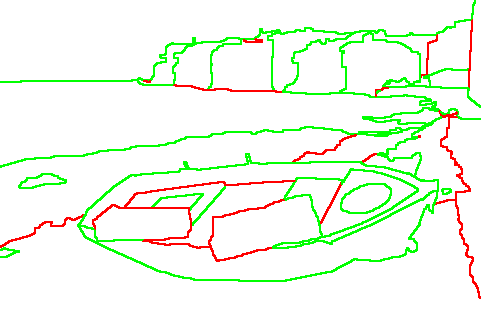}}\\
               \vspace{0.01\linewidth}
               \fbox{\includegraphics[width=1.00\linewidth]{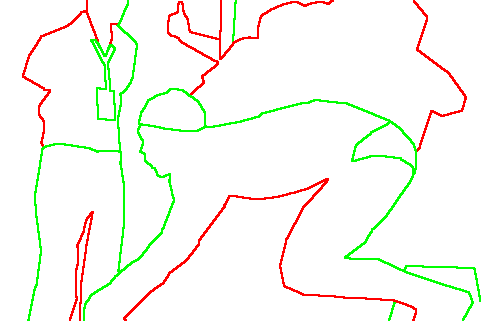}}\\
               \vspace{0.01\linewidth}
               \fbox{\includegraphics[width=1.00\linewidth]{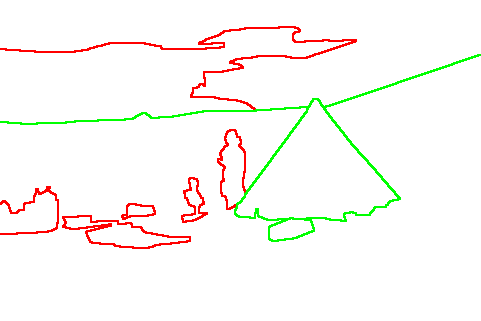}}\\
               \vspace{0.01\linewidth}
               \fbox{\includegraphics[width=1.00\linewidth]{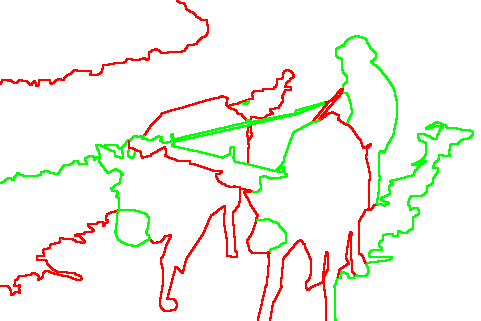}}\\
               \vspace{0.01\linewidth}
               \fbox{\includegraphics[width=1.00\linewidth]{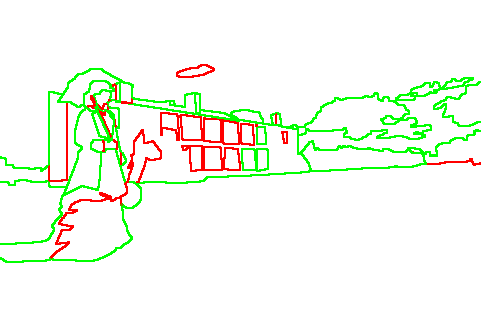}}\\
               \vspace{0.01\linewidth}
               \fbox{\includegraphics[width=1.00\linewidth]{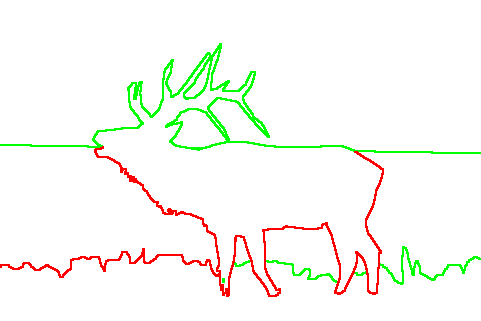}}\\
               \vspace{0.01\linewidth}
               \scriptsize{\textsf{\textbf{Maire}~\cite{Maire:ECCV:2010}}}
            \end{center}
         \end{minipage}
         \begin{minipage}[t]{0.48485\linewidth}
            \vspace{0pt}
            \begin{center}
               \fbox{\includegraphics[width=1.00\linewidth]{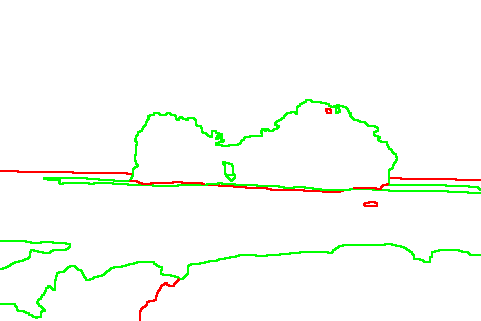}}\\
               \vspace{0.01\linewidth}
               \fbox{\includegraphics[width=1.00\linewidth]{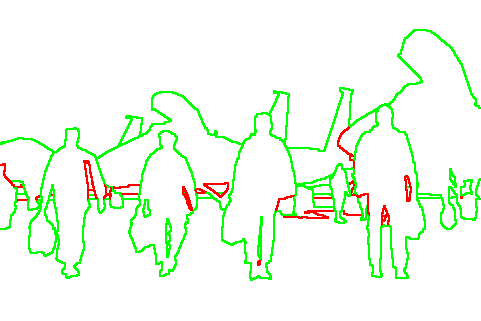}}\\
               \vspace{0.01\linewidth}
               \fbox{\includegraphics[width=1.00\linewidth]{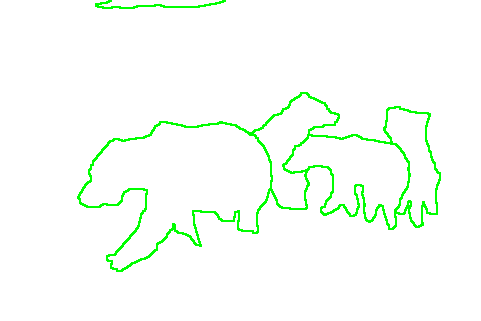}}\\
               \vspace{0.01\linewidth}
               \fbox{\includegraphics[width=1.00\linewidth]{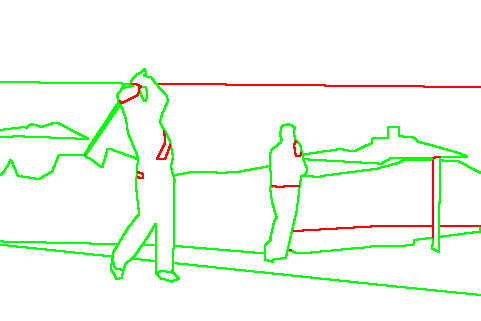}}\\
               \vspace{0.01\linewidth}
               \fbox{\includegraphics[width=1.00\linewidth]{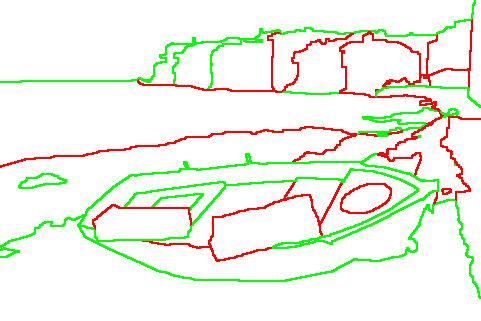}}\\
               \vspace{0.01\linewidth}
               \fbox{\includegraphics[width=1.00\linewidth]{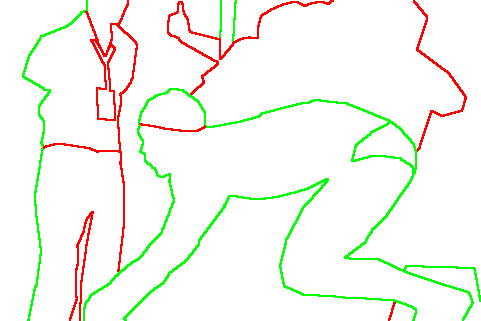}}\\
               \vspace{0.01\linewidth}
               \fbox{\includegraphics[width=1.00\linewidth]{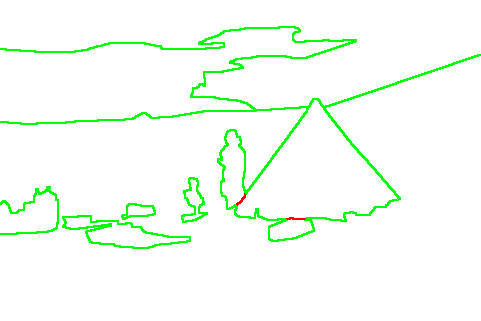}}\\
               \vspace{0.01\linewidth}
               \fbox{\includegraphics[width=1.00\linewidth]{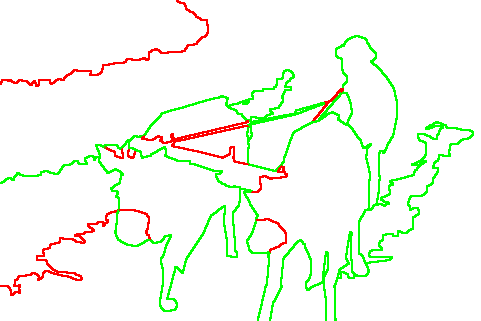}}\\
               \vspace{0.01\linewidth}
               \fbox{\includegraphics[width=1.00\linewidth]{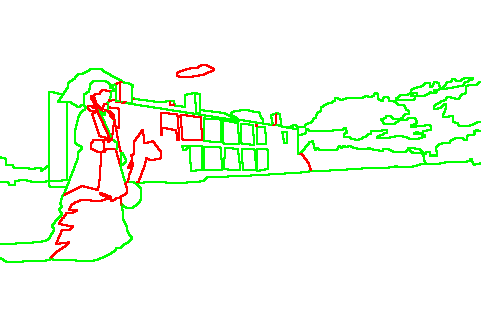}}\\
               \vspace{0.01\linewidth}
               \fbox{\includegraphics[width=1.00\linewidth]{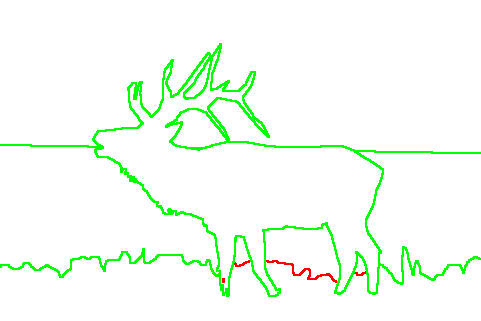}}\\
               \vspace{0.01\linewidth}
               \scriptsize{\textbf{\textsf{Ours}}}
            \end{center}
         \end{minipage}\\
         \vspace{1.0pt}
         {\rule{0.90\linewidth}{0.5pt}}\\
         \scriptsize{\textbf{\textsf{Boundary Ownership Correctness}}}
         \end{center}
      \end{minipage}
   \end{center}
   \vspace{-0.01\linewidth}
   \caption{
      \textbf{Figure/ground predictions measured on ground-truth segmentation.}
         We transfer per-pixel figure/ground predictions (columns 2~through~4
         of Figure~\ref{fig:bsds-results}) onto the ground-truth segmentation by
         taking the median value over each region.  For boundaries separating
         regions with different ground-truth figure/ground layer assignments,
         we check whether the predicted owner (more figural region) matches
         the owner according to the ground-truth.  The rightmost two columns
         mark correct boundary ownership predictions in green and errors in
         red for both the results of Maire's system~\cite{Maire:ECCV:2010}
         and our system.  Note how we correctly predict ownership of object
         lower boundaries (rows 6, 8, 10) and improve on small objects
         (row 7).  Table~\ref{tab:benchmarks} gives quantitative benchmarks.
   }
   \label{fig:fg-gt}
\end{figure*}

\begin{figure*}
   \setlength\fboxsep{0pt}
   \begin{center}
      \begin{minipage}[t]{0.165\linewidth}
         \vspace{0pt}
         \begin{center}
         \begin{minipage}[t]{0.96970\linewidth}
            \vspace{0pt}
            \begin{center}
               \fbox{\includegraphics[width=1.00\linewidth]{figs/results_bsds_images_249061_sm.png}}\\
               \vspace{0.01\linewidth}
               \fbox{\includegraphics[width=1.00\linewidth]{figs/results_bsds_images_245051_sm.png}}\\
               \vspace{0.01\linewidth}
               \fbox{\includegraphics[width=1.00\linewidth]{figs/results_bsds_images_94079_sm.png}}\\
               \vspace{0.01\linewidth}
               \fbox{\includegraphics[width=1.00\linewidth]{figs/results_bsds_images_286092_sm.png}}\\
               \vspace{0.01\linewidth}
               \fbox{\includegraphics[width=1.00\linewidth]{figs/results_bsds_images_118020_sm.png}}\\
               \vspace{0.01\linewidth}
               \fbox{\includegraphics[width=1.00\linewidth]{figs/results_bsds_images_153077_sm.png}}\\
               \vspace{0.01\linewidth}
               \fbox{\includegraphics[width=1.00\linewidth]{figs/results_bsds_images_188063_sm.png}}\\
               \vspace{0.01\linewidth}
               \fbox{\includegraphics[width=1.00\linewidth]{figs/results_bsds_images_202012_sm.png}}\\
               \vspace{0.01\linewidth}
               \fbox{\includegraphics[width=1.00\linewidth]{figs/results_bsds_images_216053_sm.png}}\\
               \vspace{0.01\linewidth}
               \fbox{\includegraphics[width=1.00\linewidth]{figs/results_bsds_images_41004_sm.png}}\\
               \vspace{0.01\linewidth}
               \scriptsize{\textbf{\textsf{Image}}}
            \end{center}
         \end{minipage}
         \end{center}
      \end{minipage}
      \hfill
      \begin{minipage}[t]{0.495\linewidth}
         \vspace{0pt}
         \begin{center}
         \begin{minipage}[t]{0.32323\linewidth}
            \vspace{0pt}
            \begin{center}
               \fbox{\includegraphics[width=1.00\linewidth]{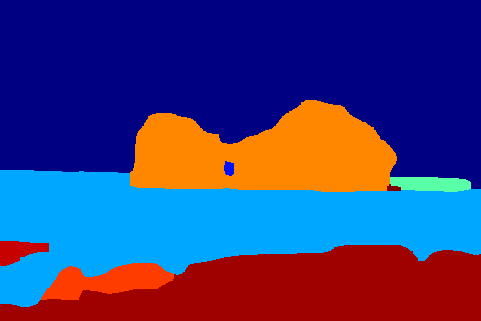}}\\
               \vspace{0.01\linewidth}
               \fbox{\includegraphics[width=1.00\linewidth]{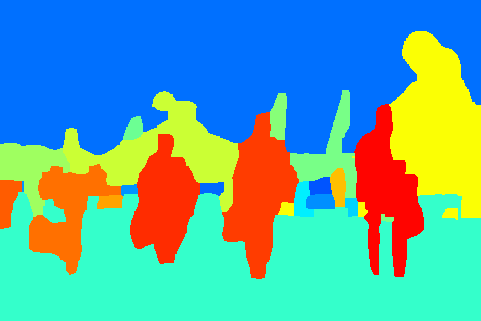}}\\
               \vspace{0.01\linewidth}
               \fbox{\includegraphics[width=1.00\linewidth]{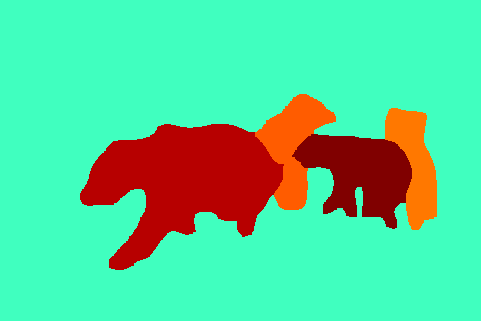}}\\
               \vspace{0.01\linewidth}
               \fbox{\includegraphics[width=1.00\linewidth]{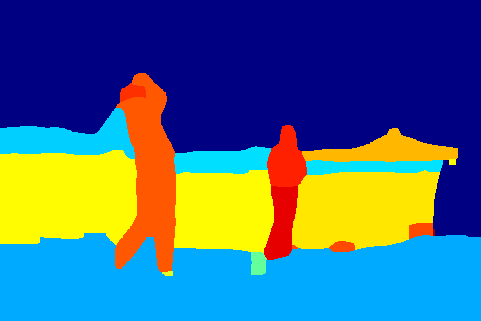}}\\
               \vspace{0.01\linewidth}
               \fbox{\includegraphics[width=1.00\linewidth]{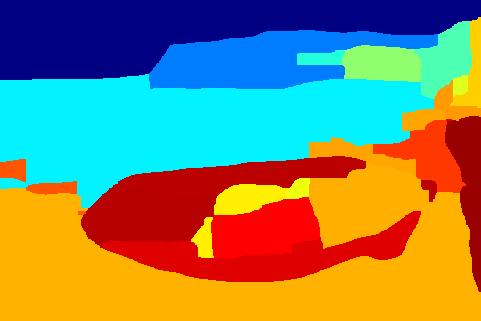}}\\
               \vspace{0.01\linewidth}
               \fbox{\includegraphics[width=1.00\linewidth]{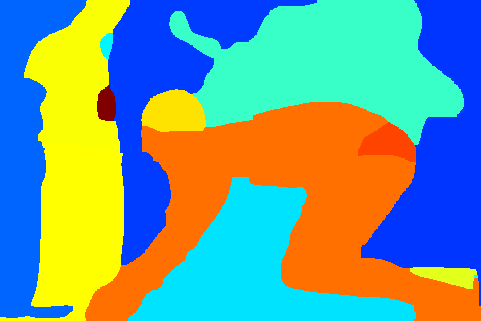}}\\
               \vspace{0.01\linewidth}
               \fbox{\includegraphics[width=1.00\linewidth]{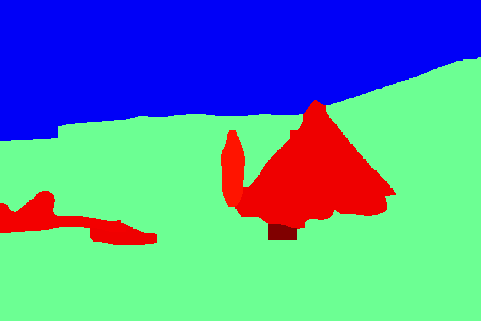}}\\
               \vspace{0.01\linewidth}
               \fbox{\includegraphics[width=1.00\linewidth]{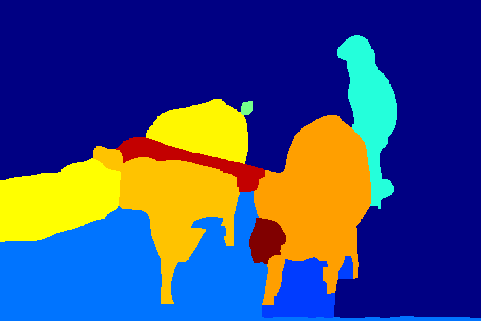}}\\
               \vspace{0.01\linewidth}
               \fbox{\includegraphics[width=1.00\linewidth]{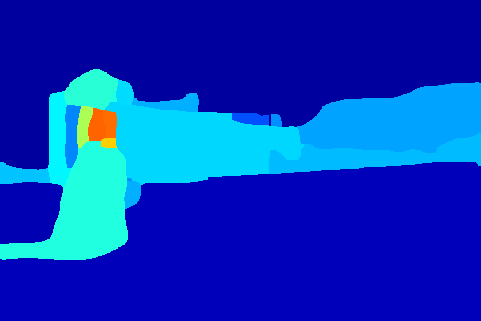}}\\
               \vspace{0.01\linewidth}
               \fbox{\includegraphics[width=1.00\linewidth]{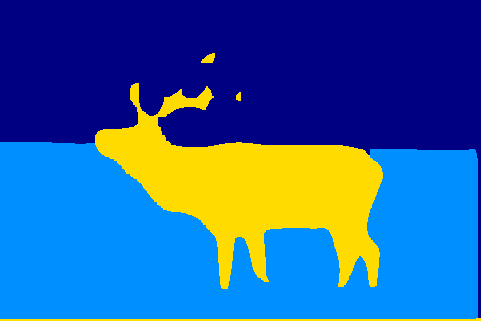}}\\
               \vspace{0.01\linewidth}
               \scriptsize{\textbf{\textsf{F/G: Ground-truth}}}
            \end{center}
         \end{minipage}
         \begin{minipage}[t]{0.32323\linewidth}
            \vspace{0pt}
            \begin{center}
               \fbox{\includegraphics[width=1.00\linewidth]{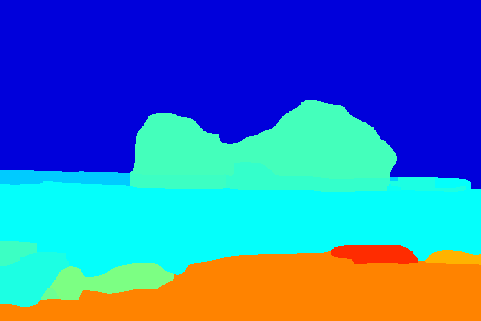}}\\
               \vspace{0.01\linewidth}
               \fbox{\includegraphics[width=1.00\linewidth]{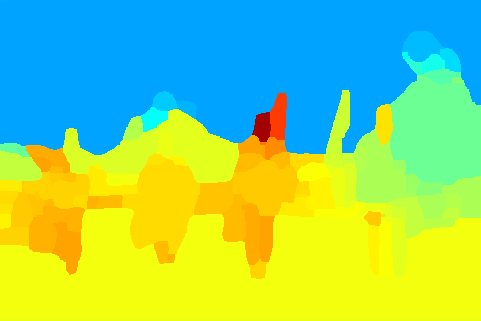}}\\
               \vspace{0.01\linewidth}
               \fbox{\includegraphics[width=1.00\linewidth]{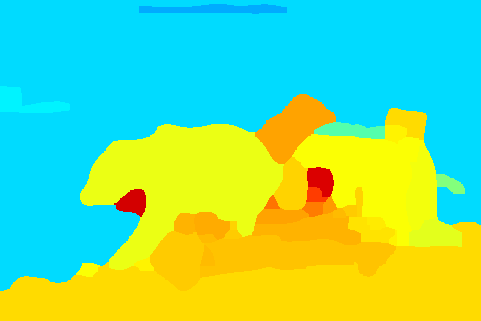}}\\
               \vspace{0.01\linewidth}
               \fbox{\includegraphics[width=1.00\linewidth]{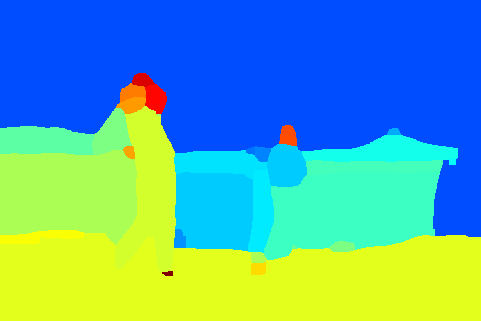}}\\
               \vspace{0.01\linewidth}
               \fbox{\includegraphics[width=1.00\linewidth]{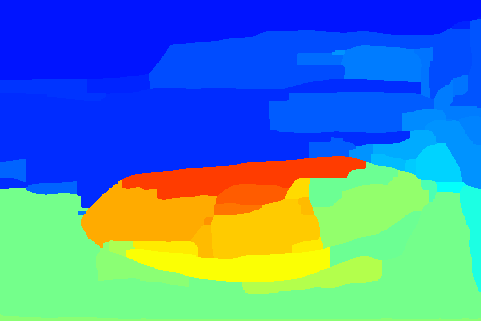}}\\
               \vspace{0.01\linewidth}
               \fbox{\includegraphics[width=1.00\linewidth]{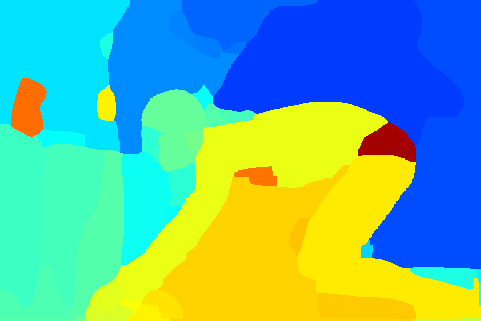}}\\
               \vspace{0.01\linewidth}
               \fbox{\includegraphics[width=1.00\linewidth]{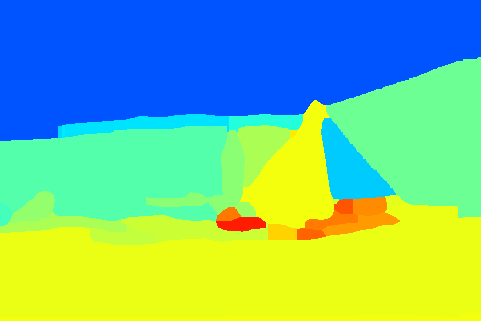}}\\
               \vspace{0.01\linewidth}
               \fbox{\includegraphics[width=1.00\linewidth]{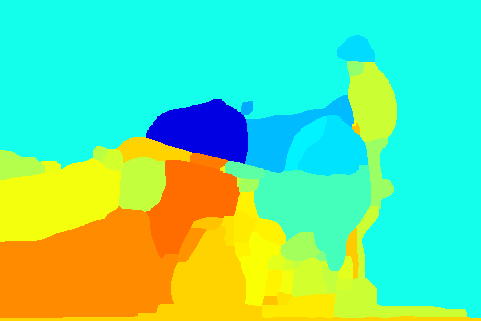}}\\
               \vspace{0.01\linewidth}
               \fbox{\includegraphics[width=1.00\linewidth]{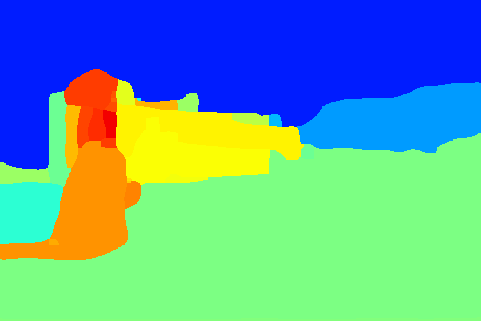}}\\
               \vspace{0.01\linewidth}
               \fbox{\includegraphics[width=1.00\linewidth]{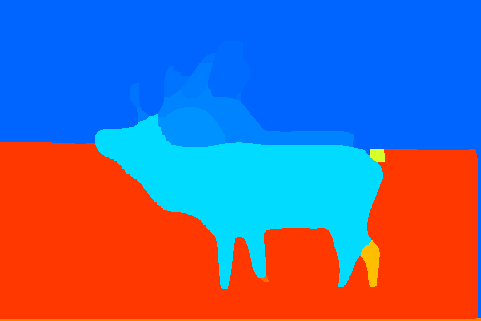}}\\
               \vspace{0.01\linewidth}
               \scriptsize{\textsf{\textbf{F/G: Maire}~\cite{Maire:ECCV:2010}}}
            \end{center}
         \end{minipage}
         \begin{minipage}[t]{0.32323\linewidth}
            \vspace{0pt}
            \begin{center}
               \fbox{\includegraphics[width=1.00\linewidth]{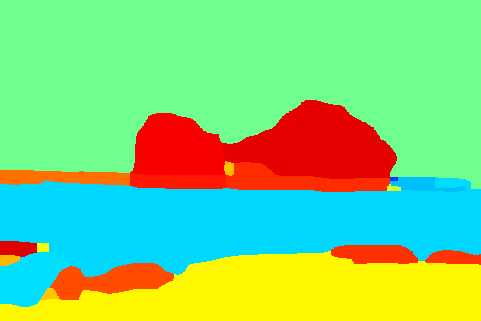}}\\
               \vspace{0.01\linewidth}
               \fbox{\includegraphics[width=1.00\linewidth]{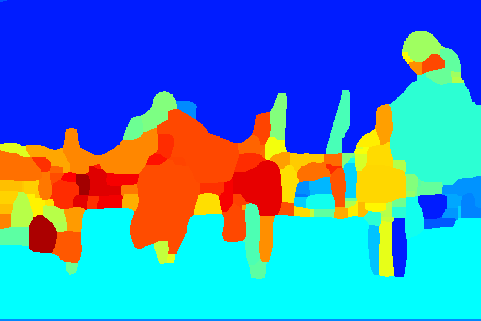}}\\
               \vspace{0.01\linewidth}
               \fbox{\includegraphics[width=1.00\linewidth]{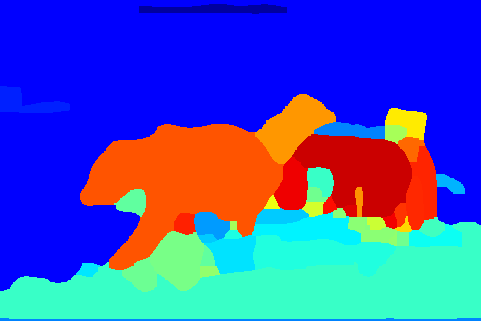}}\\
               \vspace{0.01\linewidth}
               \fbox{\includegraphics[width=1.00\linewidth]{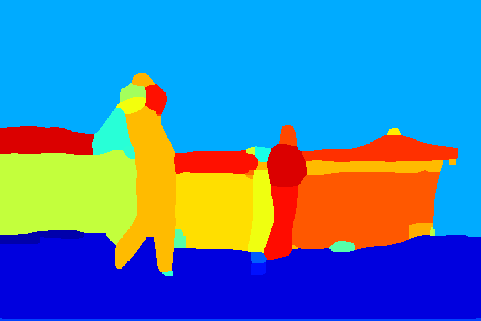}}\\
               \vspace{0.01\linewidth}
               \fbox{\includegraphics[width=1.00\linewidth]{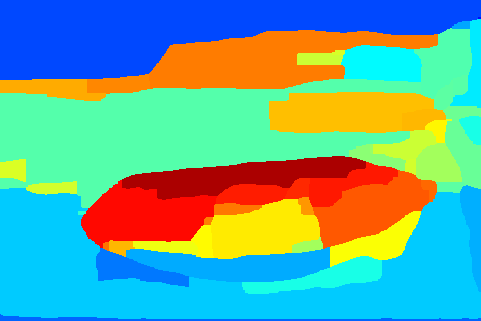}}\\
               \vspace{0.01\linewidth}
               \fbox{\includegraphics[width=1.00\linewidth]{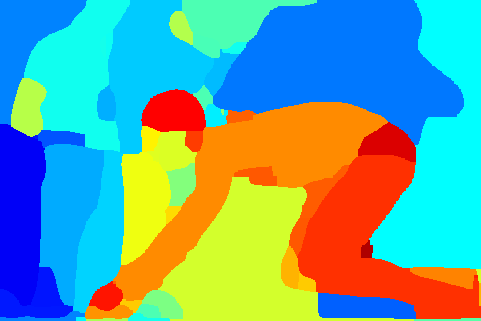}}\\
               \vspace{0.01\linewidth}
               \fbox{\includegraphics[width=1.00\linewidth]{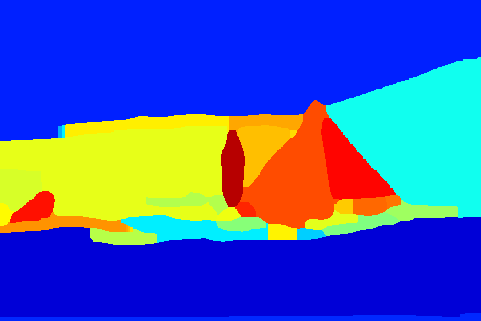}}\\
               \vspace{0.01\linewidth}
               \fbox{\includegraphics[width=1.00\linewidth]{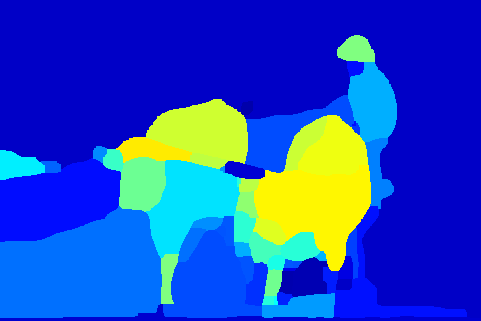}}\\
               \vspace{0.01\linewidth}
               \fbox{\includegraphics[width=1.00\linewidth]{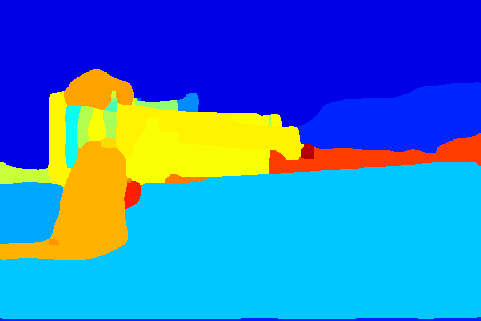}}\\
               \vspace{0.01\linewidth}
               \fbox{\includegraphics[width=1.00\linewidth]{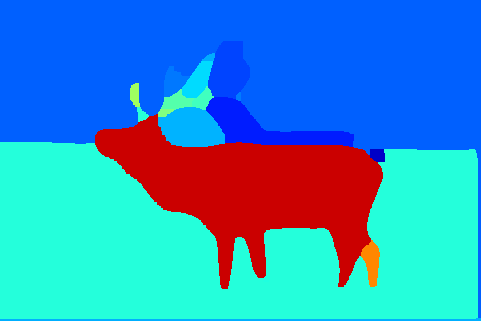}}\\
               \vspace{0.01\linewidth}
               \scriptsize{\textbf{\textsf{F/G: Ours}}}
            \end{center}
         \end{minipage}\\
         \vspace{1pt}
         \rule{0.975\linewidth}{0.5pt}\\
         \scriptsize{\textbf{\textsf{Figure/Ground Transferred onto Our Segmentation}}}
         \end{center}
      \end{minipage}
      \hfill
      \begin{minipage}[t]{0.33\linewidth}
         \vspace{0pt}
         \begin{center}
         \begin{minipage}[t]{0.48485\linewidth}
            \vspace{0pt}
            \begin{center}
               \fbox{\includegraphics[width=1.00\linewidth]{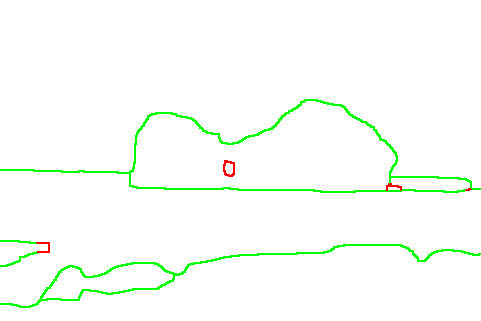}}\\
               \vspace{0.01\linewidth}
               \fbox{\includegraphics[width=1.00\linewidth]{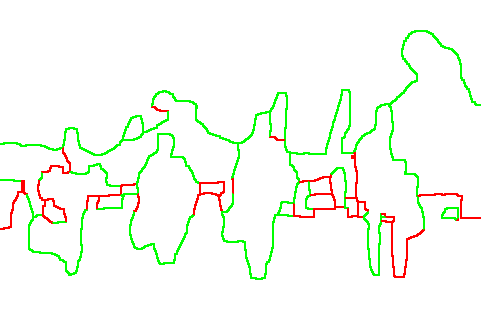}}\\
               \vspace{0.01\linewidth}
               \fbox{\includegraphics[width=1.00\linewidth]{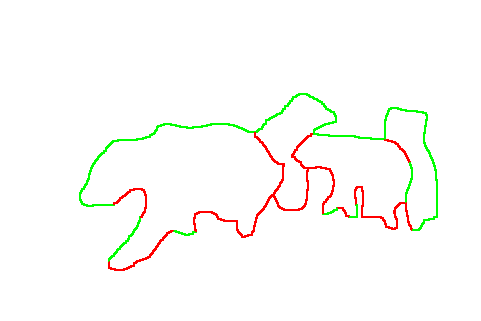}}\\
               \vspace{0.01\linewidth}
               \fbox{\includegraphics[width=1.00\linewidth]{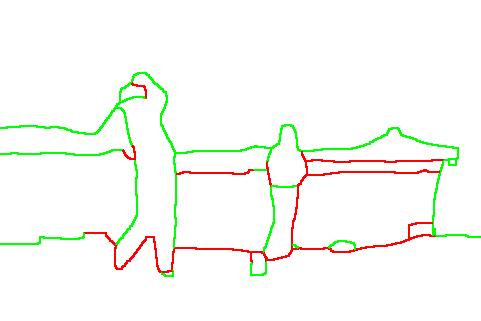}}\\
               \vspace{0.01\linewidth}
               \fbox{\includegraphics[width=1.00\linewidth]{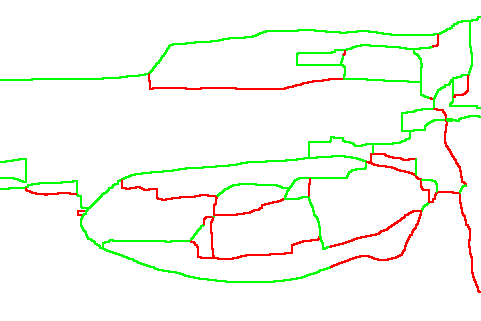}}\\
               \vspace{0.01\linewidth}
               \fbox{\includegraphics[width=1.00\linewidth]{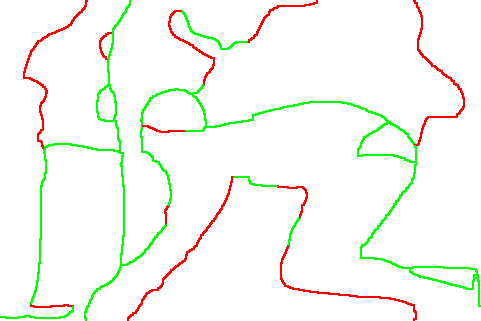}}\\
               \vspace{0.01\linewidth}
               \fbox{\includegraphics[width=1.00\linewidth]{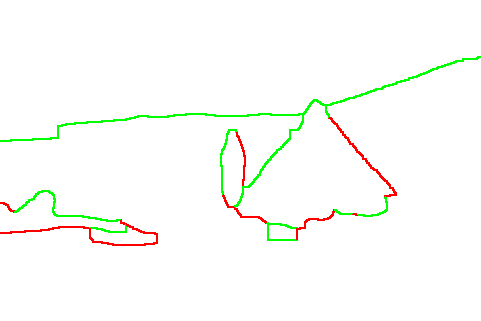}}\\
               \vspace{0.01\linewidth}
               \fbox{\includegraphics[width=1.00\linewidth]{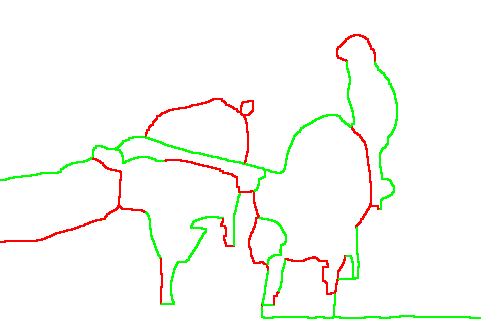}}\\
               \vspace{0.01\linewidth}
               \fbox{\includegraphics[width=1.00\linewidth]{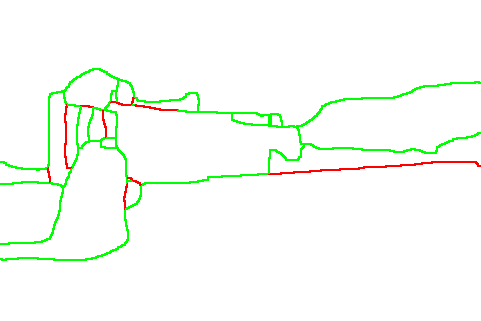}}\\
               \vspace{0.01\linewidth}
               \fbox{\includegraphics[width=1.00\linewidth]{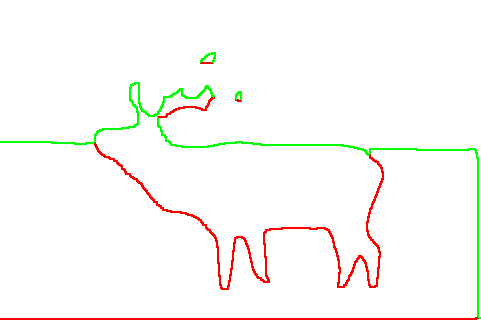}}\\
               \vspace{0.01\linewidth}
               \scriptsize{\textsf{\textbf{Maire}~\cite{Maire:ECCV:2010}}}
            \end{center}
         \end{minipage}
         \begin{minipage}[t]{0.48485\linewidth}
            \vspace{0pt}
            \begin{center}
               \fbox{\includegraphics[width=1.00\linewidth]{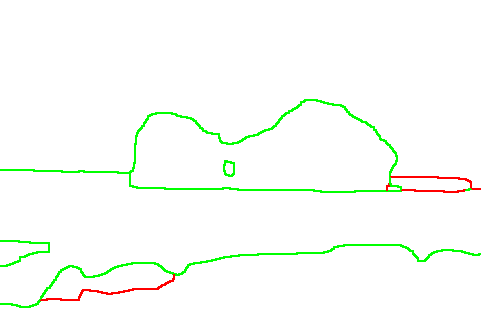}}\\
               \vspace{0.01\linewidth}
               \fbox{\includegraphics[width=1.00\linewidth]{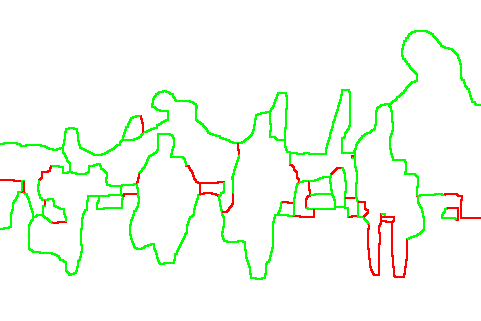}}\\
               \vspace{0.01\linewidth}
               \fbox{\includegraphics[width=1.00\linewidth]{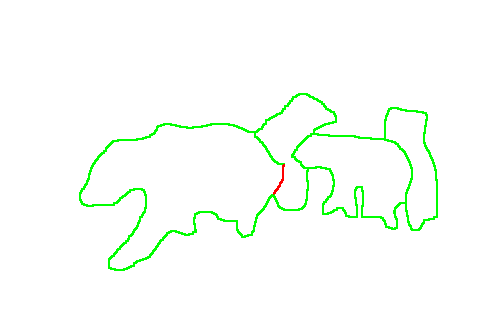}}\\
               \vspace{0.01\linewidth}
               \fbox{\includegraphics[width=1.00\linewidth]{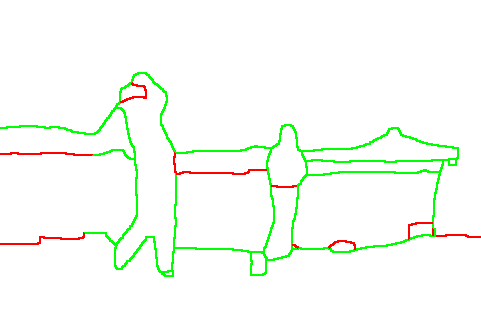}}\\
               \vspace{0.01\linewidth}
               \fbox{\includegraphics[width=1.00\linewidth]{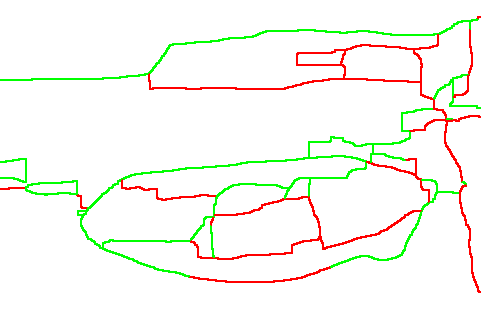}}\\
               \vspace{0.01\linewidth}
               \fbox{\includegraphics[width=1.00\linewidth]{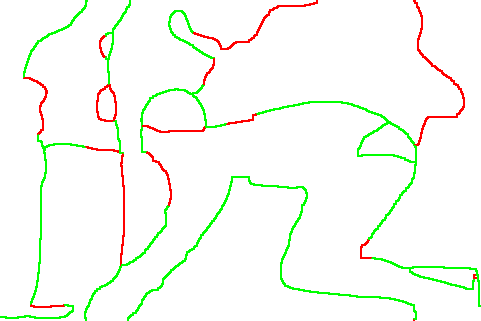}}\\
               \vspace{0.01\linewidth}
               \fbox{\includegraphics[width=1.00\linewidth]{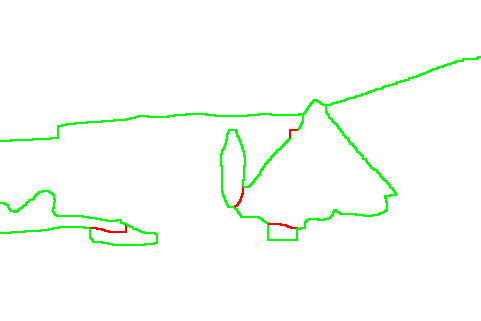}}\\
               \vspace{0.01\linewidth}
               \fbox{\includegraphics[width=1.00\linewidth]{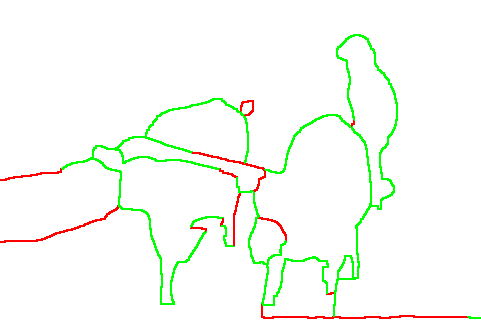}}\\
               \vspace{0.01\linewidth}
               \fbox{\includegraphics[width=1.00\linewidth]{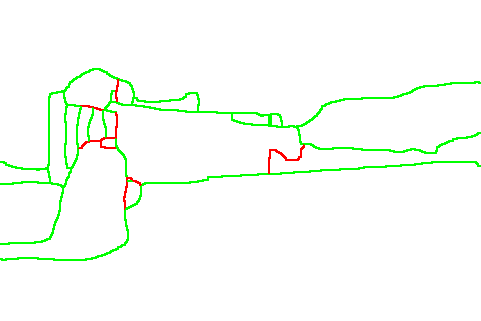}}\\
               \vspace{0.01\linewidth}
               \fbox{\includegraphics[width=1.00\linewidth]{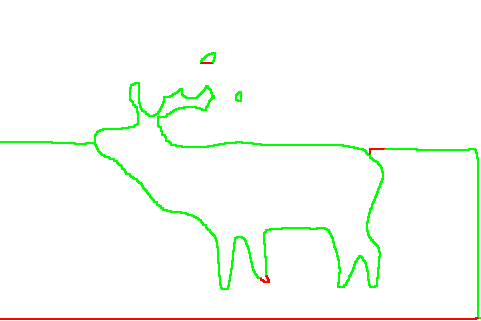}}\\
               \vspace{0.01\linewidth}
               \scriptsize{\textbf{\textsf{Ours}}}
            \end{center}
         \end{minipage}\\
         \vspace{1.0pt}
         {\rule{0.90\linewidth}{0.5pt}}\\
         \scriptsize{\textbf{\textsf{Boundary Ownership Correctness}}}
         \end{center}
      \end{minipage}
   \end{center}
   \vspace{-0.01\linewidth}
   \caption{
      \textbf{Figure/ground predictions measured on our segmentation.}
         As in Figure~\ref{fig:fg-gt}, we transfer ground-truth figure/ground,
         Maire's figure/ground predictions~\cite{Maire:ECCV:2010}, and our
         predictions onto a common segmentation.  However, instead of using
         the ground-truth segmentation, we transfer onto the segmentation
         generated by our system.  The ground-truth figure/ground transferred
         onto our regions defines the boundary ownership signal against which
         we judge predictions.  Comparing with Figure~\ref{fig:fg-gt}, our
         boundary ownership predictions are mostly consistent regardless of
         the segmentation (ground-truth or ours) to which they are applied.
         However, row 3 shows this is not the case for~\cite{Maire:ECCV:2010};
         here, its predicted correct ownership for the lower boundary relies
         on averaging out over a large ground-truth background region.
   }
   \label{fig:fg-us}
\end{figure*}

Table~\ref{tab:benchmarks} reports a complete comparison of our figure/ground
predictions and those of~\cite{Maire:ECCV:2010} against ground-truth
figure/ground on our $50$ image test subset of BSDS~\cite{BSDS}.  We
consider both projection onto ground-truth segmentation and onto our own
system's segmentation output.  For the latter, as our system produces
hierarchical segmentation, we use the region partition at a fixed level of the
hierarchy, calibrated for optimal boundary F-measure.  Figures~\ref{fig:fg-gt}
and~\ref{fig:fg-us} provide visual comparison on $10$ test images.

Across all metrics, our system significantly outperforms~\cite{
Maire:ECCV:2010}.  We achieve $69\%$ and $70\%$ boundary ownership accuracy on
ground-truth and automatic segmentation, respectively, compared to $58\%$ and
$62\%$ for~\cite{Maire:ECCV:2010}.

\begin{figure*}
   \begin{center}
      \setlength\fboxsep{0pt}
      \begin{minipage}[t]{0.495\linewidth}
         \vspace{0pt}
         \begin{center}
         \begin{minipage}[t]{0.24\linewidth}
            \vspace{0pt}
            \begin{center}
               \fbox{\includegraphics[width=1.00\linewidth]{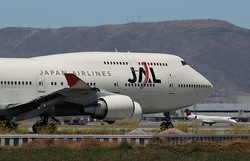}}\\
               \vspace{0.013\linewidth}
               \fbox{\includegraphics[width=1.00\linewidth]{figs/results_pascal_images_2008_000783_sm.jpg}}\\
               \vspace{0.013\linewidth}
               \fbox{\includegraphics[width=1.00\linewidth]{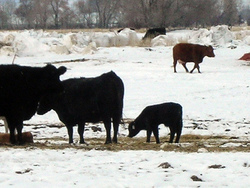}}\\
               \vspace{0.013\linewidth}
               \fbox{\includegraphics[width=1.00\linewidth]{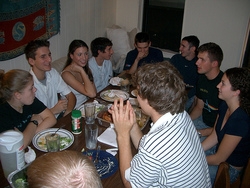}}\\
               \vspace{0.013\linewidth}
               \fbox{\includegraphics[width=1.00\linewidth]{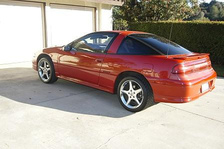}}\\
               \vspace{0.013\linewidth}
               \fbox{\includegraphics[width=1.00\linewidth]{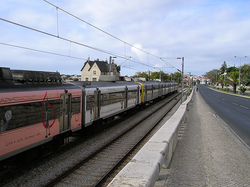}}\\
               \vspace{0.013\linewidth}
               \fbox{\includegraphics[width=1.00\linewidth]{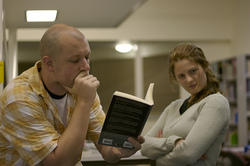}}\\
               \vspace{0.013\linewidth}
               \fbox{\includegraphics[width=1.00\linewidth]{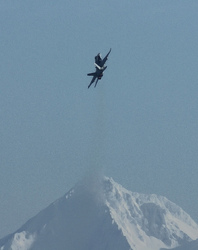}}\\
               \vspace{0.013\linewidth}
               \scriptsize{\textbf{\textsf{Image}}}
            \end{center}
         \end{minipage}
         \begin{minipage}[t]{0.24\linewidth}
            \vspace{0pt}
            \begin{center}
               \fbox{\includegraphics[width=1.00\linewidth]{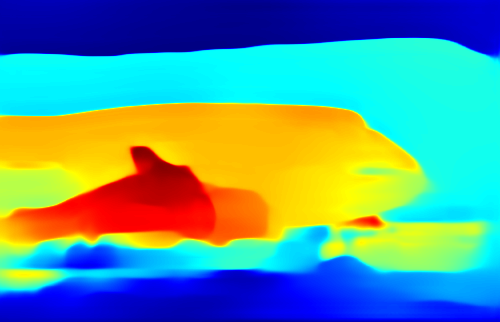}}\\
               \vspace{0.013\linewidth}
               \fbox{\includegraphics[width=1.00\linewidth]{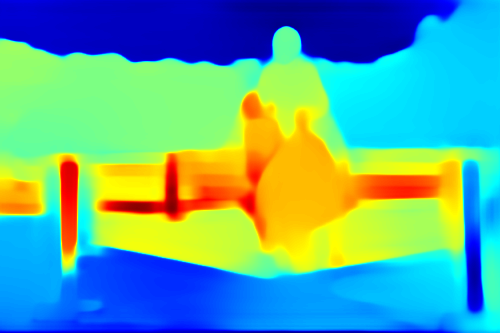}}\\
               \vspace{0.013\linewidth}
               \fbox{\includegraphics[width=1.00\linewidth]{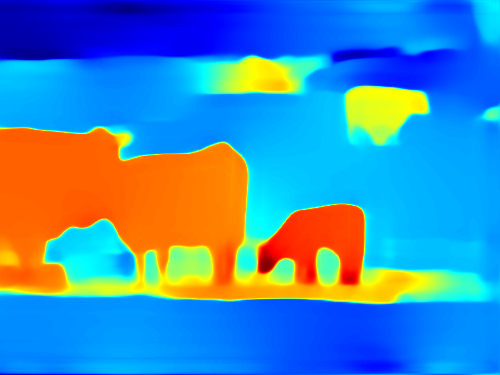}}\\
               \vspace{0.013\linewidth}
               \fbox{\includegraphics[width=1.00\linewidth]{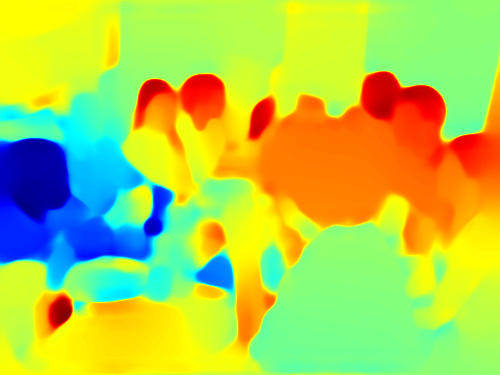}}\\
               \vspace{0.013\linewidth}
               \fbox{\includegraphics[width=1.00\linewidth]{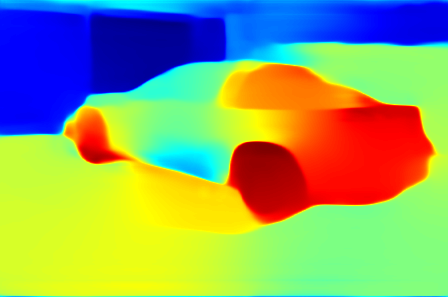}}\\
               \vspace{0.013\linewidth}
               \fbox{\includegraphics[width=1.00\linewidth]{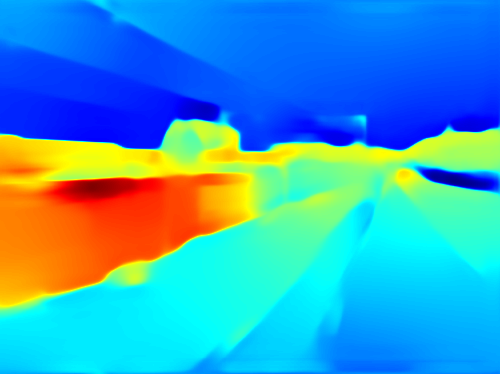}}\\
               \vspace{0.013\linewidth}
               \fbox{\includegraphics[width=1.00\linewidth]{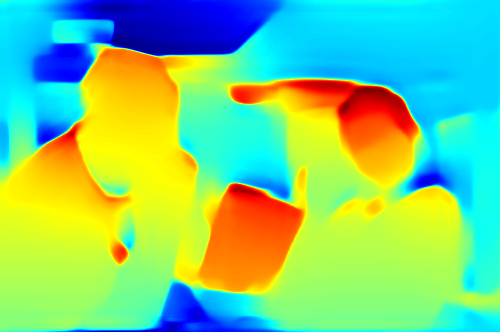}}\\
               \vspace{0.013\linewidth}
               \fbox{\includegraphics[width=1.00\linewidth]{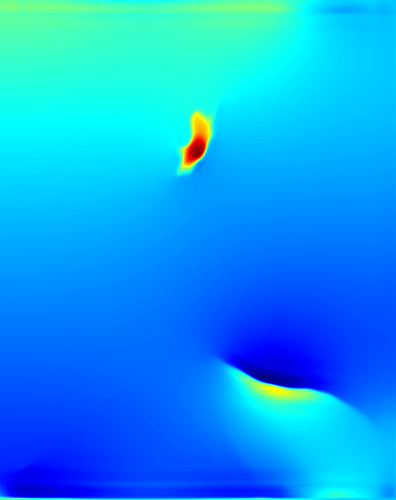}}\\
               \vspace{0.013\linewidth}
               \scriptsize{\textbf{\textsf{F/G}}}
            \end{center}
         \end{minipage}
         \begin{minipage}[t]{0.24\linewidth}
            \vspace{0pt}
            \begin{center}
               \fbox{\includegraphics[width=1.00\linewidth]{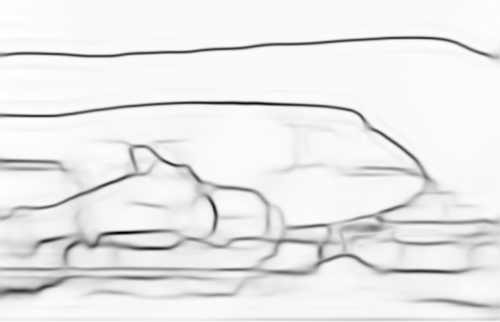}}\\
               \vspace{0.013\linewidth}
               \fbox{\includegraphics[width=1.00\linewidth]{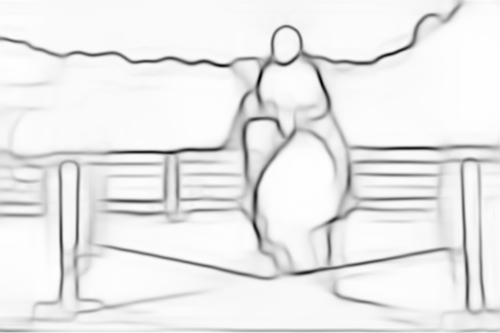}}\\
               \vspace{0.013\linewidth}
               \fbox{\includegraphics[width=1.00\linewidth]{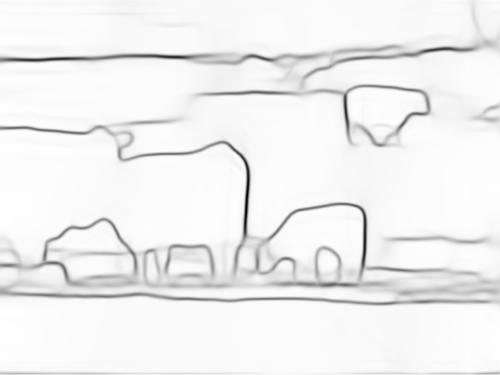}}\\
               \vspace{0.013\linewidth}
               \fbox{\includegraphics[width=1.00\linewidth]{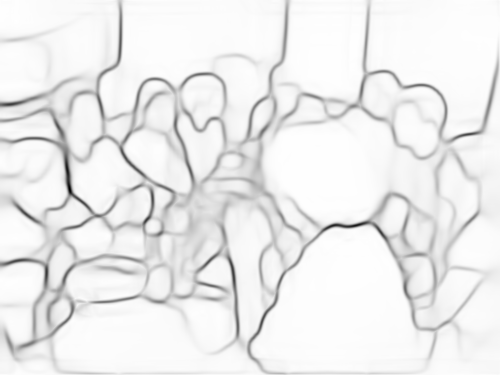}}\\
               \vspace{0.013\linewidth}
               \fbox{\includegraphics[width=1.00\linewidth]{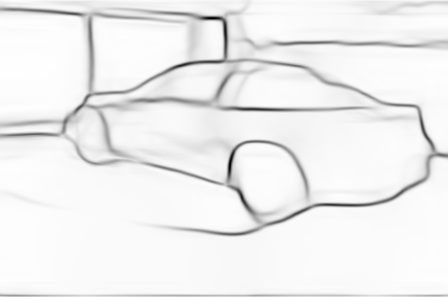}}\\
               \vspace{0.013\linewidth}
               \fbox{\includegraphics[width=1.00\linewidth]{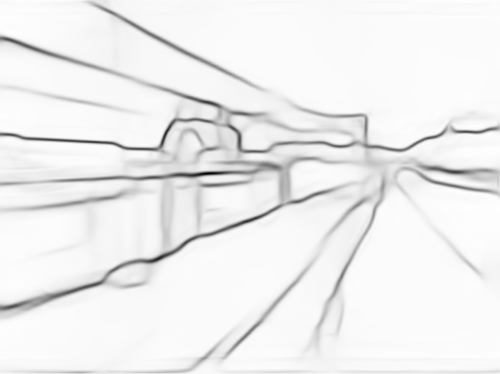}}\\
               \vspace{0.013\linewidth}
               \fbox{\includegraphics[width=1.00\linewidth]{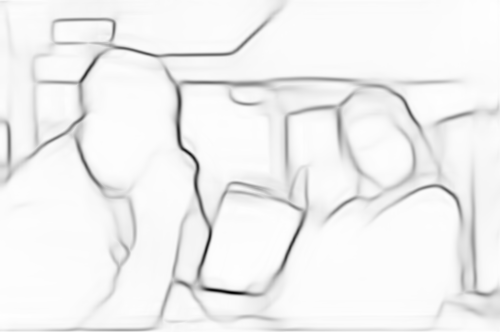}}\\
               \vspace{0.013\linewidth}
               \fbox{\includegraphics[width=1.00\linewidth]{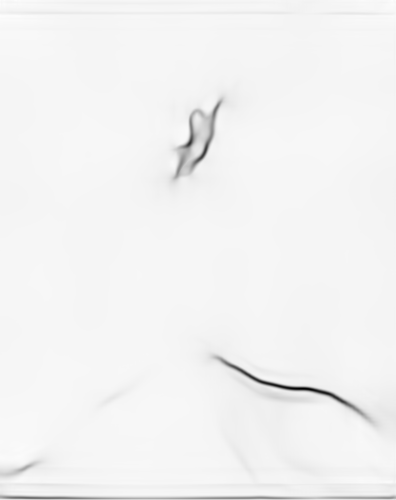}}\\
               \vspace{0.013\linewidth}
               \scriptsize{\textbf{\textsf{Boundaries}}}
            \end{center}
         \end{minipage}
         \begin{minipage}[t]{0.24\linewidth}
            \vspace{0pt}
            \begin{center}
               \fbox{\includegraphics[width=1.00\linewidth]{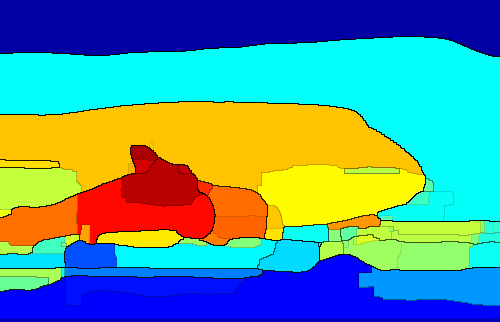}}\\
               \vspace{0.013\linewidth}
               \fbox{\includegraphics[width=1.00\linewidth]{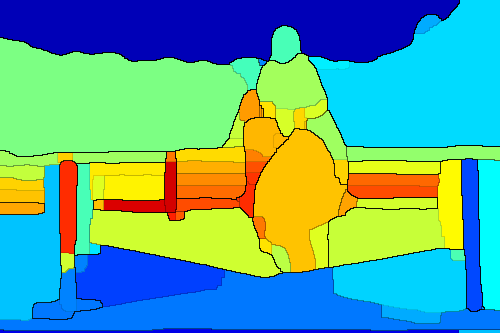}}\\
               \vspace{0.013\linewidth}
               \fbox{\includegraphics[width=1.00\linewidth]{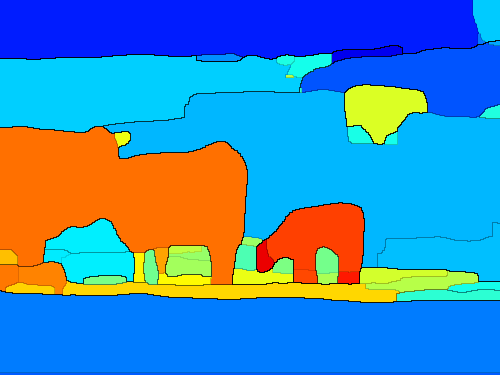}}\\
               \vspace{0.013\linewidth}
               \fbox{\includegraphics[width=1.00\linewidth]{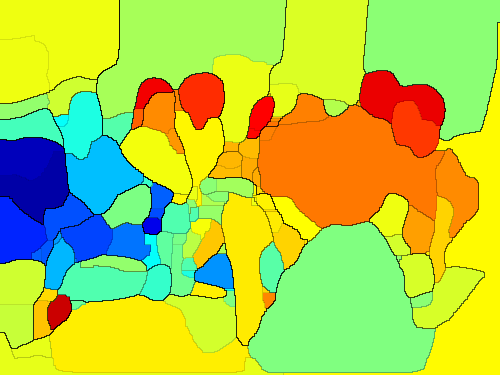}}\\
               \vspace{0.013\linewidth}
               \fbox{\includegraphics[width=1.00\linewidth]{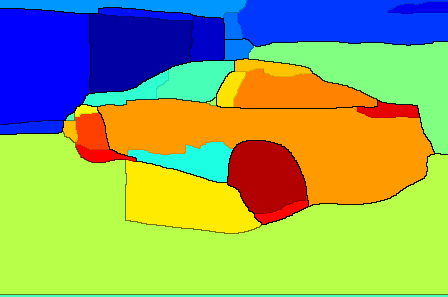}}\\
               \vspace{0.013\linewidth}
               \fbox{\includegraphics[width=1.00\linewidth]{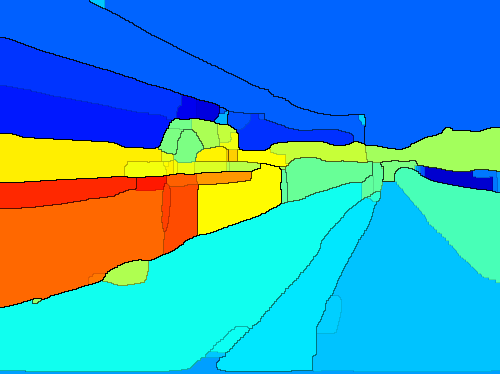}}\\
               \vspace{0.013\linewidth}
               \fbox{\includegraphics[width=1.00\linewidth]{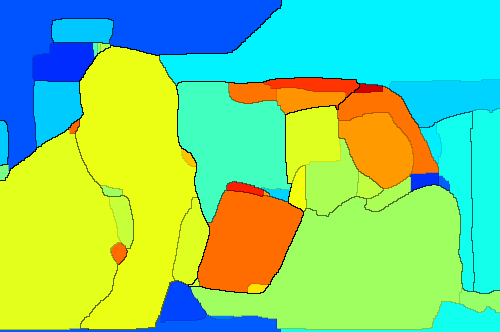}}\\
               \vspace{0.013\linewidth}
               \fbox{\includegraphics[width=1.00\linewidth]{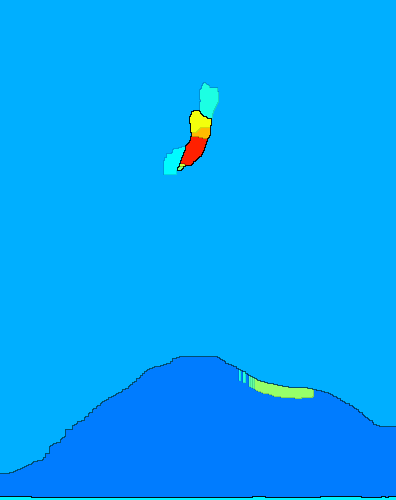}}\\
               \vspace{0.013\linewidth}
               \scriptsize{\textbf{\textsf{Seg + F/G}}}
            \end{center}
         \end{minipage}
         \end{center}
      \end{minipage}
      \hfill
      \begin{minipage}[t]{0.495\linewidth}
         \vspace{0pt}
         \begin{center}
         \begin{minipage}[t]{0.24\linewidth}
            \vspace{0pt}
            \begin{center}
               \fbox{\includegraphics[width=1.00\linewidth]{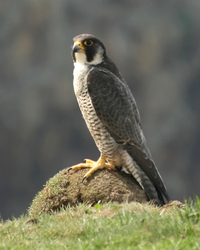}}\\
               \vspace{0.01\linewidth}
               \fbox{\includegraphics[width=1.00\linewidth]{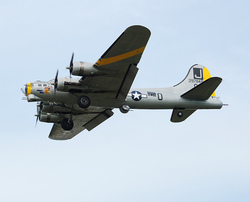}}\\
               \vspace{0.01\linewidth}
               \fbox{\includegraphics[width=1.00\linewidth]{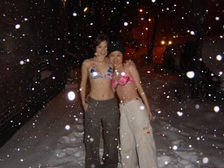}}\\
               \vspace{0.01\linewidth}
               \fbox{\includegraphics[width=1.00\linewidth]{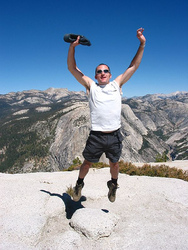}}\\
               \vspace{0.01\linewidth}
               \fbox{\includegraphics[width=1.00\linewidth]{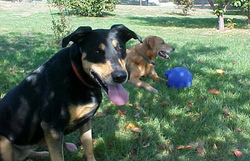}}\\
               \vspace{0.01\linewidth}
               \fbox{\includegraphics[width=1.00\linewidth]{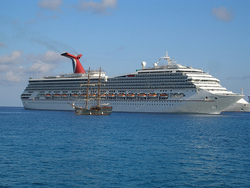}}\\
               \vspace{0.01\linewidth}
               \fbox{\includegraphics[width=1.00\linewidth]{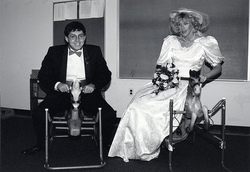}}\\
               \vspace{0.01\linewidth}
               \scriptsize{\textbf{\textsf{Image}}}
            \end{center}
         \end{minipage}
         \begin{minipage}[t]{0.24\linewidth}
            \vspace{0pt}
            \begin{center}
               \fbox{\includegraphics[width=1.00\linewidth]{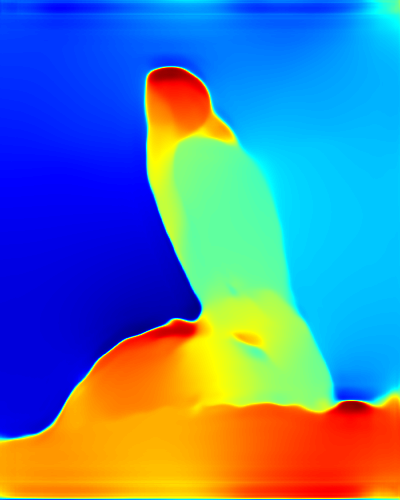}}\\
               \vspace{0.01\linewidth}
               \fbox{\includegraphics[width=1.00\linewidth]{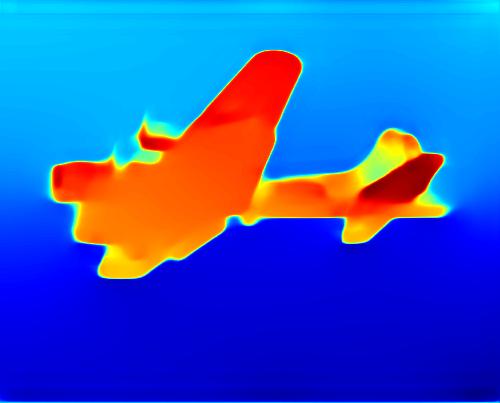}}\\
               \vspace{0.01\linewidth}
               \fbox{\includegraphics[width=1.00\linewidth]{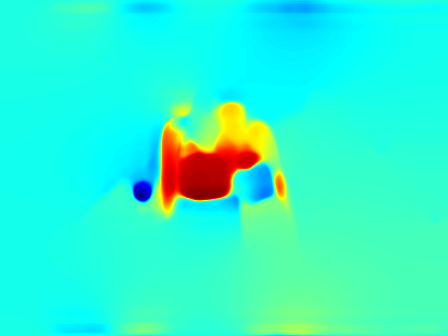}}\\
               \vspace{0.01\linewidth}
               \fbox{\includegraphics[width=1.00\linewidth]{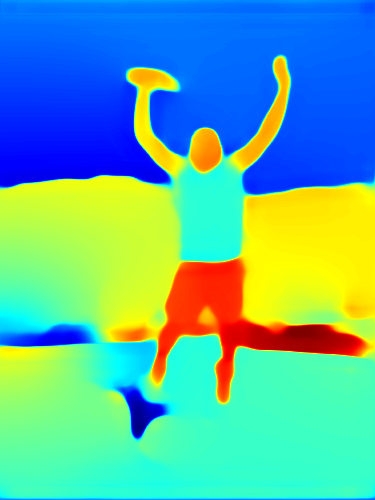}}\\
               \vspace{0.01\linewidth}
               \fbox{\includegraphics[width=1.00\linewidth]{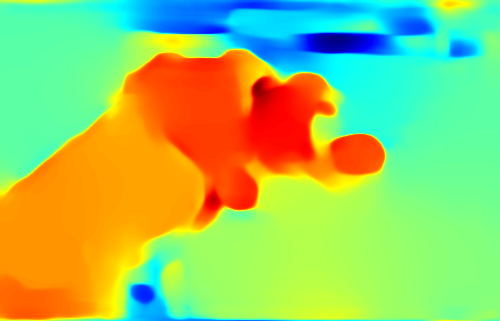}}\\
               \vspace{0.01\linewidth}
               \fbox{\includegraphics[width=1.00\linewidth]{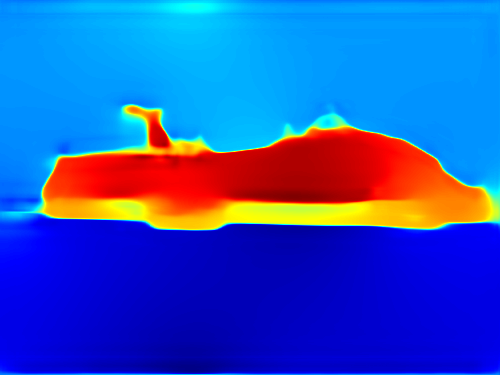}}\\
               \vspace{0.01\linewidth}
               \fbox{\includegraphics[width=1.00\linewidth]{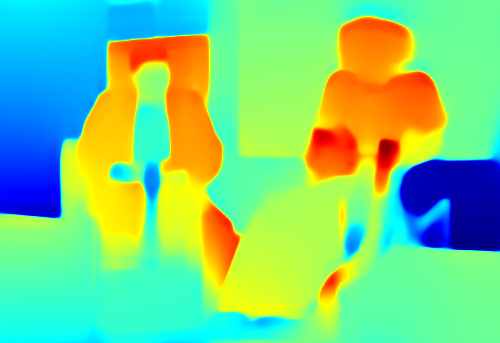}}\\
               \vspace{0.01\linewidth}
               \scriptsize{\textbf{\textsf{F/G}}}
            \end{center}
         \end{minipage}
         \begin{minipage}[t]{0.24\linewidth}
            \vspace{0pt}
            \begin{center}
               \fbox{\includegraphics[width=1.00\linewidth]{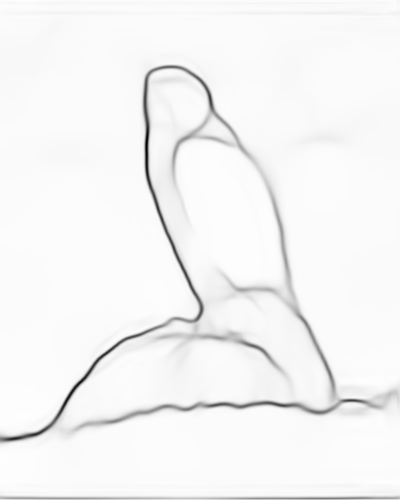}}\\
               \vspace{0.01\linewidth}
               \fbox{\includegraphics[width=1.00\linewidth]{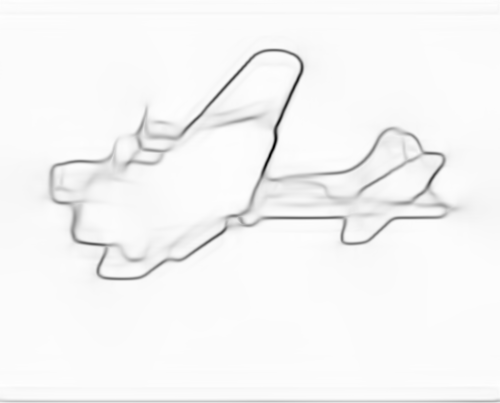}}\\
               \vspace{0.01\linewidth}
               \fbox{\includegraphics[width=1.00\linewidth]{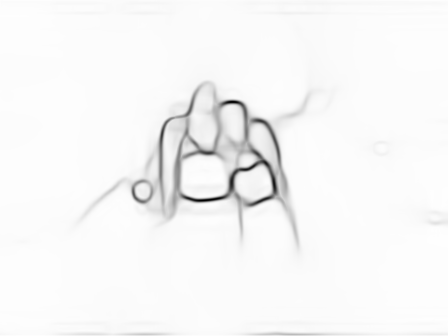}}\\
               \vspace{0.01\linewidth}
               \fbox{\includegraphics[width=1.00\linewidth]{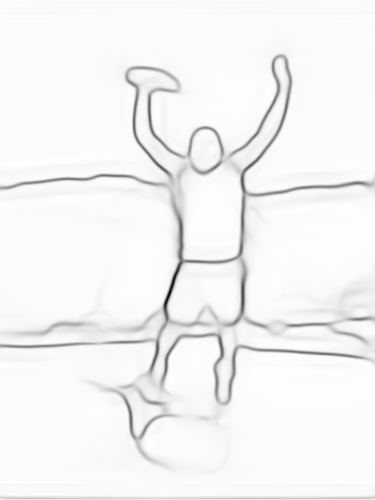}}\\
               \vspace{0.01\linewidth}
               \fbox{\includegraphics[width=1.00\linewidth]{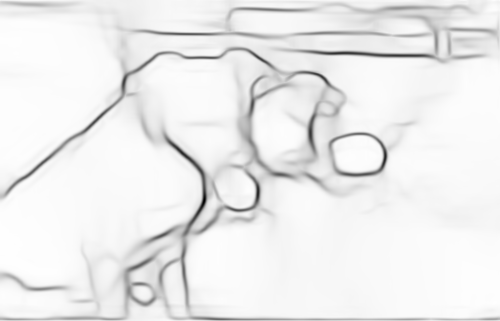}}\\
               \vspace{0.01\linewidth}
               \fbox{\includegraphics[width=1.00\linewidth]{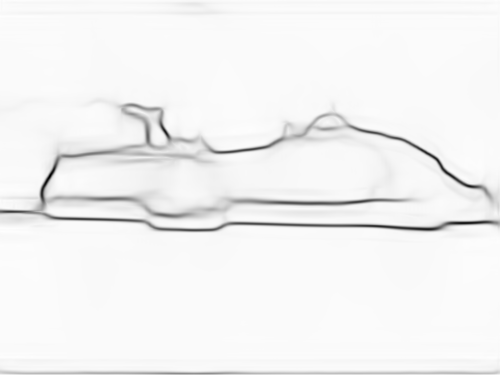}}\\
               \vspace{0.01\linewidth}
               \fbox{\includegraphics[width=1.00\linewidth]{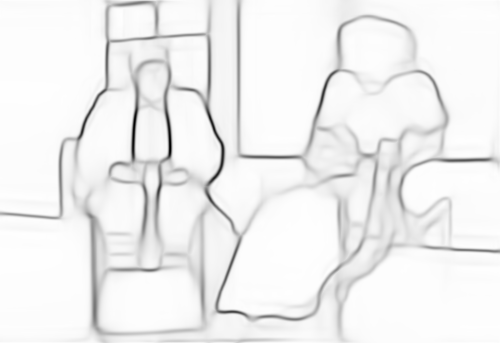}}\\
               \vspace{0.01\linewidth}
               \scriptsize{\textbf{\textsf{Boundaries}}}
            \end{center}
         \end{minipage}
         \begin{minipage}[t]{0.24\linewidth}
            \vspace{0pt}
            \begin{center}
               \fbox{\includegraphics[width=1.00\linewidth]{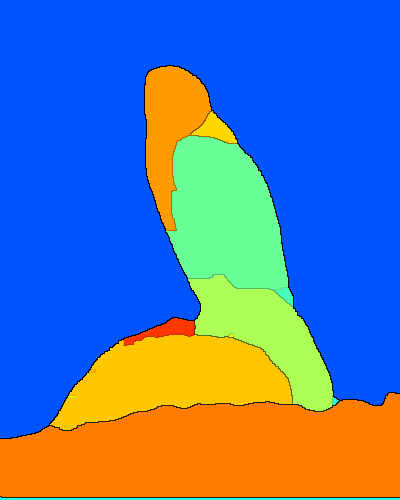}}\\
               \vspace{0.01\linewidth}
               \fbox{\includegraphics[width=1.00\linewidth]{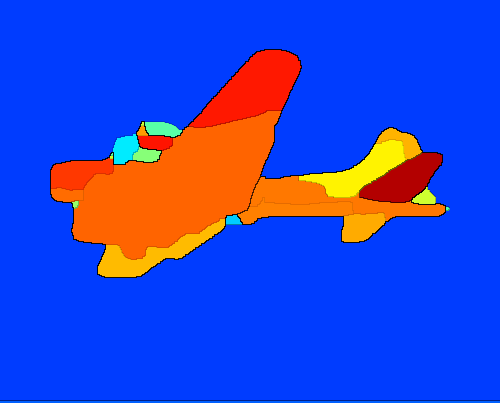}}\\
               \vspace{0.01\linewidth}
               \fbox{\includegraphics[width=1.00\linewidth]{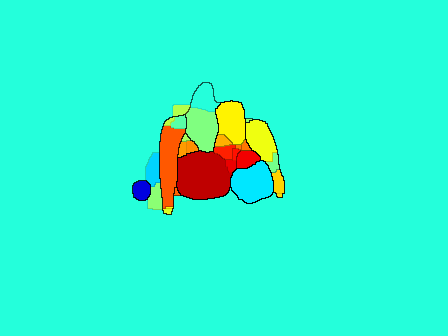}}\\
               \vspace{0.01\linewidth}
               \fbox{\includegraphics[width=1.00\linewidth]{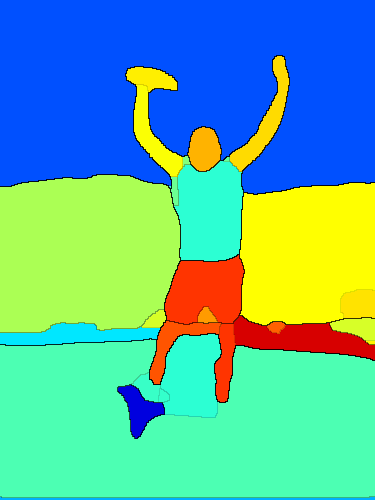}}\\
               \vspace{0.01\linewidth}
               \fbox{\includegraphics[width=1.00\linewidth]{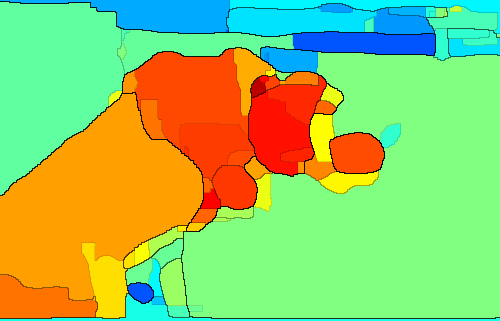}}\\
               \vspace{0.01\linewidth}
               \fbox{\includegraphics[width=1.00\linewidth]{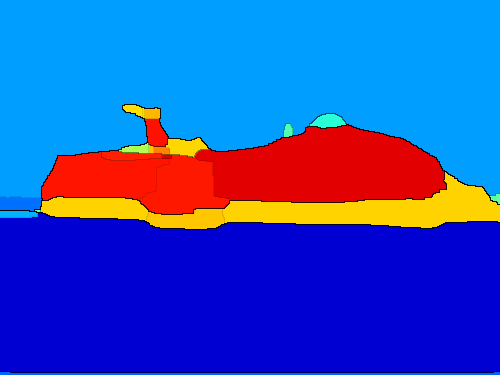}}\\
               \vspace{0.01\linewidth}
               \fbox{\includegraphics[width=1.00\linewidth]{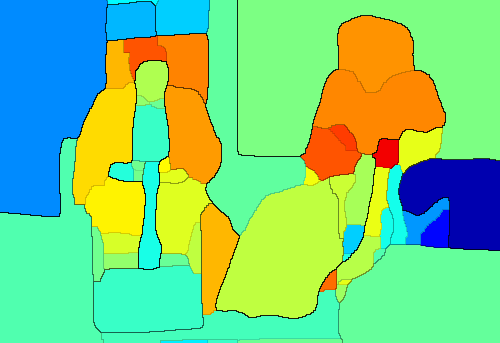}}\\
               \vspace{0.01\linewidth}
               \scriptsize{\textbf{\textsf{Seg + F/G}}}
            \end{center}
         \end{minipage}
         \end{center}
      \end{minipage}\\
      \vspace{1pt}
      \rule{0.975\linewidth}{0.5pt}\\
      \scriptsize{\textbf{\textsf{PASCAL VOC}}}
   \end{center}
   \vspace{-0.05\linewidth}
   \begin{center}
      \setlength\fboxsep{0pt}
      \begin{minipage}[t]{0.495\linewidth}
         \vspace{0pt}
         \begin{center}
         \begin{minipage}[t]{0.24\linewidth}
            \vspace{0pt}
            \begin{center}
               \fbox{\includegraphics[width=1.00\linewidth]{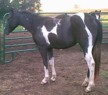}}\\
               \vspace{0.02\linewidth}
               \fbox{\includegraphics[width=1.00\linewidth]{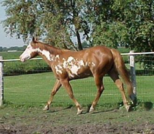}}\\
               \vspace{0.02\linewidth}
               \fbox{\includegraphics[width=1.00\linewidth]{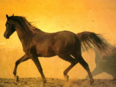}}\\
               \vspace{0.02\linewidth}
               \scriptsize{\textbf{\textsf{Image}}}
            \end{center}
         \end{minipage}
         \begin{minipage}[t]{0.24\linewidth}
            \vspace{0pt}
            \begin{center}
               \fbox{\includegraphics[width=1.00\linewidth]{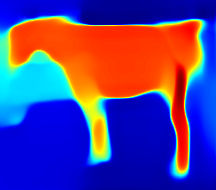}}\\
               \vspace{0.02\linewidth}
               \fbox{\includegraphics[width=1.00\linewidth]{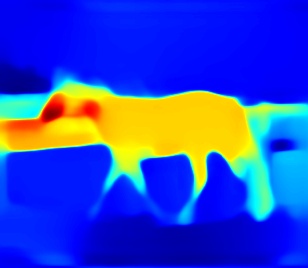}}\\
               \vspace{0.02\linewidth}
               \fbox{\includegraphics[width=1.00\linewidth]{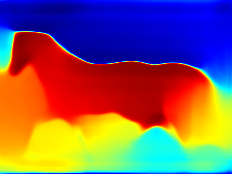}}\\
               \vspace{0.02\linewidth}
               \scriptsize{\textbf{\textsf{F/G}}}
            \end{center}
         \end{minipage}
         \begin{minipage}[t]{0.24\linewidth}
            \vspace{0pt}
            \begin{center}
               \fbox{\includegraphics[width=1.00\linewidth]{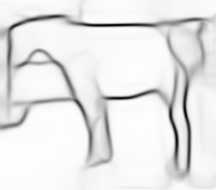}}\\
               \vspace{0.02\linewidth}
               \fbox{\includegraphics[width=1.00\linewidth]{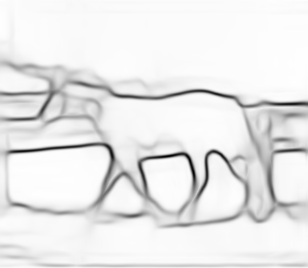}}\\
               \vspace{0.02\linewidth}
               \fbox{\includegraphics[width=1.00\linewidth]{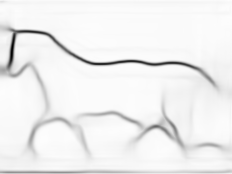}}\\
               \vspace{0.02\linewidth}
               \scriptsize{\textbf{\textsf{Boundaries}}}
            \end{center}
         \end{minipage}
         \begin{minipage}[t]{0.24\linewidth}
            \vspace{0pt}
            \begin{center}
               \fbox{\includegraphics[width=1.00\linewidth]{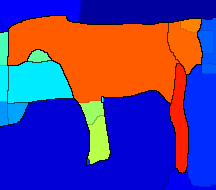}}\\
               \vspace{0.02\linewidth}
               \fbox{\includegraphics[width=1.00\linewidth]{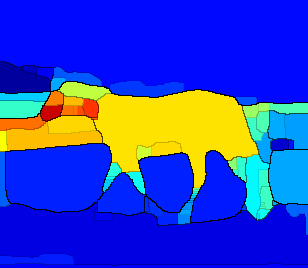}}\\
               \vspace{0.02\linewidth}
               \fbox{\includegraphics[width=1.00\linewidth]{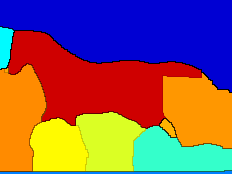}}\\
               \vspace{0.02\linewidth}
               \scriptsize{\textbf{\textsf{Seg + F/G}}}
            \end{center}
         \end{minipage}
         \end{center}
      \end{minipage}
      \hfill
      \begin{minipage}[t]{0.495\linewidth}
         \vspace{0pt}
         \begin{center}
         \begin{minipage}[t]{0.24\linewidth}
            \vspace{0pt}
            \begin{center}
               \fbox{\includegraphics[width=1.00\linewidth]{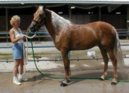}}\\
               \vspace{0.01\linewidth}
               \fbox{\includegraphics[width=1.00\linewidth]{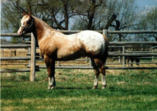}}\\
               \vspace{0.01\linewidth}
               \fbox{\includegraphics[width=1.00\linewidth]{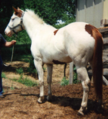}}\\
               \vspace{0.01\linewidth}
               \scriptsize{\textbf{\textsf{Image}}}
            \end{center}
         \end{minipage}
         \begin{minipage}[t]{0.24\linewidth}
            \vspace{0pt}
            \begin{center}
               \fbox{\includegraphics[width=1.00\linewidth]{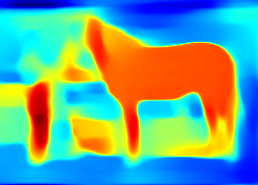}}\\
               \vspace{0.01\linewidth}
               \fbox{\includegraphics[width=1.00\linewidth]{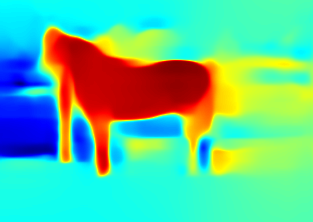}}\\
               \vspace{0.01\linewidth}
               \fbox{\includegraphics[width=1.00\linewidth]{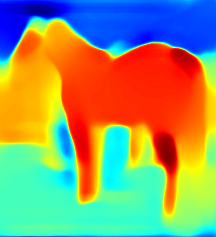}}\\
               \vspace{0.01\linewidth}
               \scriptsize{\textbf{\textsf{F/G}}}
            \end{center}
         \end{minipage}
         \begin{minipage}[t]{0.24\linewidth}
            \vspace{0pt}
            \begin{center}
               \fbox{\includegraphics[width=1.00\linewidth]{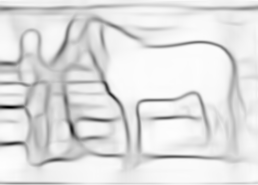}}\\
               \vspace{0.01\linewidth}
               \fbox{\includegraphics[width=1.00\linewidth]{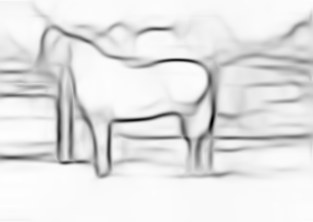}}\\
               \vspace{0.01\linewidth}
               \fbox{\includegraphics[width=1.00\linewidth]{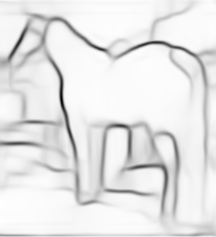}}\\
               \vspace{0.01\linewidth}
               \scriptsize{\textbf{\textsf{Boundaries}}}
            \end{center}
         \end{minipage}
         \begin{minipage}[t]{0.24\linewidth}
            \vspace{0pt}
            \begin{center}
               \fbox{\includegraphics[width=1.00\linewidth]{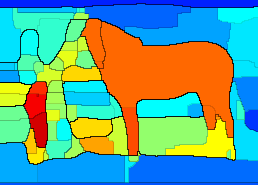}}\\
               \vspace{0.01\linewidth}
               \fbox{\includegraphics[width=1.00\linewidth]{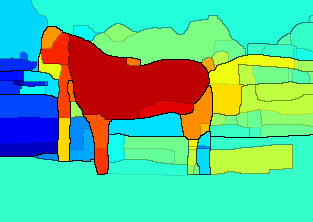}}\\
               \vspace{0.01\linewidth}
               \fbox{\includegraphics[width=1.00\linewidth]{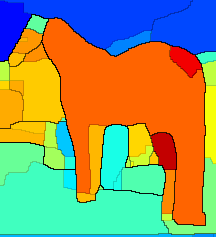}}\\
               \vspace{0.01\linewidth}
               \scriptsize{\textbf{\textsf{Seg + F/G}}}
            \end{center}
         \end{minipage}
         \end{center}
      \end{minipage}\\
      \vspace{1pt}
      \rule{0.975\linewidth}{0.5pt}\\
      \scriptsize{\textbf{\textsf{Weizmann Horses}}}
   \end{center}
   \vspace{-0.01\linewidth}
   \caption{
      \textbf{Cross-domain generalization.}
         Our system, trained only on the Berkeley segmentation dataset,
         generalizes to other datasets.  Here, we test on images from
         the PASCAL VOC dataset~\cite{PASCAL} and the Weizmann Horse
         database~\cite{BU:PAMI:2008}.  On PASCAL, our figure/ground reflects
         object attention and scene layering.  On the Weizmann images, our
         system acts essentially as a horse detection and segmentation
         algorithm despite having no object-specific training.  Its generic
         understanding of figure/ground suffices to automatically pick out
         objects.
   }
   \label{fig:pascal-horse-results}
\end{figure*}

\subsection{Additional Datasets}
\label{sec:horses-pascal}

Figure~\ref{fig:pascal-horse-results} demonstrates that our BSDS-trained
system captures generally-applicable notions of both segmentation and
figure/ground.  On both PASCAL VOC~\cite{PASCAL} and the Weizmann Horse
database~\cite{BU:PAMI:2008}, it generates figure/ground layering that
respects scene organization.  On the Weizmann examples, though having only
been trained for perceptual organization, it behaves like an object detector.

\section{Conclusion}
\label{sec:conclusion}

We demonstrate that Angular Embedding, acting on CNN predictions about pairwise
pixel relationships, provides a powerful framework for segmentation and
figure/ground organization.  Our work is the first to formulate a robust
interface between these two components.  Our results are a dramatic improvement
over prior attempts to use spectral methods for figure/ground organization.

{\small
\bibliographystyle{ieee}
\bibliography{affinity-cnn}
}

\end{document}